\documentclass{article}
\usepackage{PRIMEarxiv}
\usepackage[utf8]{inputenc}
\usepackage[T1]{fontenc}

\usepackage{amsmath,amssymb,amsfonts}
\usepackage{physics}
\usepackage{graphicx}
\usepackage{subcaption}
\usepackage{booktabs}
\usepackage{nicefrac}
\usepackage{microtype}
\usepackage{lipsum}
\usepackage{fancyhdr}
\usepackage{url}
\usepackage{hyperref}
\usepackage{orcidlink}
\usepackage{placeins}
\usepackage{float}
\usepackage{tabularx}
\usepackage{array}

\graphicspath{{media/}}
\hypersetup{colorlinks,citecolor=blue,linkcolor=red}
\title{Geometric Kolmogorov--Arnold Network (GeoKAN)} 

\author{
  Abhijit Sen \orcidlink{0000-0003-2783-1763} \textsuperscript{1*}, 
  Bikram Keshari Parida \orcidlink{0000-0003-1204-357X}\textsuperscript{1*},
  Giridas Maiti \orcidlink{0000-0002-7813-6480}\textsuperscript{2},
  Mahima Arya \orcidlink{0000-0002-1847-9705}\textsuperscript{1}, 
  Denys I. Bondar\orcidlink{0000-0002-3626-4804}\textsuperscript{1}\\
  \textsuperscript{1} Department of Physics and Engineering Physics, Tulane University,  New Orleans, Louisiana 70118, USA. \\
    \textsuperscript{2} Institute of Applied Geosciences, Karlsruhe Institute of Technology, Karlsruhe 76131, Germany. \\
  \texttt{\{asen1, bparida, dbondar\}@tulane.edu}  \\
  \textsuperscript{*} Equal Contribution. \\
}

\begin{document}
\maketitle

\begin{abstract}

We introduce \emph{Geometric Kolmogorov--Arnold Networks} (GeoKANs), a family of geometry-aware KAN-type models in which approximation is carried out in learned, geometry-adapted coordinates rather than in fixed Euclidean input coordinates. GeoKAN achieves this by learning a diagonal Riemannian metric that warps the input before basis expansion and feature mixing. The learned metric provides a geometric inductive bias through local length scaling and volume distortion, and in physics-informed settings it also affects the differential structure seen by the model. Within this framework, we develop three main variants, namely \emph{GeoKAN-NNMetric}, \emph{GeoKAN-$\gamma$}, and \emph{LM-KAN}. For \emph{LM-KAN}, we further consider three basis-specific versions, \emph{LM-KAN-RBF}, \emph{LM-KAN-Wav}, and \emph{LM-KAN-Fourier}. These variants allow us to study geometry-aware KAN models both as general function approximators and as surrogates in physics-informed learning. By stretching regions with rapid variation and compressing smoother regions, GeoKAN reallocates representational resolution in a task-dependent manner, allowing the model to place capacity where it is most needed. As a result, GeoKAN is well suited to sharp, stiff, localized, and strongly non-uniform regimes arising in scientific machine learning and differential-equation problems. Code is available at \href{https://github.com/AI-and-Quantum-Computing/GeoKAN}{https://github.com/AI-and-Quantum-Computing/GeoKAN}.

\end{abstract}

 \keywords{KAN \and Geometric KAN \and Deep Learning \and PINN \and PIKAN \and LM-KAN \and Machine Learning}

\section{Introduction}

Recently, Kolmogorov--Arnold Network (KAN) has emerged as a promising alternative to standard multilayer perceptrons \cite{KAN}. KAN is motivated by the Kolmogorov--Arnold representation theorem, which shows that multivariate continuous functions can be represented through sums and compositions of univariate functions \cite{KAT3, KAT2}. Unlike ordinary neural networks that use fixed activation functions, KAN employs learnable nonlinear functions, which gives it greater flexibility in approximation \cite{KAN}. Because of this, KAN has shown promising performance in function approximation and in the numerical solution of differential equations \cite{wang_kolmogorov_2024, koenig_kan-odes_2024, PIKAN1}. In particular, KAN-based physics-informed models have shown that they can often capture the solutions of differential equations more accurately than standard neural networks \cite{wang_kolmogorov_2024, PIKAN1, PIKAN_JMLR_2025,Sen2026EhrenfestKAN}.

Despite this flexibility, most existing KAN models still operate in a fixed coordinate system. Their basis functions are learned, but they are still evaluated directly on the original input coordinates. This can become a limitation when the target function is highly nonuniform across the domain. In many problems, especially in scientific computing, the solution is smooth in some regions but varies rapidly in others, for example near shocks, thin interfaces, sharp peaks, or boundary layers \cite{PIKAN_JMLR_2025,Sen2026EhrenfestKAN}. In such cases, a fixed representation may spend too much capacity in simple regions and not enough in the difficult ones.

This limitation motivates the main idea of the paper. Instead of learning only the approximation, we also learn the geometry of the input space. To this end, we introduce \emph{Geometric KAN} (GeoKAN), a family of geometry-aware KAN models in which the input is first warped by a learned metric and the representation is then constructed in geometry-adapted coordinates. In this way, the model can stretch regions where the target is difficult and compress regions where it is smooth, so that representational resolution is allocated more effectively across the domain. The GeoKAN framework gives rise to several variants depending on how the metric is parameterized and how the post-warp features are constructed. In this work, we introduce three main variants of GeoKAN. (1) \emph{GeoKAN-NNMetric} learns a coupled metric from the full input and combines it with a localized wavelet dictionary. (2) \emph{GeoKAN-$\gamma$} uses a more structured separable metric and builds explicit geometric features from the learned scaling and its local variation. (3) \emph{Learned-Metric KAN} (LM-KAN) uses a coupled learned metric together with localized post-warp basis functions, and serves as the main geometry-aware surrogate in the physics-informed setting considered later.

We first apply GeoKAN in a supervised function-approximation setting through matched-capacity curve-fitting benchmarks. This makes it possible to evaluate the representational behavior of the architecture before introducing differential-equation residuals, boundary conditions, or initial conditions. In this way, the curve-fitting experiments test whether geometry-aware KAN models provide an intrinsic approximation advantage on targets with oscillatory, localized, discontinuous, or multiscale structure.

We then use GeoKAN to construct physics-informed models. Physics-Informed Neural Networks (PINNs) provide a widely used framework for solving differential equations by incorporating the governing equation directly into the loss function \cite{PINN1}. Since their introduction, PINNs have been extended to many classes of problems, including fractional equations, integro-differential equations, and stochastic partial differential equations \cite{PINNReview, PINNother1, PINNother2, PINNother3}. Despite this progress, PINNs still face several challenges, including optimization instability, sensitivity to hyperparameters, and difficulty in handling solutions with sharp gradients, shocks, or boundary layers \cite{PINNdrawback2, PINNdrawback3, PINNdrawback4, PINNdrawback5}. A number of remedies have therefore been proposed to improve their robustness and training behavior \cite{PINNremedy1}. One recent direction is to replace the standard multilayer perceptron surrogate by a KAN-type model, leading to Physics-Informed KAN models, often referred to as PIKAN \cite{PIKAN1,PIKAN_JMLR_2025,Sen2026EhrenfestKAN}. Within this setting, we use the \emph{LM-KAN} variant from the broader GeoKAN family as the physics-informed surrogate. The training framework remains the standard residual-based one, but the surrogate now learns a task-dependent metric and evaluates its localized basis representation in metric-adapted coordinates. This allows a direct comparison with earlier PIKAN models based on existing KAN variants, while keeping the surrounding physics-informed methodology essentially unchanged.

Altogether this work shows GeoKAN contribution at two levels. The curve-fitting benchmarks examine its approximation properties in isolation. The physics-informed experiments examine whether the same geometry-aware mechanism improves the solution of differential equations in practice. The rest of the paper is organized as follows. Section~\ref{KAN_int} reviews the background on KAN architecture. Section~\ref{GeoKAN_int} presents the GeoKAN framework and its main variants, together with the role of the learned metric in geometry-adapted representation. Section~\ref{DataFitting_int} studies matched-capacity data-fitting benchmarks to isolate the approximation behavior of the models. Section~\ref{pikan_lmkan_results} then turns to physics-informed learning and compares the LM-KAN surrogate with earlier PIKAN models. Finally, Section~\ref{diss} summarizes the main conclusions and discusses possible future directions.

\section{Brief Overview of KAN Architecture}
\label{KAN_int}

Before introducing Kolmogorov--Arnold Networks (KANs), we briefly recall the standard deep neural network (DNN), including the multilayer perceptron (MLP), whose approximation capability is classically justified by the universal approximation theorem. In one standard form, if $\sigma$ is a continuous sigmoidal activation function, then functions of the form
\begin{equation}
k(\mathbf{x})=\sum_{j=1}^{N}\alpha_j\,\sigma\!\left(\mathbf{w}_j^{T}\mathbf{x}+b_j\right),
\qquad
\mathbf{x},\mathbf{w}_j\in\mathbb{R}^n,\;\alpha_j,b_j\in\mathbb{R},
\end{equation}
are dense in the space of continuous functions on $[0,1]^n$ \cite{UAT}. Hence, for any continuous target function $\sigma(\mathbf{x})$ and any $\varepsilon>0$, there exists such a network satisfying
\begin{equation}
|k(\mathbf{x})-\sigma(\mathbf{x})|<\varepsilon
\end{equation}
on the given compact domain \cite{UAT1,UAT2}. This result provides the theoretical basis for using sufficiently wide feed-forward networks to approximate nonlinear maps.

\begin{figure}
    \centering
    \includegraphics[width=\textwidth]{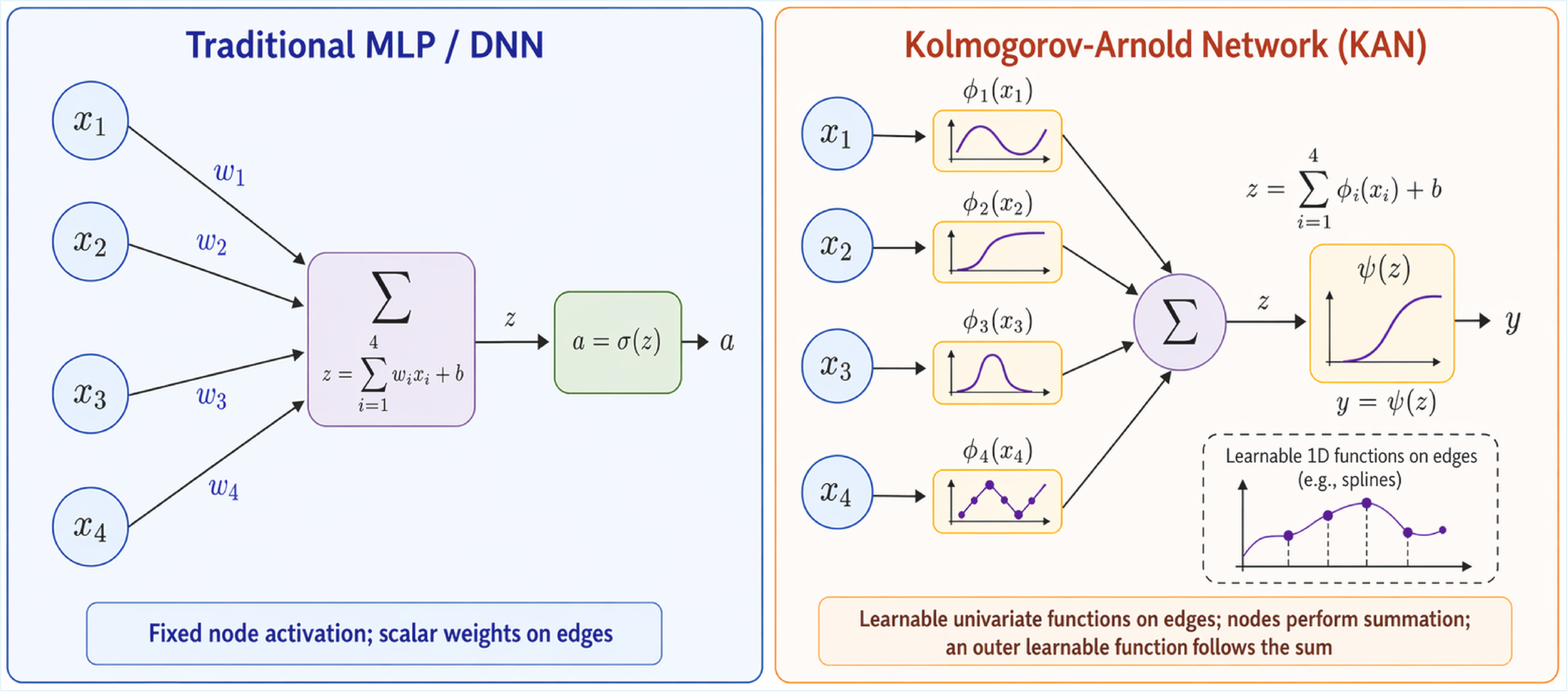}
    \caption{Comparison of the fundamental input-to-node operation in a traditional DNN and a KAN. For simplicity, the figure focuses on the local neuron/layer mechanism rather than the full deep architecture. In a traditional DNN, inputs are first combined through learnable scalar weights and summed, after which a fixed nonlinearity is applied at the node. In a KAN, by contrast, each input-edge carries its own learnable univariate nonlinear function, and the node primarily performs additive aggregation of these transformed inputs.}
    \label{fig:KAN vs DNN}
\end{figure}

In a standard feed-forward network, each neuron first forms an affine combination of its inputs and then applies a fixed nonlinear activation (Fig.~\ref{fig:KAN vs DNN}). Thus, if $\{x_i\}_{i=1}^n$ are the inputs, with weights $\{w_i\}_{i=1}^n$ and bias $b$, then the neuron computes
\begin{equation}
z=\sum_{i=1}^{n} w_i x_i+b,
\end{equation}
followed by
\begin{equation}
a=\sigma(z),
\end{equation}
where $\sigma$ is typically chosen in advance, for example ReLU, $\tanh$, or GELU. In this framework, the trainable parameters are the linear weights and biases, while the nonlinearity is fixed at the node level.

KAN is based instead on the Kolmogorov--Arnold representation theorem, which states that any continuous multivariate function can be represented through superpositions of continuous univariate functions and addition \cite{KAT3,KAT2,KAT1}. More precisely, if $f:[0,1]^n\to\mathbb{R}$ is continuous, then there exist continuous univariate functions $\phi_{i,j}$ and $\psi_i$ such that
\begin{equation}
f(x_1,\dots,x_n)
=
\sum_{i=0}^{2n}
\psi_i\!\left(\sum_{j=1}^{n}\phi_{i,j}(x_j)\right).
\end{equation}
This theorem, originating from the resolution of Hilbert's thirteenth problem by Kolmogorov and its refinement by Arnold, provides the conceptual foundation for KAN architectures (Fig.~\ref{fig:KAN vs DNN}).

\subsection{Network Architectures of KAN in Comparison to Traditional DNN}
Although earlier attempts had explored neural architectures motivated by the Kolmogorov--Arnold theorem \cite{FKAT1,FKAT2,FKAT3,FKAT4,FKAT5,FKAT6}, the work in Ref. \cite{KAN} gave a practical and systematic realization of this idea. In KAN, the role of a scalar edge weight is replaced by a learnable univariate nonlinear function. In practice, these edge functions are often parameterized by spline representations, especially B-splines, whose coefficients are learned during training.

The structural contrast with a traditional DNN is immediate. In a standard DNN, a neuron first forms a weighted sum of its inputs and only then applies a fixed nonlinearity:
\begin{equation}
z=\sum_{i=1}^{n} w_i x_i+b,
\qquad
a=\sigma(z).
\label{eq:dnn_basic}
\end{equation}
Thus, the learnable parameters are the scalar weights $w_i$, while the activation function $\sigma$ is prescribed in advance.

In a KAN, the order is reversed at the local level. Each input is first transformed by its own learnable univariate map $\phi_i:\mathbb{R}\to\mathbb{R}$, and the node then aggregates these transformed signals additively:
\begin{equation}
z=\sum_{i=1}^{n}\phi_i(x_i)+b.
\label{eq:kan_scalar}
\end{equation}

Because the edge maps \(\phi_i\) are themselves nonlinear and trainable, a separate fixed activation at the node is not essential. If desired, one may still apply an additional outer univariate map \(\psi\), giving
\begin{equation}
y=\psi(z).
\label{eq:kan_outer}
\end{equation}
This expression should be understood as a single KAN node, or equivalently as one outer-function component of the Kolmogorov--Arnold representation. The full Kolmogorov--Arnold form contains multiple such outer functions, typically indexed by \(q\), whose outputs are summed.

This simple scalar picture can also be written in the same feature-construction language used later for GeoKAN. Define the KAN feature vector
\begin{equation}
F_{\mathrm{KAN}}(x)
=
\big[\phi_1(x_1),\phi_2(x_2),\dots,\phi_n(x_n)\big]^\top.
\label{eq:kan_feature_vector}
\end{equation}
Here, \(F_{\mathrm{KAN}}(x)\in\mathbb{R}^{n}\) collects the learned univariate edge features associated with the input coordinates \(x_1,\dots,x_n\). The node output may then be viewed as a linear mixing of these learned edge features:
\begin{equation}
z =  \mathbb{I}^\top F_{\mathrm{KAN}}(x)+b,
\label{eq:kan_feature_form}
\end{equation}
where \(\mathbb{I}\in\mathbb{R}^{n}\) denotes the all-ones vector, so that
\(\mathbb{I}^{\top}F_{\mathrm{KAN}}(x)=\sum_{i=1}^{n}\phi_i(x_i)\), and \(b\in\mathbb{R}\) is a scalar bias term.

More generally, at the layer level, the same construction can be written as
\begin{equation}
\tilde h^{(\ell)} = W_{\mathrm{KAN}}^{(\ell)}F_{\mathrm{KAN}}^{(\ell)} + b_{\mathrm{KAN}}^{(\ell)}.
\label{eq:kan_layer_form}
\end{equation}
Here, \(F_{\mathrm{KAN}}^{(\ell)}\) denotes the vector of learned univariate edge features at layer \(\ell\), \(W_{\mathrm{KAN}}^{(\ell)}\) is a trainable linear mixing matrix, \(b_{\mathrm{KAN}}^{(\ell)}\) is the corresponding bias vector, and \(\tilde h^{(\ell)}\) denotes the pre-activation output of the layer.

Thus, standard KAN may be interpreted as a feature-construction architecture in which the features are learnable univariate edge functions, whereas the node itself mainly performs additive aggregation. This perspective is useful because it places KAN and GeoKAN in a common framework. Both first construct a set of nonlinear features and then linearly combine them. The essential difference is that in KAN the features are built directly from one-dimensional edge maps, whereas in GeoKAN they are constructed after learning a geometry-dependent warp of the representation.




For the sake of understanding, the final output in a DNN is denoted as $a$ (see Eq. \eqref{eq:dnn_basic} ), while in a KAN, it is denoted as $y$ (see Eq. \eqref{eq:kan_outer}). This distinction is intentional to avoid confusion during comparison or analysis.

Hence, whereas a DNN learns linear weights followed by a fixed node activation, a KAN places the learnable nonlinearities directly on the edges. The node in a KAN mainly performs summation, while the trainable parameters specify the univariate edge functions.

\begin{figure}
    \centering
    \includegraphics[width=\textwidth]{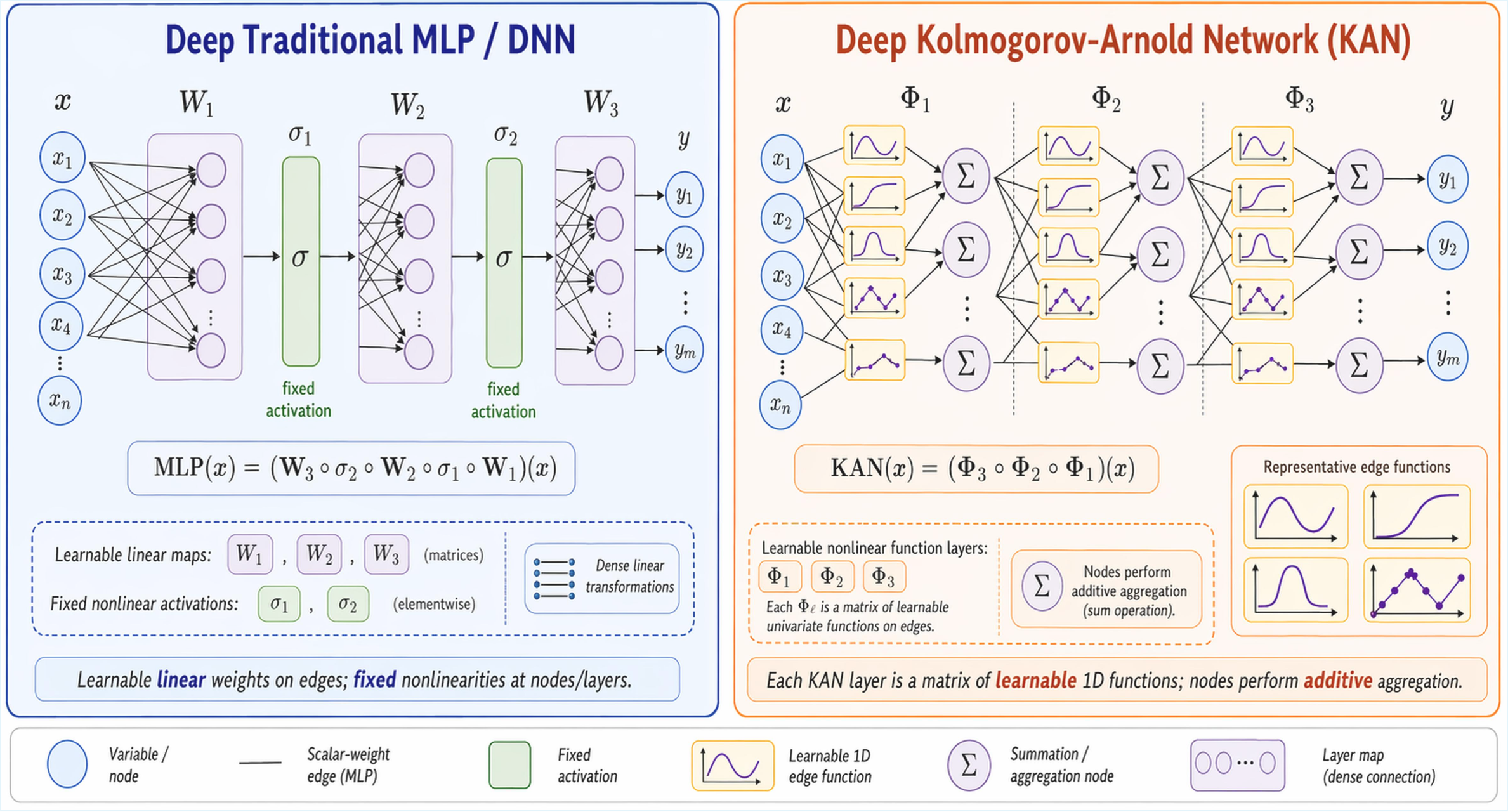}
    \caption{Comparison of deep architectures in a traditional DNN and a KAN. For clarity, the figure emphasizes the basic stacked structure rather than full implementation detail. A traditional DNN is built by alternating learnable linear maps with fixed activations, whereas a KAN is built by stacking layers of learnable univariate edge functions whose outputs are additively aggregated at nodes.
}
    \label{fig:deep_KAN vs DNN}
\end{figure}

This distinction extends naturally to depth. A deep DNN is built by stacking linear maps and fixed activations (Fig.~\ref{fig:deep_KAN vs DNN}), whereas a deep KAN is formed by stacking layers whose entries are learnable univariate functions (Fig.~\ref{fig:deep_KAN vs DNN}). Denoting the $\ell^{th}$ KAN layer by $\Phi^{(\ell)}$, the resulting architecture may be written as
\begin{equation}
\mathrm{KAN}(x)=\Phi^{(L-1)}\circ \Phi^{(L-2)}\circ \cdots \circ \Phi^{(0)}(x),
\label{eq:deep-kan}
\end{equation}
where each $\Phi^{(\ell)}$ represents a matrix of learnable one-dimensional functions acting on edges, followed by additive aggregation at the nodes. In this way, the single-node picture shown in the earlier figure extends directly to deep architectures of arbitrary width and depth.

\subsection{Variants of KAN in the Literature}

\begin{figure}
    \centering
    \includegraphics[width=\textwidth]{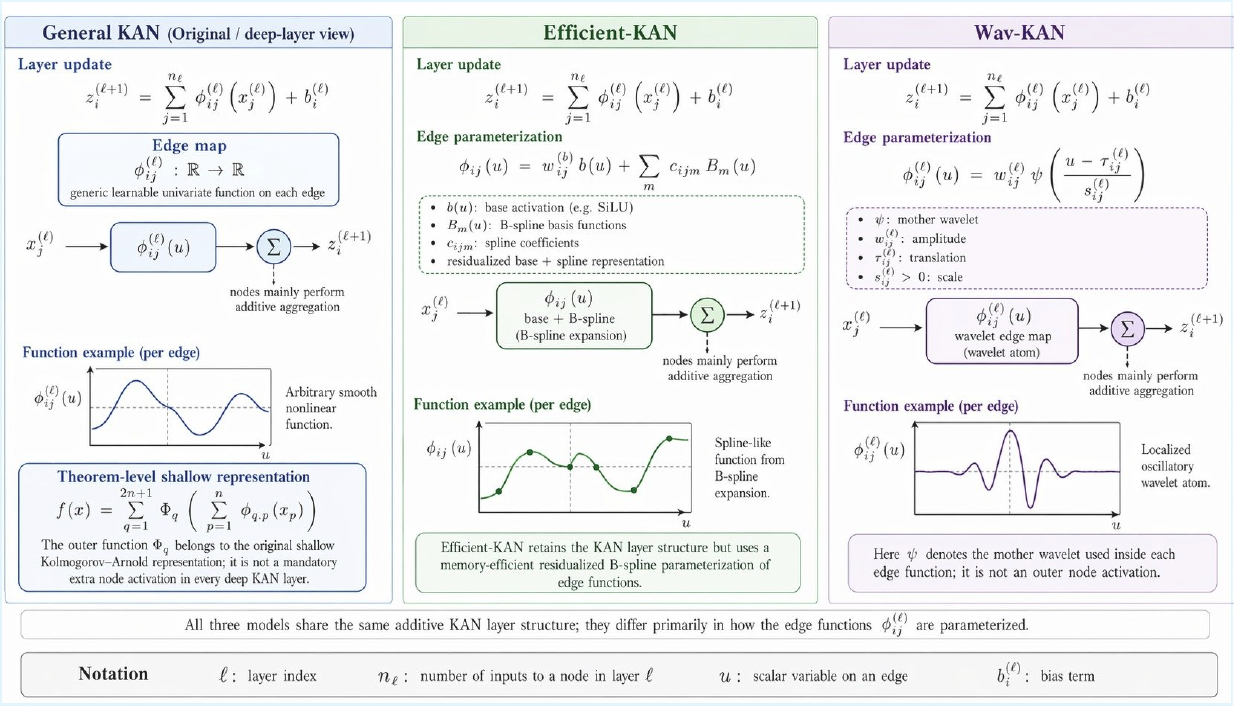}
    \caption{Comparison of the general KAN formulation and two important variants. In all cases, the layer update is $z_i^{(\ell+1)}=\sum_{j=1}^{n_\ell}\phi_{ij}^{(\ell)}\!\left(x_j^{(\ell)}\right)+b_i^{(\ell)}$. Efficient-KAN parameterizes the edge functions $\phi_{ij}^{(\ell)}$ by spline basis expansions, whereas Wav-KAN employs wavelet-based edge functions with explicit localization and scale.}
    \label{fig:ekan_wavkan}
\end{figure}

Recent years have seen the rapid development of several variants of the Kolmogorov--Arnold Network (KAN), including FastKAN, Efficient-KAN, Wav-KAN, T-KAN, and related architectures, each designed to address specific computational or application-driven requirements \cite{ PIKAN_JMLR_2025, Sen2026EhrenfestKAN, Somvanshi2025}. Among these, Efficient-KAN and Wav-KAN are especially relevant here. Efficient-KAN reformulates the original architecture using spline-based edge
functions (Fig.~\ref{fig:ekan_wavkan}) and improves computational efficiency and memory usage, making KAN more practical in large-scale settings \cite{Blealtan2024}. Wav-KAN, in contrast, replaces spline-based edge functions with wavelet-based ones (Fig.~\ref{fig:ekan_wavkan}), thereby introducing an explicitly localized and multiscale representation \cite{Bozorgasl2024}. Since Geometric KAN developed later in this work also relies on localized multiscale structure, Wav-KAN is particularly relevant to discuss in greater detail.

Wavelet-based KAN (Wav-KAN) modifies the original KAN construction by replacing spline-based learnable edge functions with wavelet-based ones. The motivation is that wavelets provide basis functions that are localized both in position and in scale, which is advantageous when the target mapping contains sharp transitions, oscillations, or features distributed across multiple resolutions.

The construction begins with a mother wavelet
$\psi:\mathbb{R}\to\mathbb{R}$, from which one generates translated and dilated wavelets
\begin{equation}
  \psi_{s,\tau}(u)
  \;=\;
  \frac{1}{\sqrt{s}}\,
  \psi\!\left(\frac{u-\tau}{s}\right),
  \qquad s>0,\ \tau\in\mathbb{R}.
  \label{eq:wav-family}
\end{equation}
Here, $s$ controls scale, with larger values capturing coarser behavior and smaller values resolving finer localized structure, while $\tau$ determines the position of the wavelet along the input axis. This is the fundamental idea underlying the continuous wavelet transform,
\begin{equation}
  C(s,\tau)
  \;=\;
  \int_{-\infty}^{+\infty}
  g(t)\,
  \frac{1}{\sqrt{s}}\,
  \psi\!\left(\frac{t-\tau}{s}\right)\,dt,
  \label{eq:cwt}
\end{equation}
which analyzes a signal across locations and resolutions.

Wav-KAN incorporates this multiscale structure directly into the KAN edge functions. Instead of expressing $\phi^{(\ell)}_{ij}$ through a spline basis, each edge is modeled by a learnable wavelet atom with trainable amplitude, translation, and scale:
\begin{equation}
  \phi^{(\ell)}_{ij}(u)
  \;=\;
  w^{(\ell)}_{ij}\,
  \psi\!\left(\frac{u-\tau^{(\ell)}_{ij}}{s^{(\ell)}_{ij}}\right),
  \qquad s^{(\ell)}_{ij}>0,
  \label{eq:wav-edge}
\end{equation}
where $w^{(\ell)}_{ij}$ is the amplitude coefficient, $\tau^{(\ell)}_{ij}$ determines the location of the wavelet, and $s^{(\ell)}_{ij}$ controls its effective support or bandwidth. Substituting Eq. \eqref{eq:wav-edge} into the KAN layer definition gives
\begin{equation}
  z^{(\ell+1)}_i
  \;=\;
  \sum_{j=1}^{n_\ell}
  w^{(\ell)}_{ij}\,
  \psi\!\left(\frac{x^{(\ell)}_j-\tau^{(\ell)}_{ij}}{s^{(\ell)}_{ij}}\right)
  \;+\;
  b^{(\ell)}_i,
  \qquad i=1,\ldots,m_\ell,
  \label{eq:wav-preact}
\end{equation}
and, if desired, an outer nonlinearity may be applied as
\begin{equation}
  a^{(\ell+1)}_i
  \;=\;
  \sigma^{(\ell)}\!\bigl(z^{(\ell+1)}_i\bigr),
  \qquad i=1,\ldots,m_\ell.
  \label{eq:wav-out}
\end{equation}

The importance of wavelets here goes beyond computational convenience. Because they encode locality and scale simultaneously, wavelets are well suited to problems in which relevant structure is concentrated near interfaces, singularities, sharp gradients, or other localized regions. In multiscale settings such as PDEs, dynamical systems, and geometrically structured data, they provide a natural way to separate coarse global behavior from fine local corrections. Compared with spline-based parameterizations tied to a fixed partition, wavelet-based edge functions can therefore adapt more naturally to irregular structure through their learned translations and dilations.

From this perspective, Wav-KAN is more than a computational variant of KAN; it provides a useful bridge toward geometry-aware learning. Its edge functions resolve features according to both location and scale, which is precisely the type of structure that later motivates Geometric KAN. While Wav-KAN enriches KAN through multiscale localized representations, Geometric KAN goes further by learning not only the representation but also the geometry of the input domain.

\section{Geometric KAN: Motivation, Input Geometry, Variants, and Mathematical Formulation}\label{GeoKAN_int}

Most existing deep learning models are defined on an input domain $\mathbb{R}^W$ endowed with the standard Euclidean geometry. This fixes both the metric and the effective resolution of the representation across the whole domain. In many applications, however, the target map is not uniformly regular. It may be smooth in some regions and vary sharply near interfaces, shocks, boundary layers, stiff transients, or localized singular structures. In such cases, fixed coordinates allocate model capacity inefficiently. Smooth regions are often over-resolved, while the regions that matter most remain under-resolved.

This limitation also appears in KAN-type models. Methods such as Efficient-KAN improve the flexibility of approximation by replacing scalar weights with learnable one-dimensional edge functions. However, these functions are still evaluated on the original coordinates. Thus, the representation becomes more expressive, but the underlying geometry of the input domain remains fixed. As a consequence, capturing sharp local variation typically requires either a finer basis grid or a larger model, even when the difficulty is highly localized.

GeoKAN is motivated by a different principle. The model should adapt not only the representation, namely the basis functions or feature map used to approximate the target, but also the geometry on which that representation is constructed. Here, geometry is understood in the differential-geometric sense, through a learned Riemannian metric on the input domain. The central idea is to learn a task-dependent metric that stretches regions of rapid variation and compresses smoother regions, so that basis evaluation is performed in geometry-adapted coordinates rather than in the original Euclidean coordinates.

\subsection{Mathematical Prerequisites}
\subsubsection{Prerequisite 1: Learned Metric and Geometric Quantities}

The central geometric object in GeoKAN is a learned diagonal Riemannian metric
\cite{Ryder2009}. For a representation
\(u=(u_1,\ldots,u_d)\in\mathbb{R}^d\), we write
\begin{equation}
g(u)=\mathrm{diag}\!\big(g_1(u),\dots,g_d(u)\big),
\qquad g_i(u)>0. \label{eq:metdiag}
\end{equation}
Here, \(g(u)\in\mathbb{R}^{d\times d}\) is a positive diagonal metric matrix
whose \(i\)-th diagonal entry is \(g_i(u)\). 

This metric defines the local inner product
\begin{equation}
\langle p,q\rangle_{g(u)} = p^\top g(u)\,q
= \sum_{i=1}^d g_i(u)\,p_i q_i,
\end{equation}

Here, the vectors
\(p=(p_1,\ldots,p_d)\in\mathbb{R}^d\) and
\(q=(q_1,\ldots,q_d)\in\mathbb{R}^d\) denote arbitrary tangent vectors, or
equivalently local perturbation directions, at the representation \(u\).

 And the corresponding infinitesimal length element
\begin{equation}
ds^2 = \sum_{i=1}^d g_i(u)\,(du_i)^2.
\end{equation}
Each component \(g_i(u)\) controls local stretching or compression along
direction \(u_i\). Large values of \(g_i(u)\) increase geometric distance along
that coordinate and therefore increase effective representational resolution in
that region.

A second useful quantity is the volume distortion. Since the metric is diagonal,
\begin{equation}
\det g(u)=\prod_{i=1}^d g_i(u),
\qquad
\log \det g(u)=\sum_{i=1}^d \log g_i(u).
\end{equation}
This scalar summarizes how the learned geometry expands or contracts local volume. In several GeoKAN variants, $\log\det g(u)$ is appended as an additional feature.

\subsubsection{Prerequisite 2: Geometry-Adapted Feature Construction}

\begin{figure}
    \centering
    \includegraphics[width=\textwidth]{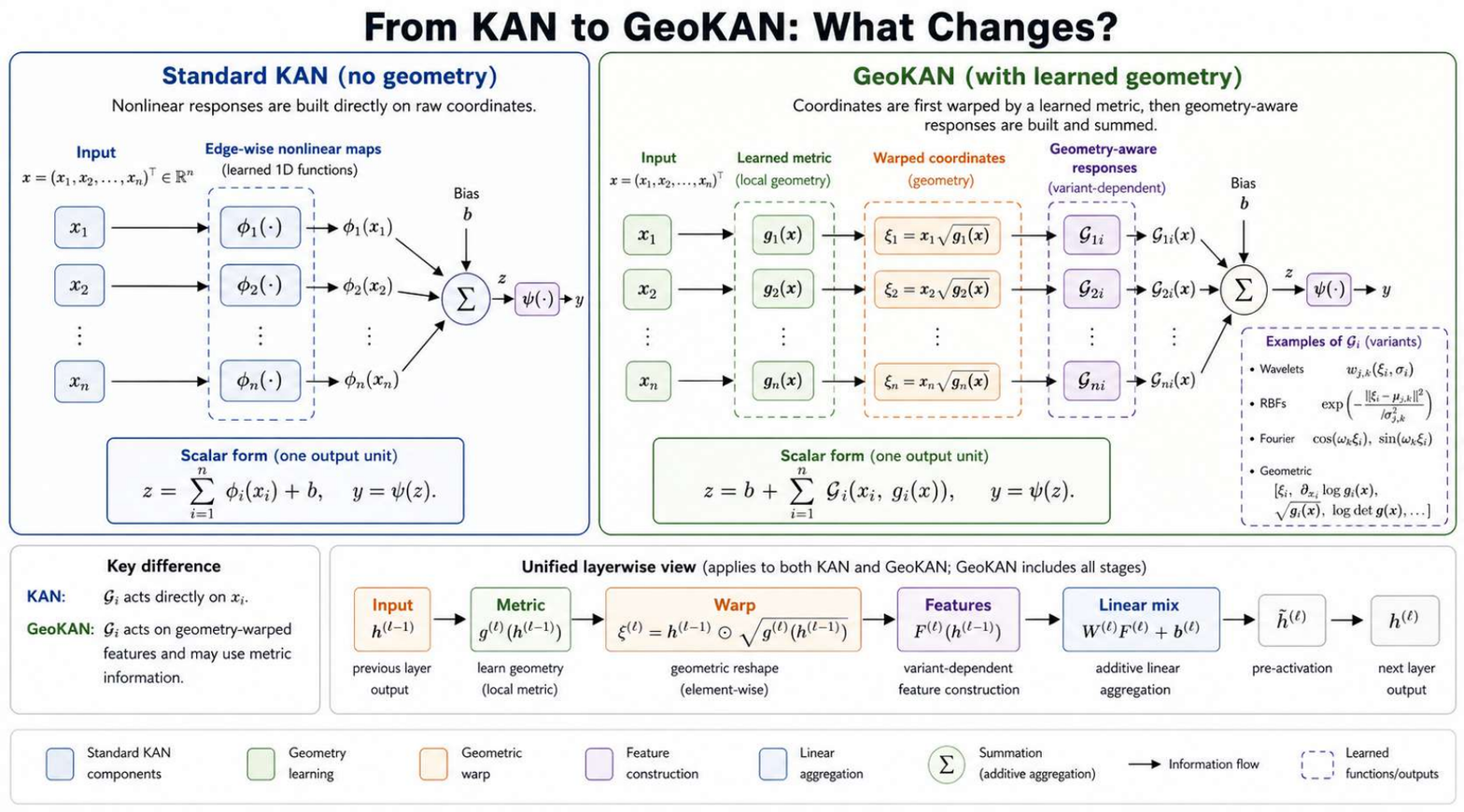}
    \caption{Comparison between a standard KAN and GeoKAN at the level of feature construction. In a standard KAN, each input coordinate is passed directly through a learnable univariate edge function, and the resulting responses are additively combined, schematically \( z=\sum_i \phi_i(x_i)+b \). Thus, the representation is learned directly in the original coordinate system. In GeoKAN, the same additive principle is retained, but feature construction is preceded by a learned geometric warp: a positive metric \( g_i(x) \) first reshapes each coordinate through \( z_i=x_i\sqrt{g_i(x)} \), after which geometry-aware responses are constructed from the warped representation. These responses may be realized by localized basis functions, such as wavelet, RBF, or Fourier, or by explicit metric-derived quantities. The figure therefore highlights the central distinction: KAN learns nonlinear responses in the original coordinates, whereas GeoKAN first learns the geometry of the representation space and then builds nonlinear responses within that learned geometry.}
    \label{fig:kan_geokan}
\end{figure}

The main difference between GeoKAN and standard KAN-type models is that feature evaluation is performed after a learned geometric warp (Fig.~\ref{fig:kan_geokan}). For a coordinate $u_i$, define the warped variable
\begin{equation}
z_i(u)=u_i\sqrt{g_i(u)}, \qquad i=1,\dots,d. \label{eq:metwarp}
\end{equation}
A canonical geometry-adapted building block has the form
\begin{equation}
\varphi_{i,k}(u)
=
\psi\!\left(\frac{z_i(u)-c_{i,k}}{s_{i,k}}\right)
=
\psi\!\left(\frac{u_i\sqrt{g_i(u)}-c_{i,k}}{s_{i,k}}\right),
\end{equation}
where $\psi$ is a chosen basis function, $c_{i,k}$ is a learnable center, and $s_{i,k}>0$ is a learnable scale.

At a general level, the GeoKAN family is determined by three choices (see Fig.~\ref{fig:kan_geokan}):
\begin{equation}
\text{(i) the metric model } g,
\qquad
\text{(ii) the feature dictionary } F,
\qquad
\text{(iii) any additional geometric features.}
\end{equation}

A central component of this construction is the choice of basis function $\psi$ used in the post-warp feature dictionary. Since different basis families encode different approximation biases, the basis choice strongly influences which kinds of structures are represented most naturally after geometric warping. Before introducing the unified layerwise formulation and the specific GeoKAN variants, we briefly summarize the basis families used in this work.

\subsubsection{Prerequisite 3: Basis Functions to Be Used in GeoKAN Variants}

In GeoKAN-type models, the nonlinear representation is constructed from a chosen family of basis functions applied after the learned geometric warp. The basis determines the functional language in which the model expresses its nonlinear map, and therefore provides an important inductive bias. In the variants considered in this work, the principal basis families are the wavelet basis, the radial basis function (RBF) basis, and the Fourier basis. These bases enter the model after the input or hidden representation has been transformed into geometry-adapted coordinates.

\paragraph{Wavelet basis.}
A wavelet basis represents a function using shifted and scaled copies of a mother wavelet \(\psi\):
\begin{equation}
f(x)=\sum_{j,k} c_{j,k}\,\psi_{j,k}(x),
\qquad
\psi_{j,k}(x)=2^{j/2}\psi(2^j x-k).
\end{equation}
Here, \(j\) denotes the scale, \(k\) denotes the location, and \(c_{j,k}\) are coefficients. In contrast to global trigonometric modes, wavelets are localized in both position and scale. This makes them well suited for representing localized structure, sharp transitions, singular behavior, and multi-resolution features.

Within GeoKAN, wavelet basis are evaluated in the warped coordinates \(z_i(u)=u_i\sqrt{g_i(u)}\) rather than in the original Euclidean coordinates. A typical geometry-adapted wavelet atom therefore has the form
\begin{equation}
\phi_{i,k}(u)
=
\psi\!\left(
\frac{z_i(u)-c_{i,k}}{s_{i,k}}
\right),
\end{equation}
where \(c_{i,k}\) and \(s_{i,k}\) are learnable centers and scales. In this way, local multi-scale approximation is combined with the learned geometric stretching or compression of the input domain.

\paragraph{Radial basis function.}
An RBF expansion represents a function as
\begin{equation}
f(x)=\sum_{m=1}^{M} c_m\,\phi\!\left(\frac{\|x-\mu_m\|}{\sigma_m}\right),
\end{equation}
where \(\mu_m\) is the center of the \(m\)-th basis function, \(\sigma_m\) determines its width, and \(c_m\) is a coefficient. A common choice is the Gaussian RBF,
\begin{equation}
\phi(r)=e^{-\gamma r^2},
\end{equation}
which yields
\begin{equation}
f(x)=\sum_{m=1}^{M} c_m\,e^{-\gamma \|x-\mu_m\|^2}.
\end{equation}

RBFs are localized smooth bump functions centered around specific points, and they are especially effective for smooth local interpolation and approximation. In GeoKAN, these basis functions are likewise evaluated in geometry-adapted coordinates. A typical warped RBF atom takes the form
\begin{equation}
\phi_{i,k}(u)
=
\exp\!\left(
-\gamma
\left(
\frac{z_i(u)-c_{i,k}}{s_{i,k}}
\right)^2
\right).
\end{equation}
Accordingly, the locality of the approximation is controlled jointly by the learned geometry and by the atom parameters.

\paragraph{Fourier basis.}
A Fourier basis represents a function as a linear combination of sinusoidal modes:
\begin{equation}
f(x)=a_0+\sum_{k=1}^{K}\left[a_k\cos(k\omega x)+b_k\sin(k\omega x)\right],
\end{equation}
where \(a_k\) and \(b_k\) are coefficients, \(\omega\) is a base frequency, and \(K\) determines the number of harmonics retained in the expansion. This basis is particularly natural for oscillatory or approximately periodic phenomena, since sine and cosine functions already encode wave-like structure.

In the GeoKAN framework, Fourier features may also be evaluated after geometric warping. A geometry-adapted Fourier atom may be written as
\begin{equation}
\phi_{i,k}^{\cos}(u)=\cos\!\big(k\omega z_i(u)\big),
\qquad
\phi_{i,k}^{\sin}(u)=\sin\!\big(k\omega z_i(u)\big),
\end{equation}
where again \(z_i(u)=u_i\sqrt{g_i(u)}\). This allows oscillatory structure to be represented in coordinates that have already been adapted to the local geometry of the target map. Such a construction is particularly appropriate when the underlying problem exhibits strong wave-like behavior, as in the Helmholtz setting considered later in this work.

\paragraph{Basis choice within the GeoKAN variants.}
The preceding basis families enter the GeoKAN variants in different ways depending on how the post-warp representation is constructed. In variants based on localized dictionaries, the geometry-adapted coordinates may be expanded using wavelet, Gaussian RBF, or, for strongly oscillatory problems, Fourier basis function. In more structured variants, the model may instead bypass an explicit post-warp basis expansion and construct features directly from the learned metric and its derivatives. This distinction in feature construction is one of the main mechanisms through which the GeoKAN variants differ.

\subsection{GeoKAN}

In a standard KAN, one may first describe the computation at the level of a
single output unit in a hidden layer (see Fig \ref{fig:kan_geokan}). Let
\[
x=(x_1,\dots,x_n)
\]
denote the vector received by this unit from the previous layer. Each coordinate
\(x_i\) is passed through a learnable univariate edge function \(\phi_i\), and
the resulting scalar responses are additively combined to produce the unit's
pre-activation output,
\[
z=\sum_{i=1}^{n}\phi_i(x_i)+b.
\]
The quantity \(z\) therefore represents the scalar output of one hidden-layer
unit before the outer nonlinearity is applied, rather than the output of the
entire layer. Applying the node function gives
\[
y=\psi(z).
\]
Hence, the local computation of a KAN unit can be viewed as first extracting
coordinate-wise nonlinear responses and then aggregating them into a single
hidden-unit response.
Here, the learnable nonlinear responses are applied directly in the original input space. Each coordinate contributes through its own edge-wise function \(\phi_i\), and the node or unit then combines these responses by summation. In this sense, KAN learns which nonlinear one-dimensional responses should be extracted from the raw coordinates and then aggregates them.

GeoKAN can be introduced most clearly by first looking at the computation
performed by a single hidden-layer unit. Suppose this unit receives the same
input vector as in KAN from the previous layer,
GeoKAN keeps the one-unit additive structure of standard KAN, but changes the nature of the coordinate-wise response. Instead of evaluating the nonlinear response directly on the raw coordinate \(x_i\), GeoKAN first introduces a learned positive metric \(g_i(x)\)~\eqref{eq:metdiag}, which locally reshapes the input space. Similar to Eq. \eqref{eq:metwarp}, a simple metric-warped coordinate is
\[
\xi_i=x_i\sqrt{g_i(x)} .
\]
The contribution of the \(i\)-th coordinate is then evaluated in this warped
geometry. Therefore, for one hidden-layer unit, the GeoKAN response can be
written schematically as
\[
z
=
b+\sum_{i=1}^{n}
\mathcal{G}_i\!\left(x_i,g_i(x)\right),
\]
where \(\mathcal{G}_i\) denotes a geometry-aware nonlinear contribution
associated with the \(i\)-th input coordinate. In the simplest case, this may
take the form
\[
\mathcal{G}_i\!\left(x_i,g_i(x)\right)
=
\phi_i\!\left(\xi_i\right)
=
\phi_i\!\left(x_i\sqrt{g_i(x)}\right).
\]
so that the ordinary KAN response \(\phi_i(x_i)\) is replaced by a response
evaluated after a learned metric warp.

More flexible GeoKAN variants may enrich \(\mathcal{G}_i\) further by using
localized basis functions or additional geometric quantities. For example,
\(\mathcal{G}_i\) may be constructed from wavelet, Gaussian RBF,
Fourier basis, or explicit metric-dependent features such as
\[
x_i\sqrt{g_i(x)}, \qquad
\partial_{x_i}\log g_i(x), \qquad
\sqrt{g_i(x)}, \qquad
\log\det g(x).
\]
The essential idea is that the nonlinear response is no longer learned only in
the original coordinate system; it is learned after the input has been locally
reshaped by a data-dependent geometry.

For a full hidden layer with multiple output units, the same idea is applied to
each unit. If \(j\) indexes the output unit, then a GeoKAN layer can be written as
\[
z_j
=
b_j+
\sum_{i=1}^{n}
\mathcal{G}_{ji}\!\left(x_i,g_i(x)\right),
\qquad
j=1,\ldots,m .
\]
Here, \(\mathcal{G}_{ji}\) is the geometry-aware response connecting input
coordinate \(i\) to output unit \(j\). In vector form, the layer maps
\[
x\longmapsto z=(z_1,\ldots,z_m).
\]

In the implemented GeoKAN/LM-KAN models used here, this geometry-aware layer
output is followed by a node-wise nonlinearity between hidden layers:
\[
y_j=\psi(z_j),
\qquad
\psi=\tanh .
\]
This final nonlinearity should be understood as an implementation-level
hybridization. The standard KAN formulation keeps the node as a pure summation
operator, whereas the present GeoKAN/LM-KAN implementation combines
KAN-style geometry-aware responses with an MLP-style outer activation.

This direct scalar viewpoint also makes it easier to understand the full
layerwise construction. Let
\begin{equation}
h^{(0)}=x\in\mathbb{R}^{d_0},
\end{equation}
and let \(h^{(\ell-1)}\in\mathbb{R}^{d_{\ell-1}}\) denote the input to layer
\(\ell\). A generic GeoKAN layer first learns a positive diagonal metric over
the incoming representation,
\begin{equation}
g^{(\ell)}\!\left(h^{(\ell-1)}\right)
=
\mathrm{diag}\!\Big(
g_1^{(\ell)},\ldots,g_{d_{\ell-1}}^{(\ell)}
\Big),
\qquad
g_i^{(\ell)}>0 .
\end{equation}
This metric defines a locally warped representation
\begin{equation}
\xi^{(\ell)}
=
h^{(\ell-1)}
\odot
\sqrt{
g^{(\ell)}\!\left(h^{(\ell-1)}\right)
}.
\end{equation}
Here, \(\odot\) denotes element-wise multiplication, so the \(i\)-th warped coordinate is
$
\xi_i^{(\ell)}
=
h_i^{(\ell-1)}
\sqrt{
g_i^{(\ell)}\!\left(h^{(\ell-1)}\right)
}.
$
The layer then constructs a variant-dependent geometry-aware feature vector
\begin{equation}
F^{(\ell)}\!\left(h^{(\ell-1)}\right)\in\mathbb{R}^{M_\ell},
\end{equation}
whose entries may depend on the warped coordinates \(\xi^{(\ell)}\), the metric
values \(g^{(\ell)}\), and, in some variants, additional metric-derived
quantities. The pre-activation output of the layer is then computed as
\begin{equation}
\tilde h^{(\ell)}
=
W^{(\ell)}
F^{(\ell)}\!\left(h^{(\ell-1)}\right)
+
b^{(\ell)} .
\label{eq:generic_geokan_layer}
\end{equation}

Thus, the scalar quantity \(z\) introduced for one hidden unit corresponds,
in the full layer notation, to one component of the pre-activation vector:
\[
z=\tilde h^{(\ell)}_j .
\]

For hidden layers in the implemented GeoKAN/LM-KAN models, this pre-activation
is followed by a pointwise nonlinearity, for example
\begin{equation}
h^{(\ell)}
=
\psi\!\left(\tilde h^{(\ell)}\right),
\qquad
\psi=\tanh ,
\qquad
\ell=1,\ldots,L .
\label{eq:generic_geokan_hidden}
\end{equation}
Here, each component \(\tilde h^{(\ell)}_j\) is precisely the layerwise analogue of the scalar unit output \(z\); that is, for the \(j\)-th hidden unit, \(z=\tilde h^{(\ell)}_j\).
A final linear readout then produces the model prediction,
\begin{equation}
\hat y(x)
=
W_{\mathrm{out}}h^{(L)}
+
b_{\mathrm{out}} .
\label{eq:generic_geokan_readout}
\end{equation}
Thus, the scalar intuition and the layerwise formalism are consistent. At the
scalar level GeoKAN forms geometry-aware coordinate responses and combines them,
while at the layer level it builds a geometry-aware feature vector and linearly
mixes those features to produce the next representation.

\subsection{GeoKAN Variants}
The main GeoKAN variants arise from three natural modeling choices. The first
is the parameterization of the metric. One may learn the metric from the full
incoming representation through a coupled model such as
\begin{equation}
g_i(u)
=
\mathrm{softplus}\!\big(\mathrm{MetricNet}_i(u)\big)
+
\varepsilon,
\end{equation}
where \(\mathrm{MetricNet}_i\) denotes the \(i\)-th scalar output of a trainable
metric network
\[
\mathrm{MetricNet}: \mathbb{R}^{d_{\mathrm{in}}} \to \mathbb{R}^{d_{\mathrm{in}}},
\]
which maps the full incoming representation \(u=(u_1,\ldots,u_{d_{\mathrm{in}}})\)
to unconstrained metric logits. The softplus transformation, together with the
small offset \(\varepsilon>0\), ensures that each diagonal metric component
\(g_i(u)\) is strictly positive. This coupled parameterization allows the local
scaling of one coordinate to depend on the others.

Alternatively, one may use a more structured separable form,
\begin{equation}
g_i(u_i)
=
\exp\!\left(
\sum_{k=1}^{K} c_{i,k}\,\rho_k(u_i)
\right),
\end{equation}
where \(\rho_k\) denotes a localized basis function. This separable form is
lighter, more interpretable, and less expensive.

The second choice is the construction of post-warp features. After the geometry
has transformed \(u\) into warped coordinates
\begin{equation}
\xi_i(u)
=
u_i\sqrt{g_i(u)},
\end{equation}
the model may build features using wavelet, Gaussian RBF, Fourier basis functions, or, in the most structured case, explicit geometric quantities derived
directly from the metric itself. 

The third choice is whether to append
additional geometric summaries, such as the volume term \(\log\det g(u)\), or
derivative-based quantities such as
\begin{equation}
\gamma_i(u_i)
=
\frac{1}{2}\partial_{u_i}\log g_i(u_i).
\end{equation}
These choices determine the major GeoKAN variants in a systematic way rather
than in an ad hoc manner.

From this perspective, GeoKAN-NNMetric corresponds to the case in which the metric is learned from the full representation and the post-warp features are
constructed using localized wavelet basis, typically together with a volume term. GeoKAN-\(\gamma\) corresponds to the more structured setting in which the
metric is separable and the features are built explicitly from metric-derived geometric quantities such as the warped coordinate, the local logarithmic
derivative of the metric, and the metric magnitude itself. LM-KAN again uses a coupled learned metric, but allows the post-warp feature dictionary to be chosen
according to the target problem, for example using wavelet, RBF, or Fourier
basis functions. In this way, all variants remain instances of the same general GeoKAN
principle: learn a geometry, construct nonlinear features within that learned
geometry, and then combine them to produce the output.

Accordingly, the distinction between KAN and GeoKAN can be summarized in a
simple and precise way. KAN may be described as learning nonlinear edge
functions and summing them, whereas GeoKAN first learns the local geometry,
constructs nonlinear responses in that learned geometry, and then sums or mixes
those responses. A compact generic scalar template for one GeoKAN unit is
therefore
\begin{equation}
z
=
b+\sum_{i=1}^{n}
\mathcal{G}_i\!\left(
x_i\sqrt{g_i(x)},\,
g_i(x),\,
\partial_{x_i}\log g_i(x),\,
\ldots
\right)
\end{equation}
where the precise arguments of \(\Phi_i\) depend on the chosen variant. In the
implemented hidden layers used here, this scalar pre-activation may then be
passed through an additional pointwise nonlinearity,
\begin{equation}
y=\psi(z),
\qquad
\psi=\tanh .
\end{equation}
This expression captures the central idea of the framework: GeoKAN preserves
the additive response structure of KAN, but extends it by making the response
itself geometry-aware through a learned metric warp.

In one sentence, the direct-scalar interpretation of GeoKAN is this: instead of
summing learnable nonlinear functions of the raw inputs, GeoKAN sums learnable
or structured nonlinear functions of geometrically warped inputs and related
metric features.







\subsection{GeoKAN-NNMetric}

GeoKAN-NNMetric is a flexible coupled-metric variant of the GeoKAN framework.
Within the generic GeoKAN template introduced above, it corresponds to the
choice of a learned metric that depends on the full incoming representation,
followed by a post-warp wavelet feature dictionary and an additional volume
summary term. In this way, the model first learns a data-adaptive local geometry
of the representation space and then constructs localized nonlinear responses
within that learned geometry. Because the metric depends on the full input to
the layer, the local scaling of one coordinate may depend on the others, allowing
the model to capture coupled nonlinear structure that may not be well represented
by more separable alternatives. The resulting warped representation is then
expanded using Mexican-hat wavelet basis, making GeoKAN-NNMetric a natural
variant when one seeks a flexible geometry-aware model with localized multiscale
features.

Let \(h^{(\ell-1)}\in\mathbb{R}^{d_{\ell-1}}\) denote the input to layer
\(\ell\). GeoKAN-NNMetric learns a positive diagonal metric whose \(i\)-th
component is given by
\begin{equation}
g_i^{(\ell)}\!\left(h^{(\ell-1)}\right)
=
\mathrm{softplus}\!\left(
\mathrm{MetricNet}^{(\ell)}_i\!\left(h^{(\ell-1)}\right)
\right)
+
\varepsilon,
\qquad
i=1,\dots,d_{\ell-1}.
\end{equation}
Here, the metric component \(g_i^{(\ell)}\) is a function of the full incoming
representation \(h^{(\ell-1)}\), not only of the coordinate \(h_i^{(\ell-1)}\).
The corresponding warped coordinate is written consistently as
\begin{equation}
\xi_i^{(\ell)}
=
h_i^{(\ell-1)}
\sqrt{
g_i^{(\ell)}\!\left(h^{(\ell-1)}\right)
}.
\end{equation}

For each coordinate \(i\) and wavelet basis \(k=1,\dots,K\), define the normalized
post-warp coordinate
\begin{equation}
r_{i,k}^{(\ell)}
=
\frac{
\xi_i^{(\ell)}-c_{i,k}^{(\ell)}
}{
s_{i,k}^{(\ell)}
},
\qquad
s_{i,k}^{(\ell)}>0,
\end{equation}
where \(c_{i,k}^{(\ell)}\) and \(s_{i,k}^{(\ell)}\) are the center and scale of
the \(k\)-th basis associated with coordinate \(i\). GeoKAN-NNMetric uses the
Mexican-hat wavelet
\begin{equation}
\psi_{\mathrm{wav}}(r)
=
(1-r^2)\exp\!\left(-\frac{r^2}{2}\right),
\end{equation}
and hence the corresponding wavelet feature is
\begin{equation}
\eta_{i,k}^{(\ell)}
=
\psi_{\mathrm{wav}}\!\left(r_{i,k}^{(\ell)}\right).
\end{equation}
We use \(\eta_{i,k}^{(\ell)}\) for these dictionary features in order to avoid
confusing them with the standard KAN edge functions \(\phi_i\).

The layer also includes the volume feature
\begin{equation}
v^{(\ell)}
=
\sum_{i=1}^{d_{\ell-1}}
\log
g_i^{(\ell)}\!\left(h^{(\ell-1)}\right)
=
\log\det
g^{(\ell)}\!\left(h^{(\ell-1)}\right),
\end{equation}
where the last equality follows because the learned metric is diagonal. The
GeoKAN-NNMetric feature vector is therefore
\begin{equation}
F_{\mathrm{NNMetric}}^{(\ell)}
=
\Big[
\eta_{1,1}^{(\ell)},\dots,\eta_{d_{\ell-1},K}^{(\ell)},v^{(\ell)}
\Big]^\top .
\end{equation}

The layerwise pre-activation output is computed by linearly mixing these
geometry-aware features:
\begin{equation}
\tilde h_{\mathrm{NNMetric}}^{(\ell)}
=
W_{\mathrm{NNMetric}}^{(\ell)}
F_{\mathrm{NNMetric}}^{(\ell)}
+
b_{\mathrm{NNMetric}}^{(\ell)}.
\label{eq:nnmetric_forward_layer_rewrite}
\end{equation}
As in the generic GeoKAN notation, each component of
\(\tilde h_{\mathrm{NNMetric}}^{(\ell)}\) corresponds to the scalar output
\(z\) of one hidden-layer unit.

Equivalently, for output channel \(j=1,\dots,d_\ell\), the expanded scalar form
is
\begin{equation}
\tilde h_{\mathrm{NNMetric},j}^{(\ell)}
=
b_{\mathrm{NNMetric},j}^{(\ell)}
+
\sum_{i=1}^{d_{\ell-1}}\sum_{k=1}^{K}
a_{j,i,k}^{(\ell)}
\psi_{\mathrm{wav}}\!\left(
\frac{
h_i^{(\ell-1)}
\sqrt{g_i^{(\ell)}(h^{(\ell-1)})}
-
c_{i,k}^{(\ell)}
}{
s_{i,k}^{(\ell)}
}
\right)
+
a_{j,\mathrm{vol}}^{(\ell)}
\sum_{i=1}^{d_{\ell-1}}
\log
g_i^{(\ell)}\!\left(h^{(\ell-1)}\right).
\label{eq:nnmetric_forward_expanded_rewrite}
\end{equation}
Thus, GeoKAN-NNMetric realizes the general GeoKAN principle by learning a
coupled positive metric, warping each coordinate by that learned geometry,
building localized wavelet responses in the warped coordinates, appending a
volume summary, and finally mixing these features to produce the next-layer
pre-activation.

\begin{figure}
    \centering
    \includegraphics[width=\textwidth]{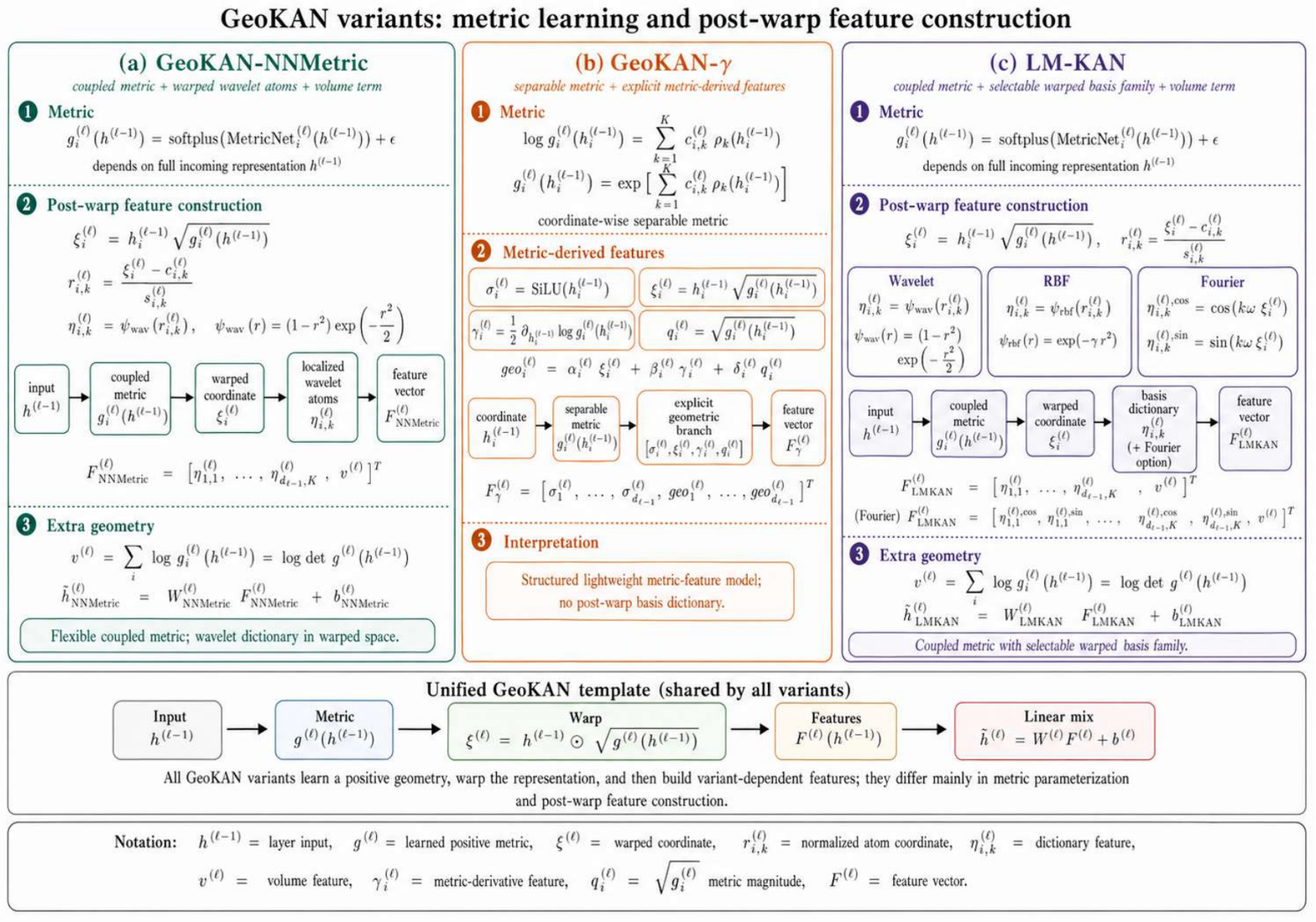}
    \caption{Schematic comparison of the main GeoKAN variants through their metric and post-warp feature construction. GeoKAN-NNMetric uses a coupled learned metric \( g_i(h) = \mathrm{softplus}(\mathrm{MetricNet}_i(h)) + \varepsilon \) together with warped wavelet features and a volume term \( \sum_i \log g_i(h) \). GeoKAN-$\gamma$ uses a separable metric \( g_i(h_i) \) and replaces a warped basis dictionary by explicit metric-derived features such as \( h_i \sqrt{g_i} \), \( \gamma_i = \frac{1}{2} \partial_{h_i} \log g_i \), and \( \sqrt{g_i} \). LM-KAN again uses a coupled learned metric, but builds the post-warp dictionary from wavelet, RBF, or Fourier basis, making it the most flexible basis-driven variant for PDE-oriented settings.}
    \label{fig:geokan_variants}
\end{figure}

\subsection{GeoKAN-$\gamma$}

GeoKAN-\(\gamma\) is the lightest and most structured variant in the GeoKAN
family. It uses a separable metric, so each coordinate learns its own local
scaling independently. Rather than applying a learned wavelet or RBF dictionary
after the warp, it constructs explicit geometric features from the metric and
its local variation (Fig.~\ref{fig:geokan_variants}b). This gives the model a
stronger inductive bias, makes the geometric role of each feature easier to
interpret, and keeps the architecture relatively compact.

The metric is defined by
\begin{equation}
\log g_i^{(\ell)}\!\left(h_i^{(\ell-1)}\right)
=
\sum_{k=1}^{K}
c_{i,k}^{(\ell)}
\rho_k\!\left(h_i^{(\ell-1)}\right),
\end{equation}
and hence
\begin{equation}
g_i^{(\ell)}\!\left(h_i^{(\ell-1)}\right)
=
\exp\!\left[
\sum_{k=1}^{K}
c_{i,k}^{(\ell)}
\rho_k\!\left(h_i^{(\ell-1)}\right)
\right].
\end{equation}
In the implementation, the basis functions \(\rho_k\) are Gaussian RBFs on a
one-dimensional grid.

This variant does not use a post-warp wavelet or RBF dictionary. Instead, it
constructs explicit geometric features:
\begin{align}
\sigma_i^{(\ell)}
&=
\mathrm{SiLU}\!\left(h_i^{(\ell-1)}\right),
\\
\xi_i^{(\ell)}
&=
h_i^{(\ell-1)}
\sqrt{
g_i^{(\ell)}\!\left(h_i^{(\ell-1)}\right)
},
\\
\gamma_i^{(\ell)}
&=
\frac{1}{2}
\partial_{h_i^{(\ell-1)}}
\log g_i^{(\ell)}\!\left(h_i^{(\ell-1)}\right),
\\
q_i^{(\ell)}
&=
\sqrt{
g_i^{(\ell)}\!\left(h_i^{(\ell-1)}\right)
}.
\end{align}
Here, \(\xi_i^{(\ell)}\) is the warped coordinate, \(\gamma_i^{(\ell)}\) measures
the local logarithmic variation of the metric, and \(q_i^{(\ell)}\) represents
the metric magnitude through its square root.

These metric-derived quantities are combined into a geometric branch
\begin{equation}
\mathrm{geo}_i^{(\ell)}
=
\alpha_i^{(\ell)} \xi_i^{(\ell)}
+
\beta_i^{(\ell)} \gamma_i^{(\ell)}
+
\delta_i^{(\ell)} q_i^{(\ell)} .
\end{equation}
The feature vector is
\begin{equation}
F_{\gamma}^{(\ell)}
=
\Big[
\sigma_1^{(\ell)},\dots,\sigma_{d_{\ell-1}}^{(\ell)},
\mathrm{geo}_1^{(\ell)},\dots,\mathrm{geo}_{d_{\ell-1}}^{(\ell)}
\Big]^\top .
\end{equation}
The layerwise pre-activation output is
\begin{equation}
\tilde h_{\gamma}^{(\ell)}
=
W_{\gamma}^{(\ell)}F_{\gamma}^{(\ell)}
+
b_{\gamma}^{(\ell)} .
\label{eq:gamma_forward_layer_rewrite}
\end{equation}
As in the generic GeoKAN notation, each component of
\(\tilde h_{\gamma}^{(\ell)}\) corresponds to the scalar output \(z\) of one
hidden-layer unit.

Equivalently, for output channel \(j=1,\dots,d_\ell\), the expanded scalar form
is
\begin{equation}
\tilde h_{\gamma,j}^{(\ell)}
=
b_{\gamma,j}^{(\ell)}
+
\sum_{i=1}^{d_{\ell-1}}
u_{j,i}^{(\ell)}
\,\mathrm{SiLU}\!\left(h_i^{(\ell-1)}\right)
+
\sum_{i=1}^{d_{\ell-1}}
v_{j,i}^{(\ell)}
\Big[
\alpha_i^{(\ell)}
\xi_i^{(\ell)}
+
\beta_i^{(\ell)}
\gamma_i^{(\ell)}
+
\delta_i^{(\ell)}
q_i^{(\ell)}
\Big].
\label{eq:gamma_forward_expanded_rewrite}
\end{equation}
Substituting the definitions of \(\xi_i^{(\ell)}\) and \(q_i^{(\ell)}\), this
can also be written as
\begin{equation}
\tilde h_{\gamma,j}^{(\ell)}
=
b_{\gamma,j}^{(\ell)}
+
\sum_{i=1}^{d_{\ell-1}}
u_{j,i}^{(\ell)}
\,\mathrm{SiLU}\!\left(h_i^{(\ell-1)}\right)
+
\sum_{i=1}^{d_{\ell-1}}
v_{j,i}^{(\ell)}
\Big[
\alpha_i^{(\ell)}
h_i^{(\ell-1)}
\sqrt{g_i^{(\ell)}\!\left(h_i^{(\ell-1)}\right)}
+
\beta_i^{(\ell)}
\gamma_i^{(\ell)}
+
\delta_i^{(\ell)}
\sqrt{g_i^{(\ell)}\!\left(h_i^{(\ell-1)}\right)}
\Big].
\end{equation}
Thus, GeoKAN-\(\gamma\) is best viewed as a metric-feature model rather than a
warped basis model.

\subsection{LM-KAN}
LM-KAN is a PDE-oriented branch of the same geometric framework. Like
GeoKAN-NNMetric, it learns a coupled metric from the full incoming
representation. In terms of feature construction, it employs a localized
post-warp dictionary, whose basis may be chosen according to the structure of
the target problem, including wavelet, Gaussian RBF, or Fourier features
(Fig.~\ref{fig:geokan_variants}c).

Let \(h^{(\ell-1)}\in\mathbb{R}^{d_{\ell-1}}\) denote the input to layer
\(\ell\). LM-KAN learns a coupled positive metric from the full incoming
representation,
\begin{equation}
g_i^{(\ell)}\!\left(h^{(\ell-1)}\right)
=
\mathrm{softplus}\!\left(
\mathrm{MetricNet}^{(\ell)}_i\!\left(h^{(\ell-1)}\right)
\right)
+
\varepsilon,
\qquad
i=1,\dots,d_{\ell-1}.
\end{equation}
Thus, the metric component \(g_i^{(\ell)}\) may depend on all coordinates of
\(h^{(\ell-1)}\), not only on \(h_i^{(\ell-1)}\).

The warped coordinate is written consistently as
\begin{equation}
\xi_i^{(\ell)}
=
h_i^{(\ell-1)}
\sqrt{
g_i^{(\ell)}\!\left(h^{(\ell-1)}\right)
}.
\end{equation}
For localized wavelet or RBF basis, we define the normalized post-warp basis
coordinate
\begin{equation}
r_{i,k}^{(\ell)}
=
\frac{
\xi_i^{(\ell)}-c_{i,k}^{(\ell)}
}{
s_{i,k}^{(\ell)}
},
\qquad
s_{i,k}^{(\ell)}>0,
\qquad
k=1,\dots,K.
\end{equation}

LM-KAN admits multiple basis choices. The wavelet version uses the Mexican-hat
wavelet
\begin{equation}
\psi_{\mathrm{wav}}(r)
=
(1-r^2)\exp\!\left(-\frac{r^2}{2}\right),
\end{equation}
while the RBF version uses the Gaussian radial basis function
\begin{equation}
\psi_{\mathrm{rbf}}(r)
=
\exp(-\gamma r^2),
\qquad
\gamma>0.
\end{equation}
For oscillatory problems, LM-KAN may also employ a Fourier basis directly in
the warped coordinate,
\begin{equation}
\psi_{\mathrm{four},k}^{\cos}\!\left(\xi_i^{(\ell)}\right)
=
\cos\!\left(k\omega \xi_i^{(\ell)}\right),
\qquad
\psi_{\mathrm{four},k}^{\sin}\!\left(\xi_i^{(\ell)}\right)
=
\sin\!\left(k\omega \xi_i^{(\ell)}\right),
\end{equation}
where \(\omega\) is a base frequency and \(k\) indexes the retained harmonics.

For the wavelet and RBF cases, the dictionary atoms are written uniformly as
\begin{equation}
\eta_{i,k}^{(\ell)}
=
\psi\!\left(r_{i,k}^{(\ell)}\right),
\qquad
\psi\in\{\psi_{\mathrm{wav}},\psi_{\mathrm{rbf}}\}.
\end{equation}
We use \(\eta_{i,k}^{(\ell)}\) for these dictionary features to avoid confusing
them with the standard KAN edge functions \(\phi_i\). For the Fourier case, the
atoms are defined directly in terms of the warped coordinate:
\begin{equation}
\eta_{i,k}^{(\ell),\cos}
=
\cos\!\left(k\omega \xi_i^{(\ell)}\right),
\qquad
\eta_{i,k}^{(\ell),\sin}
=
\sin\!\left(k\omega \xi_i^{(\ell)}\right).
\end{equation}

As in GeoKAN-NNMetric, LM-KAN also appends the volume feature
\begin{equation}
v^{(\ell)}
=
\sum_{i=1}^{d_{\ell-1}}
\log g_i^{(\ell)}\!\left(h^{(\ell-1)}\right)
=
\log\det
g^{(\ell)}\!\left(h^{(\ell-1)}\right),
\end{equation}
where the last equality follows because the learned metric is diagonal.

Accordingly, the feature vector depends on the chosen basis. For the wavelet
and RBF versions, it is
\begin{equation}
F_{\mathrm{LMKAN}}^{(\ell)}
=
\Big[
\eta_{1,1}^{(\ell)},\dots,\eta_{d_{\ell-1},K}^{(\ell)},v^{(\ell)}
\Big]^\top .
\end{equation}
For the Fourier version, it becomes
\begin{equation}
F_{\mathrm{LMKAN}}^{(\ell)}
=
\Big[
\eta_{1,1}^{(\ell),\cos},
\eta_{1,1}^{(\ell),\sin},
\dots,
\eta_{d_{\ell-1},K}^{(\ell),\cos},
\eta_{d_{\ell-1},K}^{(\ell),\sin},
v^{(\ell)}
\Big]^\top .
\end{equation}

The layerwise pre-activation output is obtained by linearly mixing these
geometry-aware features:
\begin{equation}
\tilde h_{\mathrm{LMKAN}}^{(\ell)}
=
W_{\mathrm{LMKAN}}^{(\ell)}
F_{\mathrm{LMKAN}}^{(\ell)}
+
b_{\mathrm{LMKAN}}^{(\ell)}.
\label{eq:lm_forward_layer_rewrite}
\end{equation}
As in the generic GeoKAN notation, each component of
\(\tilde h_{\mathrm{LMKAN}}^{(\ell)}\) corresponds to the scalar output \(z\) of
one hidden-layer unit.

Equivalently, for output channel \(j=1,\dots,d_\ell\), the expanded scalar form
in the wavelet/RBF case is
\begin{equation}
\tilde h_{\mathrm{LMKAN},j}^{(\ell)}
=
b_{\mathrm{LMKAN},j}^{(\ell)}
+
\sum_{i=1}^{d_{\ell-1}}\sum_{k=1}^{K}
a_{j,i,k}^{(\ell)}
\psi\!\left(
\frac{
h_i^{(\ell-1)}
\sqrt{g_i^{(\ell)}(h^{(\ell-1)})}
-
c_{i,k}^{(\ell)}
}{
s_{i,k}^{(\ell)}
}
\right)
+
a_{j,\mathrm{vol}}^{(\ell)}
\sum_{i=1}^{d_{\ell-1}}
\log g_i^{(\ell)}\!\left(h^{(\ell-1)}\right),
\qquad
\psi\in\{\psi_{\mathrm{wav}},\psi_{\mathrm{rbf}}\}.
\label{eq:lm_forward_expanded_rewrite}
\end{equation}
Equivalently, using the warped-coordinate notation, this can be written more
compactly as
\begin{equation}
\tilde h_{\mathrm{LMKAN},j}^{(\ell)}
=
b_{\mathrm{LMKAN},j}^{(\ell)}
+
\sum_{i=1}^{d_{\ell-1}}\sum_{k=1}^{K}
a_{j,i,k}^{(\ell)}
\psi\!\left(r_{i,k}^{(\ell)}\right)
+
a_{j,\mathrm{vol}}^{(\ell)}
v^{(\ell)}.
\end{equation}

For the Fourier case, the corresponding expansion is
\begin{equation}
\tilde h_{\mathrm{LMKAN},j}^{(\ell)}
=
b_{\mathrm{LMKAN},j}^{(\ell)}
+
\sum_{i=1}^{d_{\ell-1}}\sum_{k=1}^{K}
\left[
a_{j,i,k}^{(\ell),\cos}
\cos\!\left(k\omega \xi_i^{(\ell)}\right)
+
a_{j,i,k}^{(\ell),\sin}
\sin\!\left(k\omega \xi_i^{(\ell)}\right)
\right]
+
a_{j,\mathrm{vol}}^{(\ell)}
v^{(\ell)}.
\label{eq:lm_forward_expanded_fourier}
\end{equation}
For \(\ell>1\), the learned metric acts on the hidden representation
\(h^{(\ell-1)}\) rather than directly on the original input.

All GeoKAN variants follow the same geometric principle. They differ only in
how the metric is learned, how the post-warp features are constructed, and
which additional geometric quantities are included
(Table~\ref{tab:geokan_taxonomy_short}, Fig.~\ref{fig:geokan_variants}).
GeoKAN-NNMetric uses a coupled neural metric and warped wavelet atoms
(Fig.~\ref{fig:geokan_variants}a). GeoKAN-\(\gamma\) uses a separable metric
and explicit geometric features (Fig.~\ref{fig:geokan_variants}b). LM-KAN
combines a coupled learned metric with localized warped atoms
(Fig.~\ref{fig:geokan_variants}c) and is particularly suitable for PDE
approximation. In this way, GeoKAN extends the KAN idea beyond representation
alone: standard KAN variants learn how to represent a function, whereas GeoKAN
additionally learns the geometry in which that representation is constructed.

\begin{table}[H]
\centering
\caption{Comparison of the main GeoKAN variants. Here \(u\) denotes a generic
incoming representation, e.g., \(u=x\) in the first layer and
\(u=h^{(\ell-1)}\) in deeper layers.}
\label{tab:geokan_taxonomy_short}
\renewcommand{\arraystretch}{1.2}
\begin{tabular}{p{2.8cm}p{3.3cm}p{3.3cm}p{3.3cm}}
\hline
\textbf{Property} & \textbf{GeoKAN-NNMetric} & \textbf{GeoKAN-\(\gamma\)} & \textbf{LM-KAN} \\
\hline
Metric model 
& \(g_i(u)\) from full representation 
& \(g_i(u_i)\) only 
& \(g_i(u)\) from full representation \\

Metric realization 
& MLP \(+\) softplus 
& 1D RBF expansion 
& MLP \(+\) softplus \\

Feature construction 
& Warped wavelet basis 
& Metric-derived features 
& Warped wavelet, RBF, or Fourier basis \\

Extra geometry feature 
& \(\log \det g(u)\) 
& \(\gamma_i=\tfrac12 \partial_{u_i}\log g_i(u_i)\) 
& \(\log \det g(u)\) \\

Coupling across coordinates 
& Yes 
& No 
& Yes \\

Main role 
& Flexible coupled metric model 
& Structured lightweight model 
& Coupled metric basis model for PDEs \\
\hline
\end{tabular}
\end{table}

\section{Recovering Structure from Data: Data Fitting}\label{DataFitting_int}

Before turning to physics informed learning, we first study the representational behavior of the proposed models in a supervised approximation setting. This separates two questions. The first is whether the GeoKAN family has an intrinsic advantage as a function approximator. The second is whether that advantage later improves performance in a physics informed framework. To address the first question, we evaluate the models on a controlled one dimensional data fitting benchmark with targets chosen to test oscillation, nonstationarity, discontinuity, localization, and multiscale structure.

\subsection{Matched Capacity Function Approximation Benchmark}

To compare the models fairly, we use a matched capacity benchmark in which all architectures have approximately the same parameter count, about $2\times 10^4$, and the same effective depth. The compared models are the MLP baseline, Efficient-KAN, GeoKAN-$\gamma$, GeoKAN-NNMetric, LM-KAN-Wav, and LM-KAN-RBF. The architectural hyperparameters are chosen so that the parameter counts are closely aligned while preserving the natural structure of each model. The MLP uses width $140$ and depth $2$. Efficient-KAN uses layers $[1,57,57,1]$ with grid size $3$ and spline order $1$. GeoKAN-$\gamma$ uses width $90$, depth $2$, and $K=12$. GeoKAN-NNMetric, LM-KAN-Wav, and LM-KAN-RBF use width $38$, depth $2$, $K=12$, and metric hidden size $8$, while LM-KAN-RBF also uses $\gamma=2.0$. All models are trained with the same epoch budget, the same early stopping logic, and the same train test split for each target. This makes the comparison primarily a test of architectural inductive bias rather than parameter count. The architecture parameters are summarized in Table~\ref{tab:datafit_param_counts}.

\begin{table}[H]
\centering
\caption{Architectures and parameter counts used in the matched-capacity benchmark.}
\label{tab:datafit_param_counts}
\renewcommand{\arraystretch}{1.45}
\small
\begin{tabularx}{\textwidth}{|c|X|c|}
\hline
\textbf{Model} & \textbf{Architecture} & \textbf{Trainable parameters} \\
\hline
MLP & width $=140$, depth $=2$ & $20161$ \\

Efficient-KAN & layers $=[1,57,57,1]$, grid size $=3$, spline order $=1$ & $20178$ \\

GeoKAN-$\gamma$ & width $=90$, depth $=2$, $K=12$ & $20200$ \\

GeoKAN-NNMetric & width $=38$, depth $=2$, $K=12$, metric hidden $=8$ & $19734$ \\

LM-KAN-Wav & width $=38$, depth $=2$, $K=12$, metric hidden $=8$ & $19734$ \\

LM-KAN-RBF & width $=38$, depth $=2$, $K=12$, metric hidden $=8$, $\gamma=2.0$ & $19734$ \\
\hline
\end{tabularx}
\end{table}

The targets are chosen to probe different approximation difficulties. The high frequency sinusoid tests rapid oscillation over a fixed domain. The chirp tests nonstationary frequency content. The step function tests sharp nonsmooth transitions. The narrow needle tests highly localized structure. The multiscale envelope modulated sinusoid combines oscillation with spatially varying amplitude. The sawtooth tests piecewise linear nonsmooth behavior. Together, these targets cover several regimes in which fixed coordinate representations are known to be difficult.

\begin{figure}
    \centering
    \includegraphics[width=0.92\textwidth]{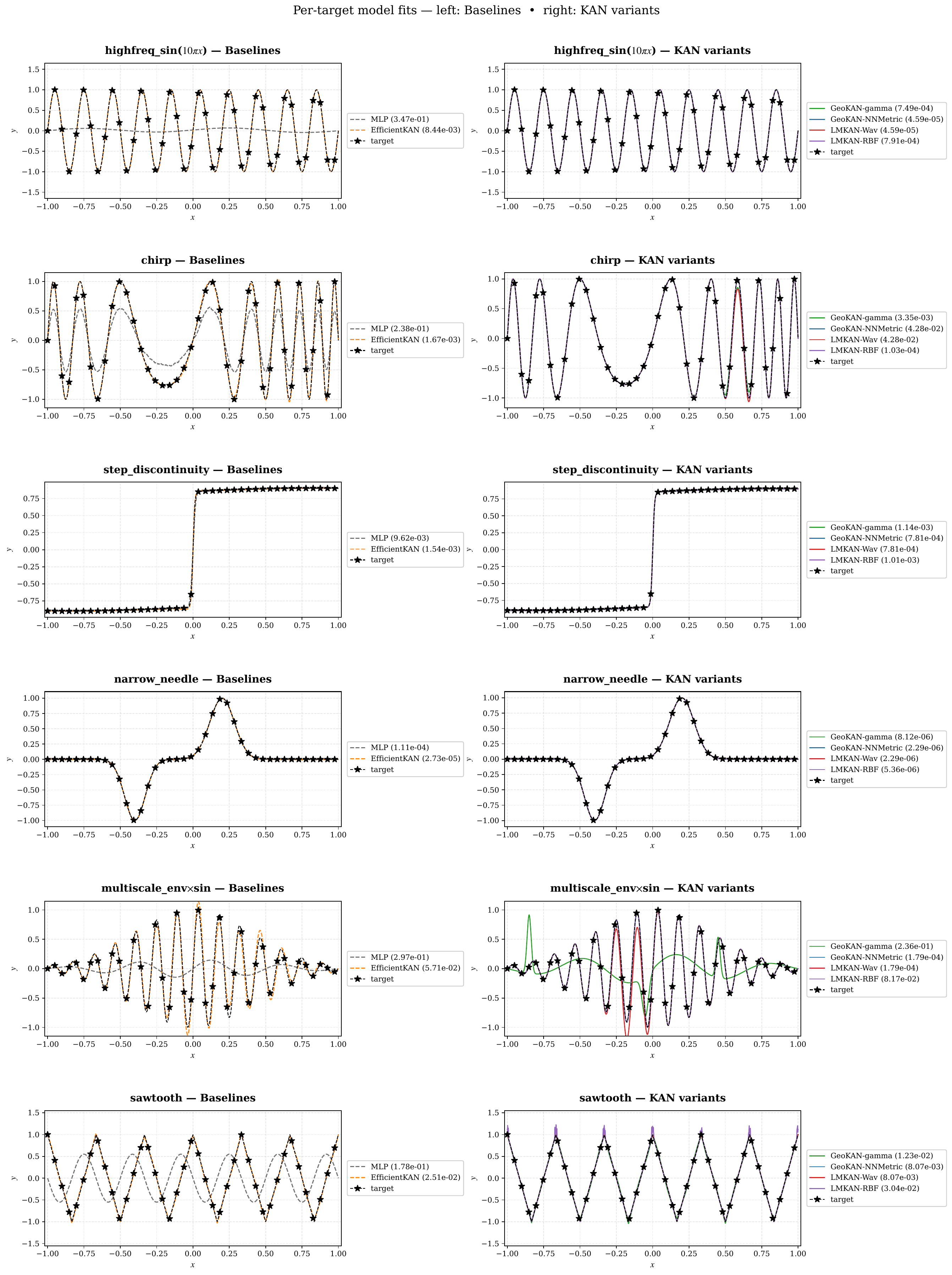}
    \caption{Per-target function approximation fits under the matched capacity benchmark. For each target, the left panel compares the baseline models against the ground truth, while the right panel compares the GeoKAN family variants against the same target. Across several structure rich targets, the geometry aware variants provide visibly tighter fits than the fixed geometry baselines, especially for high frequency, localized, and multiscale functions.}
    \label{fig:datafit_all_targets}
\end{figure}

\begin{figure}
    \centering

    \begin{subfigure}[t]{0.32\textwidth}
        \centering
        \includegraphics[width=\textwidth]{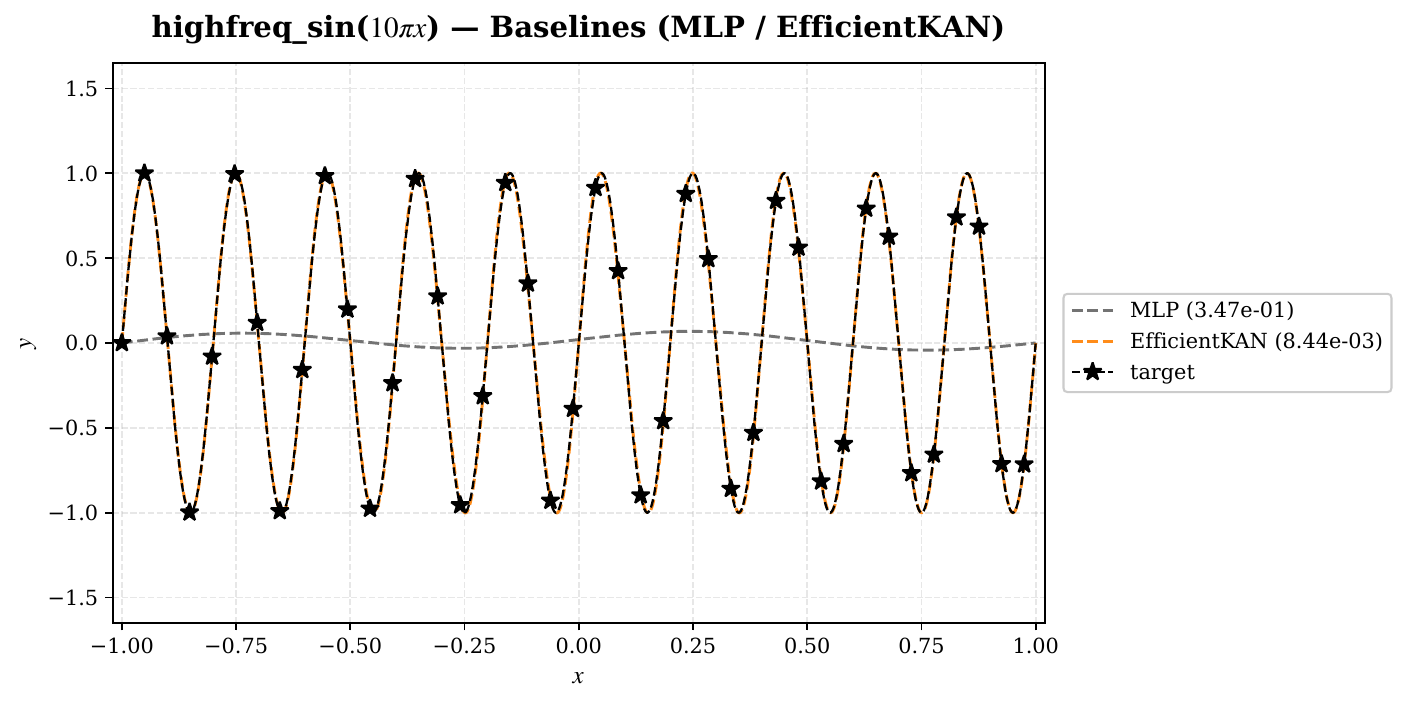}
        \caption{High frequency sinusoid: baselines.}
        \label{fig:highfreq_base}
    \end{subfigure}
    \hfill
    \begin{subfigure}[t]{0.32\textwidth}
        \centering
        \includegraphics[width=\textwidth]{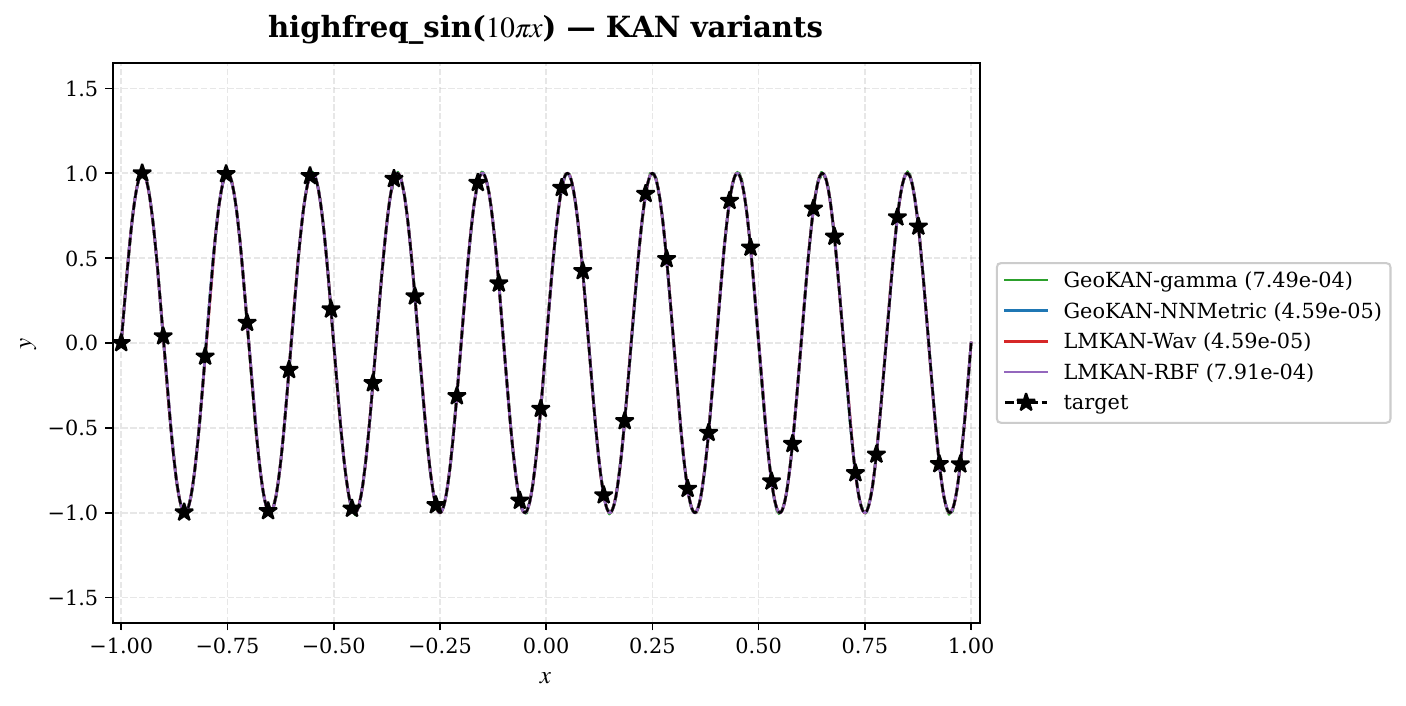}
        \caption{High frequency sinusoid: GeoKAN variants.}
        \label{fig:highfreq_adv}
    \end{subfigure}
    \hfill
    \begin{subfigure}[t]{0.32\textwidth}
        \centering
        \includegraphics[width=\textwidth]{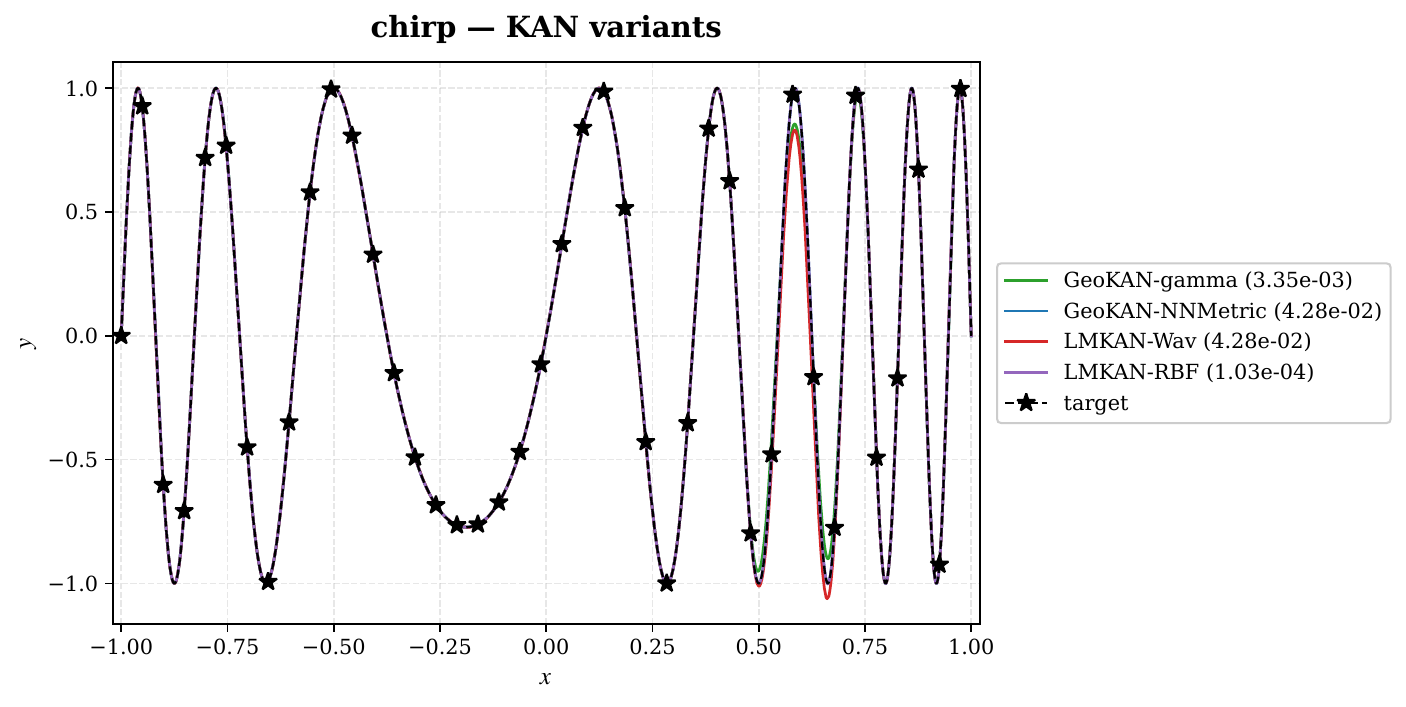}
        \caption{Chirp: GeoKAN variants.}
        \label{fig:chirp_adv}
    \end{subfigure}

    \vspace{0.8em}

    \begin{subfigure}[t]{0.32\textwidth}
        \centering
        \includegraphics[width=\textwidth]{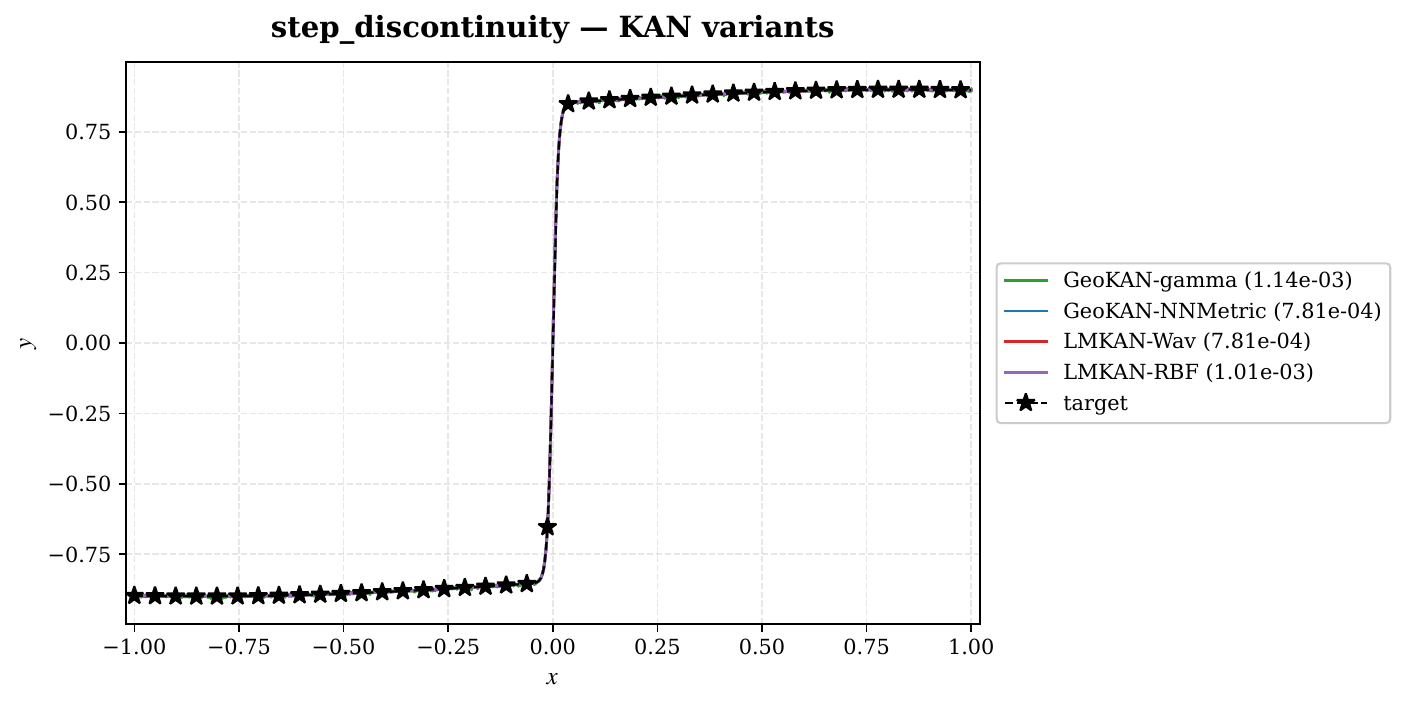}
        \caption{Step discontinuity: GeoKAN variants.}
        \label{fig:step_adv}
    \end{subfigure}
    \hfill
    \begin{subfigure}[t]{0.32\textwidth}
        \centering
        \includegraphics[width=\textwidth]{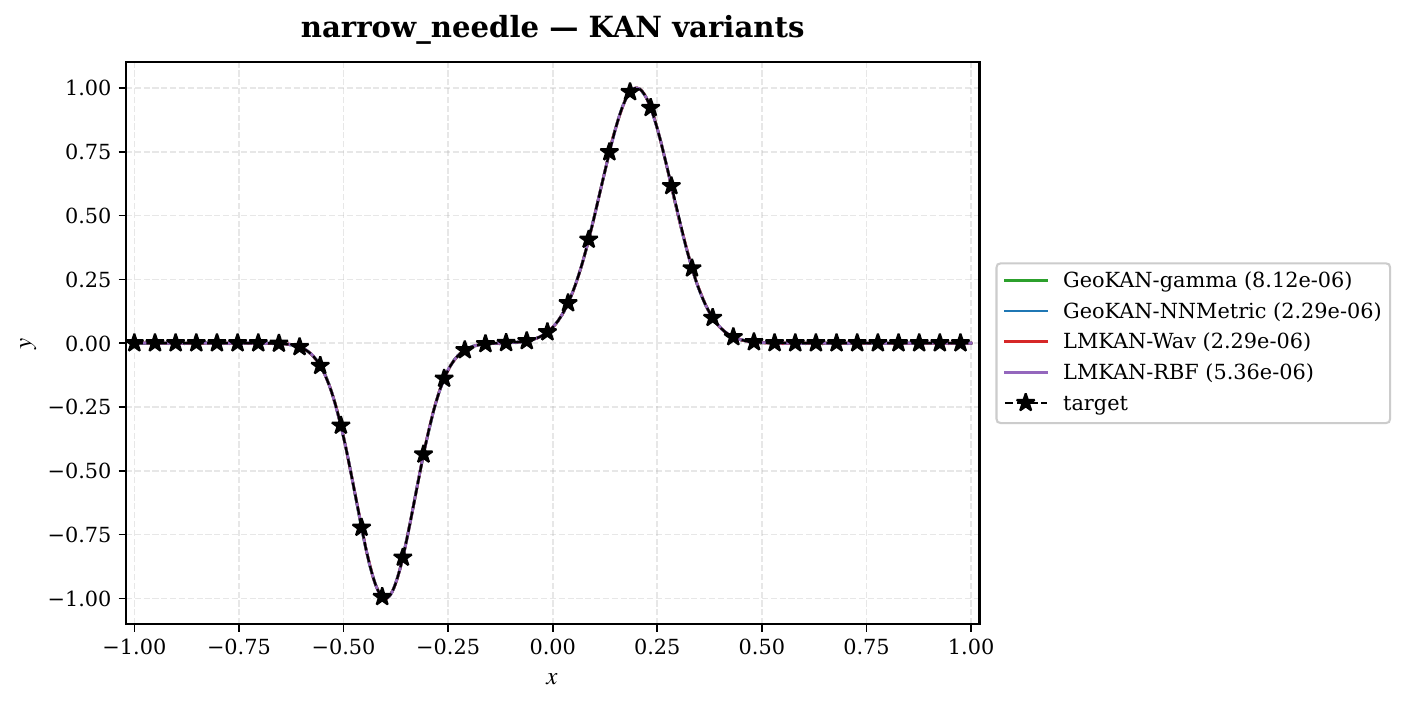}
        \caption{Narrow needle: GeoKAN variants.}
        \label{fig:needle_adv}
    \end{subfigure}
    \hfill
    \begin{subfigure}[t]{0.32\textwidth}
        \centering
        \includegraphics[width=\textwidth]{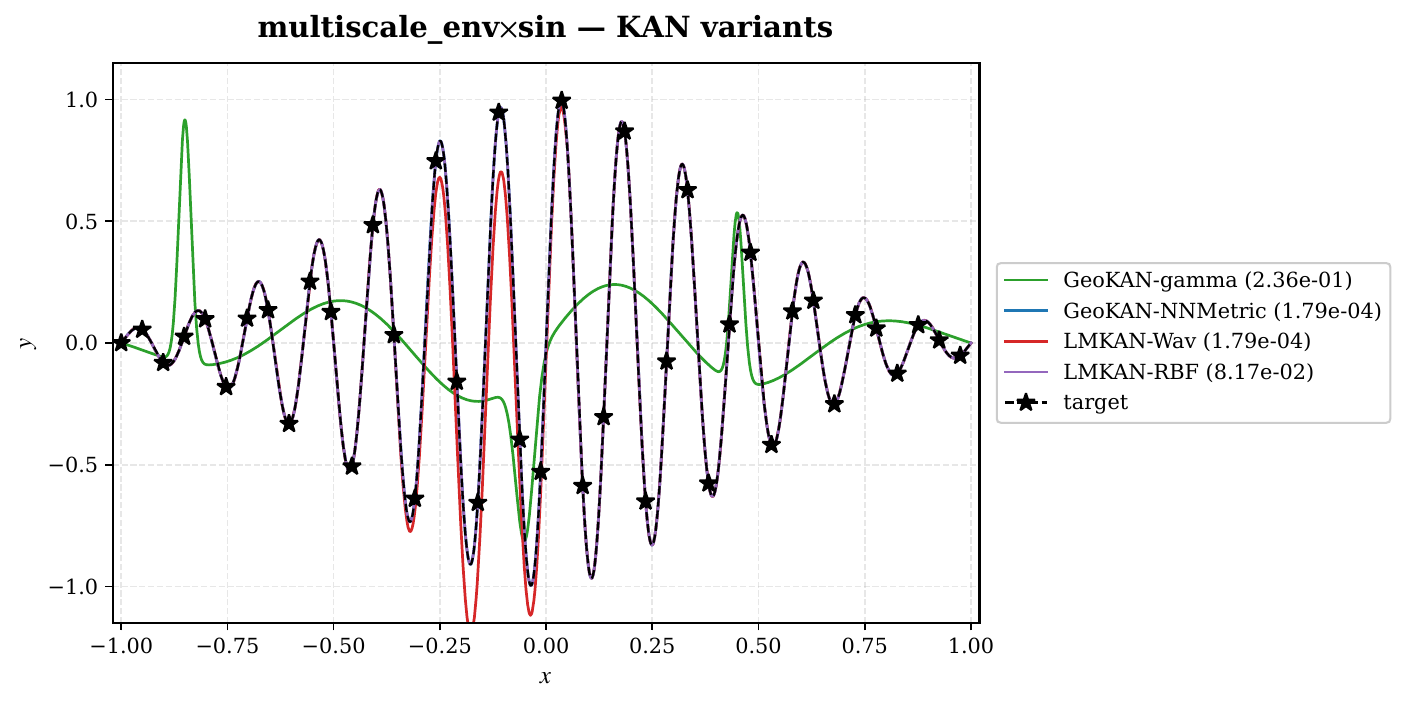}
        \caption{Multiscale envelope modulated sinusoid: GeoKAN variants.}
        \label{fig:multi_adv}
    \end{subfigure}

    \caption{Representative function fitting results from the matched capacity benchmark. The panels illustrate the ability of the GeoKAN family models to capture nonuniform structure, including high frequency oscillation, discontinuity, sharp localization, chirped frequency variation, and multiscale modulation. In most challenging cases, the learned metric variants track the target more closely than the fixed geometry baselines.}
    \label{fig:datafit_selected_targets}
\end{figure}

The results are reported in Table~\ref{tab:datafit_main_results}. GeoKAN-NNMetric and LM-KAN-Wav are the strongest models on several structure rich targets, including the high frequency sinusoid, step discontinuity, narrow needle, multiscale envelope modulated sinusoid, and sawtooth. On $\sin(10\pi x)$, they achieve a test MSE of $4.592\times 10^{-5}$, compared with $8.445\times 10^{-3}$ for Efficient-KAN and $3.472\times 10^{-1}$ for the MLP. On the multiscale envelope modulated sinusoid, they achieve $1.790\times 10^{-4}$, whereas Efficient-KAN gives $5.715\times 10^{-2}$ and the MLP gives $2.974\times 10^{-1}$. On the chirp target, however, LM-KAN-RBF is best with a test MSE of $1.035\times 10^{-4}$. Overall, the benchmark shows that GeoKAN type models provide a clear representational advantage on targets with strong local variation or multiscale structure.

\begin{table}[H]
\centering
\caption{Matched-capacity data-fitting benchmark at approximately $2\times 10^4$ parameters and depth $2$. Reported values are test MSE. Best performance in each row is shown in bold.}
\label{tab:datafit_main_results}
\renewcommand{\arraystretch}{1.5}
\small
\begin{tabularx}{\textwidth}{|X|c|c|c|c|c|c|}
\hline
\textbf{Target} & \textbf{MLP} & \textbf{Efficient-KAN} & \textbf{GeoKAN-$\gamma$} & \textbf{GeoKAN-NNMetric} & \textbf{LM-KAN-Wav} & \textbf{LM-KAN-RBF} \\
\hline
Highfreq sin$(10\pi x)$    & $3.472\times 10^{-1}$ & $8.445\times 10^{-3}$ & $7.487\times 10^{-4}$ & $\mathbf{4.592\times 10^{-5}}$ & $\mathbf{4.592\times 10^{-5}}$ & $7.910\times 10^{-4}$ \\
Chirp                       & $2.384\times 10^{-1}$ & $1.670\times 10^{-3}$ & $3.353\times 10^{-3}$ & $4.284\times 10^{-2}$ & $4.284\times 10^{-2}$ & $\mathbf{1.035\times 10^{-4}}$ \\
Step discontinuity         & $9.619\times 10^{-3}$ & $1.543\times 10^{-3}$ & $1.145\times 10^{-3}$ & $\mathbf{7.806\times 10^{-4}}$ & $\mathbf{7.806\times 10^{-4}}$ & $1.013\times 10^{-3}$ \\
Narrow needle              & $1.110\times 10^{-4}$ & $2.728\times 10^{-5}$ & $8.117\times 10^{-6}$ & $\mathbf{2.294\times 10^{-6}}$ & $\mathbf{2.294\times 10^{-6}}$ & $5.363\times 10^{-6}$ \\
Multiscale env$\times\sin$ & $2.974\times 10^{-1}$ & $5.715\times 10^{-2}$ & $2.355\times 10^{-1}$ & $\mathbf{1.790\times 10^{-4}}$ & $\mathbf{1.790\times 10^{-4}}$ & $8.170\times 10^{-2}$ \\
Sawtooth                    & $1.777\times 10^{-1}$ & $2.512\times 10^{-2}$ & $1.234\times 10^{-2}$ & $\mathbf{8.072\times 10^{-3}}$ & $\mathbf{8.072\times 10^{-3}}$ & $3.038\times 10^{-2}$ \\
\hline
\end{tabularx}
\end{table}

\subsection{Interpretation of the Learned Geometry}

The benchmark results are consistent with the geometric mechanism built into GeoKAN. Standard MLPs operate in a fixed Euclidean coordinate system. Efficient-KAN increases basis flexibility, but still evaluates its learnable functions on the original coordinates. GeoKAN first learns a positive metric field and then performs basis evaluation or geometric feature extraction in the corresponding warped coordinates. This changes how resolution is allocated across the domain. Regions that are difficult to approximate can be stretched, while smoother regions can be compressed.

This provides a simple interpretation of the benchmark results. The multiscale envelope modulated sinusoid, the narrow needle, and the step discontinuity all contain highly nonuniform local structure. A fixed coordinate model must distribute its parameters globally, even when the difficulty is confined to a small part of the domain. A learned metric allows GeoKAN models to redistribute effective resolution toward those difficult regions. Near a sharp peak or a rapidly varying patch, the learned metric can enlarge geometric distance so that a modest number of localized atoms in warped coordinates behaves like a much finer basis. In smoother regions, the geometry can contract the domain and avoid spending unnecessary capacity. This is consistent with the large gains observed on the most nonuniform targets.

A useful way to view this mechanism is through the learned metric $g(x)$ and the warped coordinate
\begin{equation}
z(x)=x\sqrt{g(x)}.
\end{equation}
Where $g(x)$ is large, the map $x\mapsto z(x)$ magnifies local detail and increases effective basis resolution. Where $g(x)$ is smaller, the representation is compressed. The learned geometry therefore acts as a task dependent coordinate transformation.

\subsection{Functional Roles of the GeoKAN Variants}

Although all GeoKAN variants are geometry aware, the benchmark suggests that they play different functional roles. GeoKAN-NNMetric is the most expressive variant. It learns a coupled metric from the full input and combines it with a wavelet style localized dictionary. In this benchmark, it attains the best or tied best result on the high frequency sinusoid, step discontinuity, narrow needle, multiscale envelope modulated sinusoid, and sawtooth. These are settings in which localized responses and geometric adaptation are especially useful.

LM-KAN-Wav behaves similarly and, in the present implementation, matches GeoKAN-NNMetric on several targets. This indicates that the combination of a learned metric with a wavelet type dictionary is particularly effective for sharp transitions, localized oscillations, and multiscale structure. LM-KAN-RBF shows a different behavior. It is the strongest model on the chirp target, where the main difficulty is smoothly varying local frequency rather than a sharp discontinuity or isolated spike. In this regime, the smoother localized character of the RBF dictionary appears to be better matched to the target.

GeoKAN-$\gamma$ is the most structured and lightweight variant. Rather than using a full post warp dictionary, it extracts explicit metric derived features. It does not dominate the benchmark, but it remains competitive and substantially outperforms the MLP and, on several targets, Efficient-KAN. Taken together, these results suggest that the GeoKAN family should be viewed as a collection of geometry aware approximators with different inductive biases rather than as a single uniform model class.

\subsection{Transition to Physics Informed Learning}

The matched capacity data fitting results establish that the advantage of the GeoKAN family appears already at the level of pure representation learning, before any physical constraints are imposed. The improvement is therefore not only a consequence of residual based training. It is already present in the surrogate architecture through the learned metric warp and the geometry adapted feature construction.

This motivates the next stage of the paper. If geometry aware coordinate adaptation improves the approximation of sharp, localized, stiff, or multiscale targets in supervised fitting, then the same mechanism may also be useful when the solution must be inferred from a differential equation together with initial and boundary conditions. We therefore turn next to physics informed Kolmogorov Arnold models built from the GeoKAN framework, with particular emphasis on the LM-KAN surrogate introduced earlier. 

\section{Physics-Informed KAN Models and the LM-KAN Surrogate} \label{pikan_lmkan_results}

Physics-Informed Kolmogorov--Arnold Networks (PIKAN) extend the general philosophy of Physics-Informed Neural Networks (PINN) by replacing the standard deep neural network surrogate with a KAN-type architecture \cite{PIKAN_JMLR_2025}. Thus, the essential idea remains unchanged. One seeks a neural approximation of the unknown solution that is trained not only from data, if available, but also from the governing differential equation together with its initial and boundary conditions. The main distinction from classical PINN lies in the choice of function approximator. Instead of using a multilayer perceptron with fixed activation functions and learnable scalar weights, PIKAN employs a KAN in which the connections carry learnable univariate nonlinear functions.

To illustrate the standard construction, consider a differential equation posed on a domain of interest, together with suitable initial and/or boundary conditions. In the physics-informed setting, the unknown solution is approximated by a neural network $\hat{u}_\theta$, where $\theta$ denotes the trainable parameters of the model. For KAN-type networks, these parameters correspond to the coefficients defining the learnable univariate edge functions. The network is then trained by minimizing a loss function built from the residual of the governing equation and the mismatch in the prescribed conditions. In this way, the network is encouraged to satisfy the physical law throughout the domain while also respecting the auxiliary constraints.

For example, for an ordinary differential equation of the form
\begin{equation}
    \frac{dy(x)}{dx}-f(y(x),x)=0,
    \qquad
    y(a)=y_a,
\end{equation}
the neural approximation $\hat{y}_\theta(x)$ is expected to satisfy
\begin{equation}
    \frac{d\hat{y}_\theta(x)}{dx}-f(\hat{y}_\theta(x),x)\approx 0,
    \qquad
    \hat{y}_\theta(a)=y_a.
\end{equation}
To enforce this, one defines the residual at collocation points $\{x_r^i\}_{i=1}^{N_r}$ by
\begin{equation}
    \mathcal{R}_\theta(x_r^i)
    =
    \frac{d\hat{y}_\theta(x_r^i)}{dx}
    -
    f(\hat{y}_\theta(x_r^i),x_r^i),
\end{equation}
and the corresponding physics loss by
\begin{equation}
    \mathcal{L}_r
    =
    \frac{1}{N_r}
    \sum_{i=1}^{N_r}
    \left|
    \mathcal{R}_\theta(x_r^i)
    \right|^2.
\end{equation}
The initial or boundary conditions may be imposed either as hard constraints, by reparameterizing the network output so that the conditions are satisfied identically, or as soft constraints, by adding penalty terms to the loss function. In the latter case, one typically arrives at a weighted loss of the form
\begin{equation}
    \mathcal{L}_{\mathrm{final}}
    =
    \lambda_r \mathcal{L}_r
    +
    \lambda_{bc}\mathcal{L}_{bc}
    +
    \lambda_{ic}\mathcal{L}_{ic},
    \label{eq:pikan_general_loss}
\end{equation}
where $\mathcal{L}_{bc}$ and $\mathcal{L}_{ic}$ denote the boundary and initial condition losses, respectively, and the coefficients $\lambda_r$, $\lambda_{bc}$, and $\lambda_{ic}$ are introduced to balance the influence of the different terms during optimization.

In some settings, one also augments the physics-based loss with observational or simulated data. This leads to the data-driven form of PIKAN, in which the total loss becomes
\begin{equation}
    \mathcal{L}_{\mathrm{final}}^{DD}
    =
    \lambda_r \mathcal{L}_r
    +
    \lambda_{bc}\mathcal{L}_{bc}
    +
    \lambda_{ic}\mathcal{L}_{ic}
    +
    \lambda_{\mathrm{data}}\mathcal{L}_{\mathrm{data}},
\end{equation}
with
\begin{equation}
    \mathcal{L}_{\mathrm{data}}
    =
    \frac{1}{N_{\mathrm{data}}}
    \sum_{i=1}^{N_{\mathrm{data}}}
    \left|
    \hat{u}_\theta(x_{\mathrm{data}}^i)-u(x_{\mathrm{data}}^i)
    \right|^2.
\end{equation}
Such a formulation is particularly useful when the purely data-free physics loss is difficult to optimize, as can happen for convection-dominated, stiff, or highly nonlinear problems. In this sense, PIKAN follows the same broad strategy as PINN. The governing equation provides the principal supervision, while additional data may be incorporated to improve convergence and accuracy.

Accordingly, much of the earlier PIKAN literature can be understood as using the standard physics-informed training framework described above while changing only the underlying KAN surrogate \cite{KAN}. In other words, the residual-based loss, the treatment of boundary and initial conditions, and the optional use of observational data remain close to the standard PINN methodology. Most successful physics-informed KAN models reported in the literature are based mainly on two KAN variants, namely Efficient-KAN \cite{PIKAN_JMLR_2025, Blealtan2024} and wavelet-KAN (Wav-KAN) \cite{PIKAN_JMLR_2025, WAV-KAN}. In these models, the main change lies in the basis representation used by the surrogate, while the physics-informed optimization framework itself remains essentially the same.

 In the present work, we keep this standard physics-informed framework, but replace the surrogate by \emph{Geometric KAN}, specifically its \emph{Learned-Metric KAN} (LM-KAN) variants (Eqs. \eqref{eq:lm_forward_expanded_rewrite}, \eqref{eq:lm_forward_expanded_fourier}). Thus, the main novelty of our approach is architectural rather than procedural. The residual-based training framework is unchanged, but the network now learns a task-dependent metric and evaluates its localized basis representation in metric-adapted coordinates.

For this reason, in the remainder of the paper we use \emph{PIKAN} mainly to refer to earlier physics-informed KAN baselines, especially those based on Efficient-KAN and Wav-KAN, while \emph{LM-KAN} denotes our proposed geometry-aware surrogate. This distinction is important for the comparisons that follow, since our goal is to evaluate whether the learned geometric adaptation in LM-KAN yields an advantage over earlier PIKAN models under the same physics-informed training setup.

\subsection{Differential Equation Case Studies}

In this work, the LM-KAN architecture is specified by the hidden width, network depth, number of basis functions per input dimension \(K\), basis family, kernel parameter \(\gamma\) when applicable, and the hidden width of the learned metric network. Unless noted otherwise, each hidden LM-KAN block consists of a \texttt{LearnedMetricBasisLayer} followed by a \(\tanh\) activation, while the final layer is a linear output head. The metric subnetwork used to generate the diagonal metric coefficients is implemented as
\[
\texttt{Linear} \rightarrow \texttt{SiLU} \rightarrow \texttt{Linear} \rightarrow \texttt{SiLU} \rightarrow \texttt{Linear} \rightarrow \texttt{Softplus},
\]
so that the learned diagonal metric remains strictly positive.

The benchmark problems considered here are trained with a small amount of additional data supervision together with the physics-informed residual and constraint losses. In all equations, data-driven loss terms are included, but they account for only 10\% of the training information. These data points provide limited guidance while preserving the predominantly physics-informed character of the training. For Burgers' equation, the additional data help resolve the steep gradient more accurately. For Allen--Cahn Case 2 and Helmholtz, they provide useful supervision in regimes where the solution structure is harder to learn. The same 10\% data-assisted setting is also used for the remaining Allen--Cahn and Lorenz experiments.

The selected benchmark equations span several characteristic challenges in scientific machine learning. Burgers' equation tests the ability of the model to resolve nonlinear advection--diffusion dynamics and sharp localized gradients. The Allen--Cahn equation probes reaction--diffusion behavior with thin transition layers, making it a useful test of how well the model captures interface dynamics and multiscale structure. The Helmholtz equation examines the representation of oscillatory and wave-like solutions, and is therefore well suited for assessing approximation quality in problems with high-frequency spatial patterns. The Lorenz system provides a strongly coupled nonlinear ODE benchmark that tests the stability and accuracy of the learned temporal dynamics. Taken together, these case studies assess whether LM-KAN improves representation efficiency and training behavior across qualitatively different regimes.

\subsection{Allen--Cahn Equation}
\paragraph{Case 1:}

The Allen--Cahn equation is a prototypical nonlinear reaction--diffusion model used to describe phase separation and interface evolution in materials science. In this case study, we consider the one-dimensional Allen--Cahn equation with diffusion coefficient \(\nu=0.001\):
\begin{align}
\label{allen_1}
u_t &= \nu u_{xx} - u^3 + u, \notag\\
&\begin{cases}
u(x,0) = 0.53x + 0.47\sin(-1.5\pi x),\\
u(1,t) = 1,\\
u(-1,t) = -1,
\end{cases}
\end{align}
for \(x\in[-1,1]\) and \(t\in[0,1]\).

Allen--Cahn Case 1 constitutes a representative benchmark for nonlinear PDE learning with localized interface dynamics. The combination of cubic reaction, diffusion, and nontrivial initial data produces evolving solution profiles with sharp internal layers and spatially non-uniform behavior. In particular, the small diffusion coefficient \(\nu=0.001\) gives rise to narrow transition regions that are difficult to approximate using fixed-resolution representations. This makes the problem well suited for evaluating whether geometry-aware models can adapt their effective representational resolution to capture steep gradients and interface motion more efficiently than standard fixed-geometry architectures.

To solve Eq.~\eqref{allen_1}, we employ physics-informed KAN-type models and compare the baseline PIKAN with the proposed LM-KAN formulation. The physics-informed residual is defined as
\begin{align}
\mathcal{R}_\theta(x,t)
=
\hat{u}_t + \hat{u}^3 - \hat{u} - \nu \hat{u}_{xx},
\end{align}
and the total training objective is given by
\begin{align}
\mathcal{L}_{\mathrm{r}} &=
\frac{1}{N_r}\sum_{i=1}^{N_r}
\left|\mathcal{R}_\theta(x_r^i,t_r^i)\right|^2, \nonumber\\
\mathcal{L}_{\mathrm{ic}} &=
\frac{1}{N_{\mathrm{ic}}}\sum_{j=1}^{N_{\mathrm{ic}}}
\left(\hat{u}(x_{\mathrm{ic}}^j,0)-u(x_{\mathrm{ic}}^j,0)\right)^2, \nonumber\\
\mathcal{L}_{\mathrm{bc}} &=
\frac{1}{N_{\mathrm{bc}}}\sum_{k=1}^{N_{\mathrm{bc}}}
\left[
\bigl(\hat{u}(1,t_{\mathrm{bc}}^k)-1\bigr)^2
+
\bigl(\hat{u}(-1,t_{\mathrm{bc}}^k)+1\bigr)^2
\right], \nonumber\\
\mathcal{L}_{\mathrm{final}}^{DF}
&=
\mathcal{L}_{\mathrm{r}}+\mathcal{L}_{\mathrm{ic}}+\mathcal{L}_{\mathrm{bc}}.
\label{eq:final_loss_allen_case1}
\end{align}

For Allen--Cahn Case 1, the PIKAN baseline uses the architecture $[2,12,8,12,1]$ with cubic B-spline edge functions on a grid of size 5 and sinusoidal base activation. The corresponding LM-KAN model uses two learned-metric basis layers of width 5, with \(K=7\) RBF basis functions per input dimension and a compact metric network of hidden size 9 (Table~\ref{tab:allen_case1_arch}). This configuration was selected by grid search to keep the model small while still allowing geometry-adaptive resolution in the \((x,t)\)-domain.

\begin{table}
\centering
\caption{Architecture comparison for Allen--Cahn Case 1 between PIKAN vs LM-KAN}
\label{tab:allen_case1_arch}
\renewcommand{\arraystretch}{1.2}
\small
\begin{tabularx}{\textwidth}{|c|X|c|X|X|c|c|}
\hline
Model & Architecture & Depth & Basis resolution & Basis type & Metric hidden & Output \\
\hline
PIKAN & $[2,12,8,12,1]$ & 3 hidden layers & Grid size $=5$ & Cubic B-spline + $\sin$ base & -- & 1 \\
\hline
LM-KAN & $(2 \rightarrow 5 \rightarrow 5 \rightarrow 1)$ & 2 metric-basis layers & $K=7$ per input dim & RBF, $\gamma=2.0$ & 9 & 1 \\
\hline
\end{tabularx}
\end{table}

\begin{figure}
    \centering
    \begin{subfigure}[t]{0.49\textwidth}
        \centering
        \includegraphics[width=\textwidth]{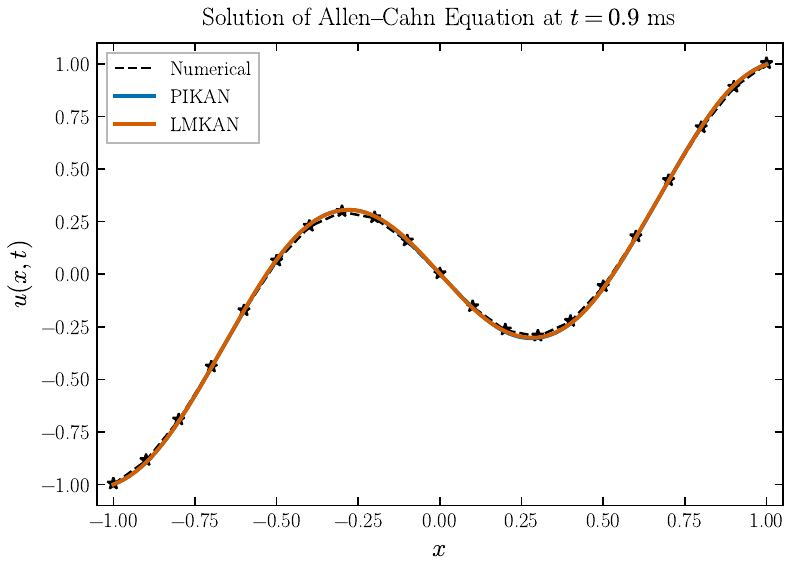}
        \caption{\(t=0.9\) ms}
    \end{subfigure}
    \hfill
    \begin{subfigure}[t]{0.49\textwidth}
        \centering
        \includegraphics[width=\textwidth]{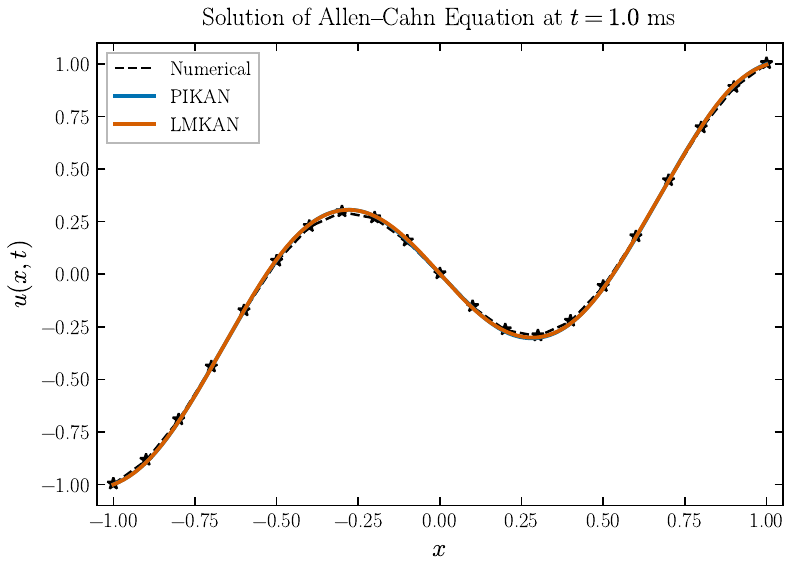}
        \caption{\(t=1.0\) ms}
    \end{subfigure}
    \hfill
    \begin{subfigure}[t]{0.49\textwidth}
        \centering
        \includegraphics[width=\textwidth]{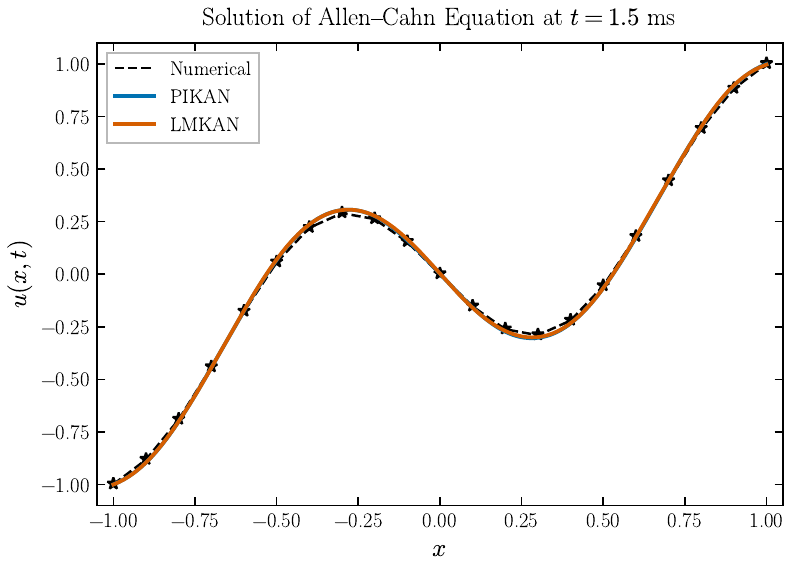}
        \caption{\(t=1.5\) ms}
    \end{subfigure}
    \caption{Solution profile comparison at three representative time instants. The PIKAN and LM-KAN predictions closely overlap with the numerical solution, demonstrating high pointwise accuracy throughout the evolution.}
    \label{fig:allen_case1_profiles}
\end{figure}

\begin{figure}
    \centering
    \begin{subfigure}[t]{0.49\textwidth}
        \centering
        \includegraphics[width=\textwidth]{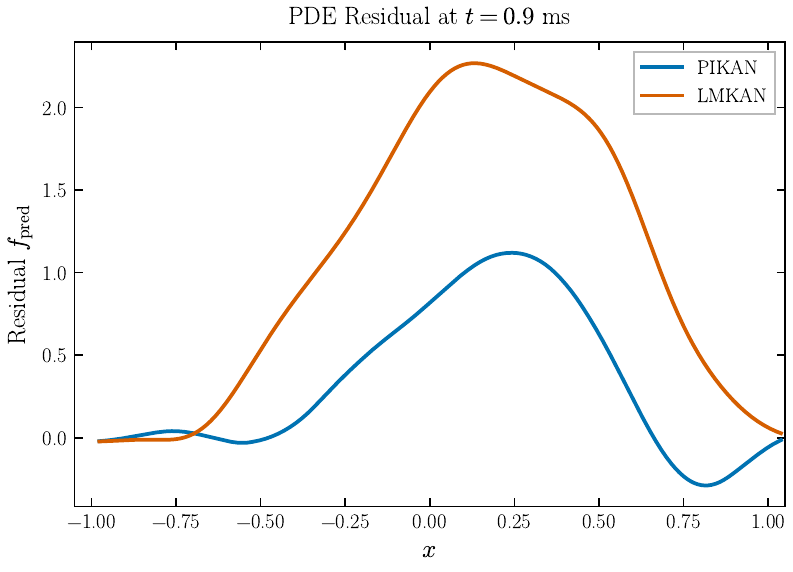}
        \caption{\(t=0.9\) ms}
    \end{subfigure}
    \hfill
    \begin{subfigure}[t]{0.49\textwidth}
        \centering
        \includegraphics[width=\textwidth]{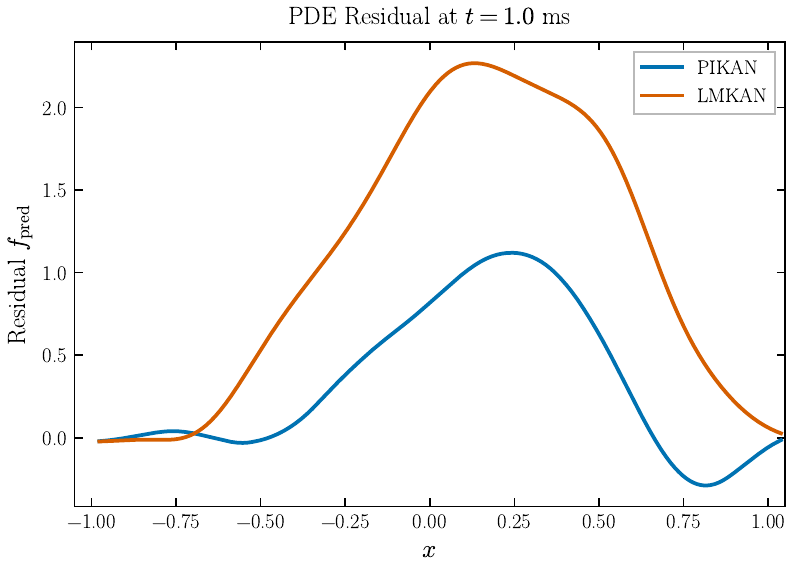}
        \caption{\(t=1.0\) ms}
    \end{subfigure}
    \hfill
    \begin{subfigure}[t]{0.49\textwidth}
        \centering
        \includegraphics[width=\textwidth]{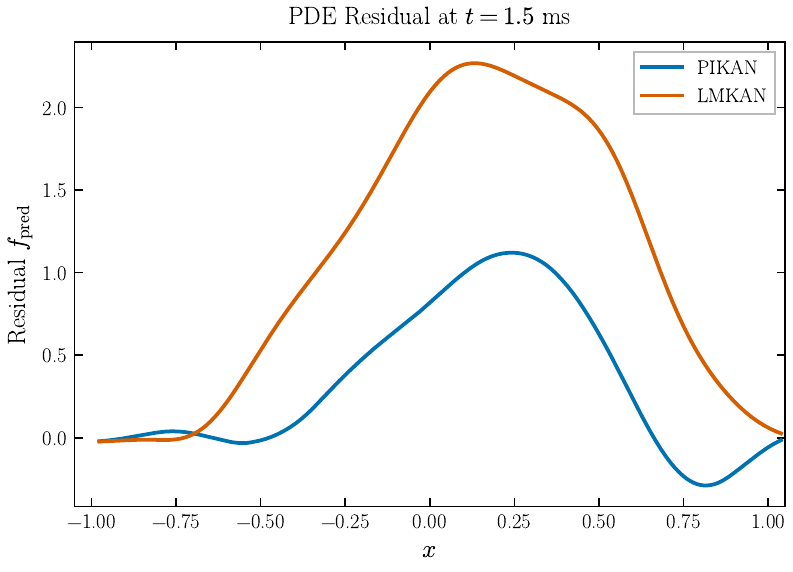}
        \caption{\(t=1.5\) ms}
    \end{subfigure}
    \caption{PDE residual comparison for PIKAN and LM-KAN at three representative time instants. Both methods satisfy the governing equation well, while PIKAN exhibits consistently smaller residual magnitudes across the spatial domain.}
    \label{fig:allen_case1_residual}
\end{figure}

For training, we use \(N_x=100\) collocation points in space and \(N_t=100\) collocation points in time, uniformly distributed over the computational domain. The network is optimized for 10,000 epochs using the AdamW optimizer with learning rate \(\eta=10^{-3}\). To first assess pointwise predictive accuracy, Fig.~\ref{fig:allen_case1_profiles} compares the numerical and predicted solution profiles at \(t=0.9\), \(1.0\), and \(1.5\) ms. In all three snapshots, the predicted curves are nearly indistinguishable from the numerical solution, indicating that both models recover the nonlinear solution profile with high accuracy. PIKAN shows slightly closer agreement with the reference curve at some locations, although LM-KAN remains comparably accurate overall.

The corresponding PDE residual distributions are shown in Fig.~\ref{fig:allen_case1_residual} at the same three time instants. The residual remains small across most of the spatial domain for both models, confirming that the learned solutions satisfy the governing equation well. However, PIKAN consistently exhibits lower residual magnitudes than LM-KAN, suggesting somewhat stronger PDE enforcement in this test case.

\begin{figure}
    \centering
    \begin{subfigure}[t]{0.6\textwidth}
        \centering
        \includegraphics[width=\textwidth]{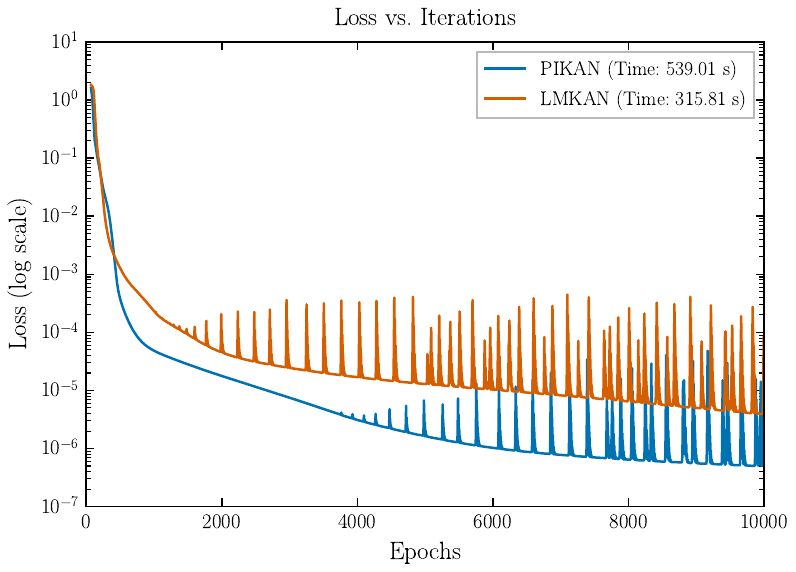}
        \caption{}
    \end{subfigure}
    \hfill
    \begin{subfigure}[t]{1.0\textwidth}
        \centering
        \includegraphics[width=\textwidth]{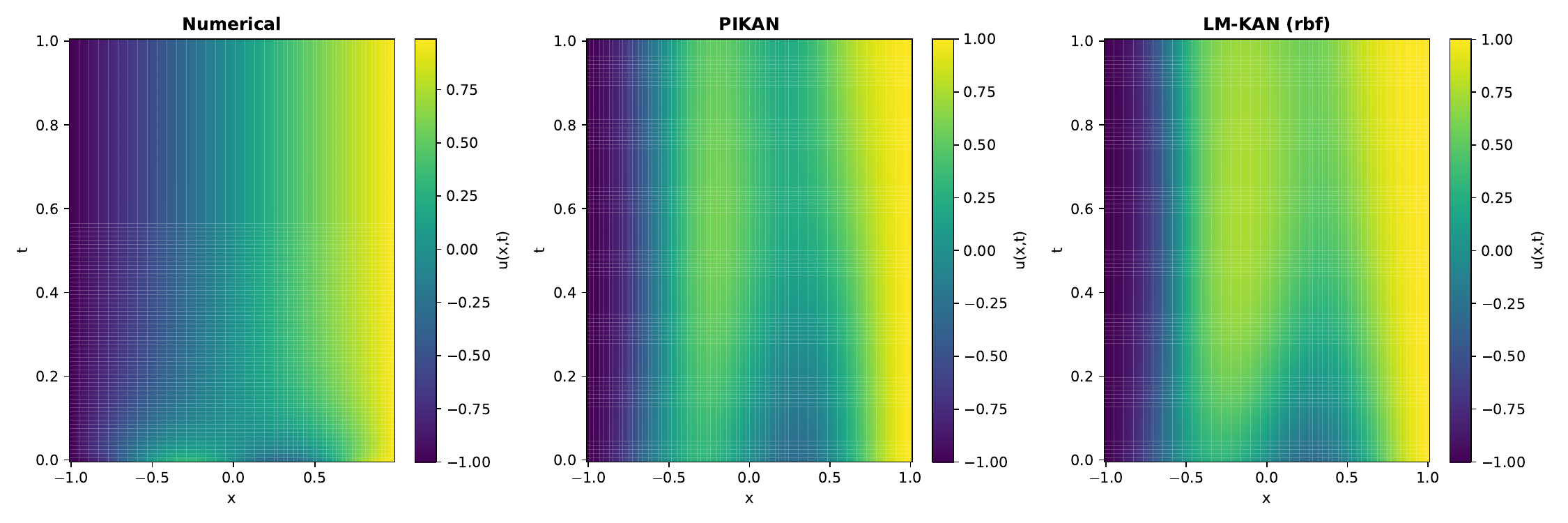}
        \caption{}
    \end{subfigure}
    \caption{Comparison of PIKAN and LM-KAN for Allen--Cahn Case 1: (a) training loss versus epochs, and (b) spatiotemporal solution heatmaps for the numerical reference, PIKAN, and LM-KAN predictions. Both models recover the overall solution evolution accurately, while LM-KAN converges in less wall-clock time and PIKAN reaches a lower final loss.}
    \label{fig:allen_case1_loss_heatmap}
\end{figure}

Finally, Fig.~\ref{fig:allen_case1_loss_heatmap} presents a broader comparison of the training behavior and spatiotemporal predictions of PIKAN and LM-KAN for Allen--Cahn Case 1 after \(10{,}000\) training epochs. As shown in Fig.~\ref{fig:allen_case1_loss_heatmap}(a), both models exhibit stable convergence, demonstrating that each architecture is capable of learning the underlying solution dynamics robustly. PIKAN attains the smaller final loss, \(1\times10^{-6}\), whereas LM-KAN reaches \(5\times10^{-6}\). Despite this, LM-KAN completes training in substantially less wall-clock time, requiring \(315.81\)~s compared with \(539.01\)~s for PIKAN. The model sizes are also similar, with \(2280\) trainable parameters for PIKAN and \(2136\) for LM-KAN. A quantitative comparison between the two models is reported in Table~\ref{tab:allen_case1_comparison}. Fig.~\ref{fig:allen_case1_loss_heatmap}(b) further shows that both methods accurately reconstruct the spatiotemporal solution field, capturing the overall evolution and principal interface structure throughout the considered time interval.

Overall, the results for Allen--Cahn Case 1 show that both PIKAN and LM-KAN provide accurate physics-informed surrogates. The profile comparisons in Fig.~\ref{fig:allen_case1_profiles} and the residual plots in Fig.~\ref{fig:allen_case1_residual} confirm that both models achieve high solution accuracy and good PDE satisfaction, while the global comparison in Fig.~\ref{fig:allen_case1_loss_heatmap} highlights the main trade-off: PIKAN achieves slightly better residual accuracy and a lower final loss, whereas LM-KAN offers a substantial reduction in training time while preserving a highly accurate approximation of the solution field.

\begin{table}
\centering
\caption{Comparison between PIKAN and LM-KAN for Allen--Cahn Case 1 after \(10{,}000\) training epochs. PIKAN attains the smaller final loss, while LM-KAN completes training in less wall-clock time with a slightly smaller parameter count.}
\renewcommand{\arraystretch}{1.2}
\begin{tabular}{c|c|c|c}
\hline
Model & Parameters & Training time (s) & Final loss \\
\hline
\textbf{PIKAN}  & 2280 & 539.01 & \(1\times10^{-6}\) \\
\textbf{LM-KAN} & 2136 & 315.81 & \(5\times10^{-6}\) \\
\hline
\end{tabular}

\label{tab:allen_case1_comparison}
\end{table}

\paragraph{Case 2:}

We next consider a more challenging Allen--Cahn test problem, characterized by a smaller diffusion coefficient and a stronger nonlinear reaction term. The governing equation is given by \cite{allen-cohen}
\begin{align}
\label{allen_2}
u_t &= \nu u_{xx} - 5u^3 + 5u, \notag\\
&\begin{cases}
u(x,0)=x^2\cos(\pi x),\\
u(1,t)=1,\\
u(-1,t)=-1,
\end{cases}
\end{align}
where \(\nu=0.0001\), \(x\in[-1,1]\), and \(t\in[0,1]\).

Compared with Case~1, this problem introduces additional complexity through both the smaller diffusion coefficient and the stronger nonlinear reaction term. The reduced diffusion, \(\nu=0.0001\), generates narrower transition layers, while the amplified reaction term accelerates the local phase dynamics and increases the stiffness of the equation. Consequently, the solution exhibits sharper interfaces and stronger spatial non-uniformity, making accurate approximation more challenging. This setting therefore provides a more stringent benchmark for evaluating whether geometry-aware models can adapt their effective resolution to localized structures more efficiently than fixed-geometry architectures.

\begin{figure}
    \centering
    \begin{subfigure}[t]{0.49\textwidth}
        \centering
        \includegraphics[width=\textwidth]{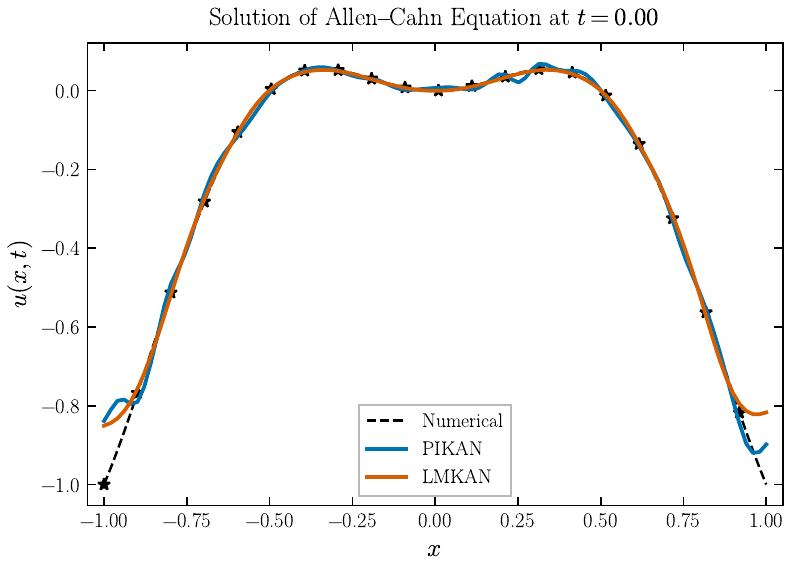}
        \caption{\(t=0.00\)}
    \end{subfigure}
    \hfill
    \begin{subfigure}[t]{0.49\textwidth}
        \centering
        \includegraphics[width=\textwidth]{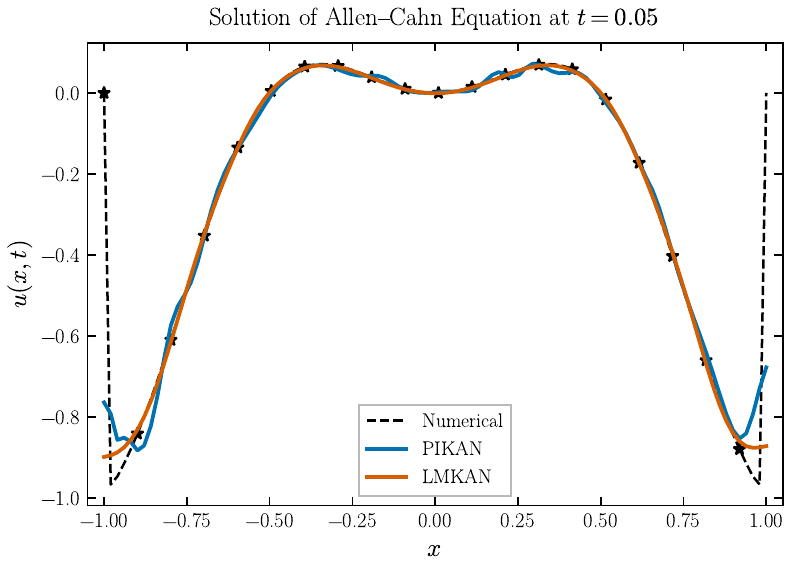}
        \caption{\(t=0.05\)}
    \end{subfigure}
    \hfill
    \begin{subfigure}[t]{0.49\textwidth}
        \centering
        \includegraphics[width=\textwidth]{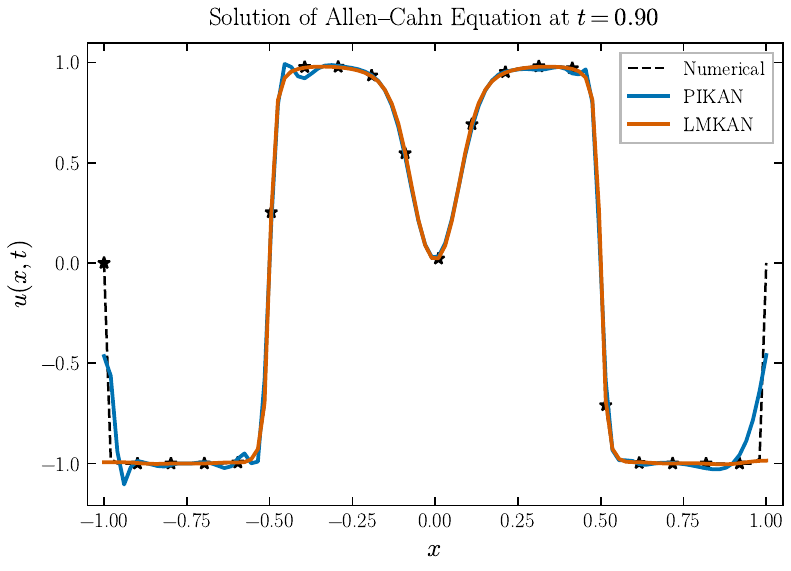}
        \caption{\(t=0.90\)}
    \end{subfigure}
    \caption{Solution profile comparison at representative time instants for Allen--Cahn Case 2. Both models follow the numerical solution closely, while LM-KAN shows improved agreement across the evolving interface region.}
    \label{fig:allen_case2_profiles}
\end{figure}

To solve Eq.~\eqref{allen_2}, we use a data-assisted physics-informed formulation. In addition to the PDE residual, initial-condition, and boundary-condition losses, a supervised data term is introduced using a sparse subset of the numerical reference solution. The residual and training losses are defined as
\begin{align}
\mathcal{R}_\theta(x,t) &= \hat{u}_t + 5\hat{u}^3 - 5\hat{u} - \nu \hat{u}_{xx}, \nonumber\\
\mathcal{L}_{\mathrm{r}} &= \frac{1}{N_r}\sum_{i=1}^{N_r}\left|\mathcal{R}_\theta(x_r^i,t_r^i)\right|^2, \nonumber\\
\mathcal{L}_{\mathrm{ic}} &= \frac{1}{N_{\mathrm{ic}}}\sum_{j=1}^{N_{\mathrm{ic}}}
\left(\hat{u}(x_{\mathrm{ic}}^j,0)-(x_{\mathrm{ic}}^j)^2\cos(\pi x_{\mathrm{ic}}^j)\right)^2, \nonumber\\
\mathcal{L}_{\mathrm{bc}} &= \frac{1}{N_{\mathrm{bc}}}\sum_{k=1}^{N_{\mathrm{bc}}}
\left[
\bigl(\hat{u}(1,t_{\mathrm{bc}}^k)-1\bigr)^2
+
\bigl(\hat{u}(-1,t_{\mathrm{bc}}^k)+1\bigr)^2
\right], \nonumber\\
\mathcal{L}_{\mathrm{data}} &= \frac{1}{N_{\mathrm{data}}}\sum_{p=1}^{N_{\mathrm{data}}}
\left|\hat{u}(x_{\mathrm{data}}^p)-u(x_{\mathrm{data}}^p)\right|^2, \nonumber\\
\mathcal{L}_{\mathrm{final}}^{DD}
&=
\mathcal{L}_{\mathrm{r}}+\mathcal{L}_{\mathrm{ic}}+\mathcal{L}_{\mathrm{bc}}+\mathcal{L}_{\mathrm{data}}.
\label{allen_loss2}
\end{align}

For Allen--Cahn Case 2, the PIKAN baseline again uses the architecture $[2,12,8,12,1]$ with cubic B-spline parameterization and sinusoidal base activation. To enable a controlled comparison at similar model capacity, both LM-KAN variants use the same compact architecture, consisting of two learned-metric basis layers of width 5 with \(K=7\) basis functions per input dimension and a metric-network hidden width of 9 (Table~\ref{tab:allen_case2_arch}). The only difference between the two LM-KAN models is the basis family. One uses Gaussian radial basis functions, while the other uses a Mexican-hat wavelet basis. This isolates the effect of the basis choice from the geometric adaptation mechanism itself.The LM-KAN hyperparameters for this case were chosen by grid search so as to match the parameter count of the PIKAN baseline as closely as possible, thereby making the comparison focus on representation mechanism rather than model size.

In this experiment, the numerical reference solution is first computed, and only \(10\%\) of the data is randomly sampled for inclusion through the data-driven loss term. The model is trained using \(N_x=100\) spatial and \(N_t=100\) temporal collocation points, uniformly distributed over the computational domain. Training is carried out for \(20{,}000\) epochs with the AdamW optimizer and learning rate \(\eta=10^{-3}\).

\begin{table}
\centering
\caption{Architecture comparison for Allen--Cahn Case 2. The two LM-KAN variants share the same learned-metric architecture and differ only in the basis family.}
\label{tab:allen_case2_arch}
\renewcommand{\arraystretch}{1.2}
\small
\begin{tabularx}{\textwidth}{|c|X|c|c|X|c|c|}
\hline
Model & Architecture & Depth & Resolution & Basis & Metric hidden & Output \\
\hline
PIKAN & $[2,12,8,12,1]$ & 3 & Grid $=5$ & Cubic B-spline + $\sin$ & -- & 1 \\
\hline
LM-KAN (RBF) & $(2 \rightarrow 5 \rightarrow 5 \rightarrow 1)$ & 2 & $K=7$ & RBF, $\gamma=2.0$ & 9 & 1 \\
\hline
LM-KAN (Wavelet) & $(2 \rightarrow 5 \rightarrow 5 \rightarrow 1)$ & 2 & $K=7$ & Mexican-hat wavelet & 9 & 1 \\
\hline
\end{tabularx}

\end{table}

To first examine pointwise predictive accuracy, Fig.~\ref{fig:allen_case2_profiles} presents one-dimensional solution profiles at \(t=0.00\), \(0.05\), and \(0.90\). At all three time instants, both PIKAN and LM-KAN capture the main shape of the numerical solution and reproduce the dominant interface structure. However, LM-KAN generally follows the reference profile more closely, particularly in the interior region and near the evolving transition zone, whereas PIKAN exhibits slightly larger deviations. These profile comparisons indicate that both models are able to recover the nonlinear solution behavior, but LM-KAN provides the more accurate local approximation in this more challenging setting.

\begin{figure}
    \centering
    \begin{subfigure}[t]{1.0\textwidth}
        \centering
        \includegraphics[width=\textwidth]{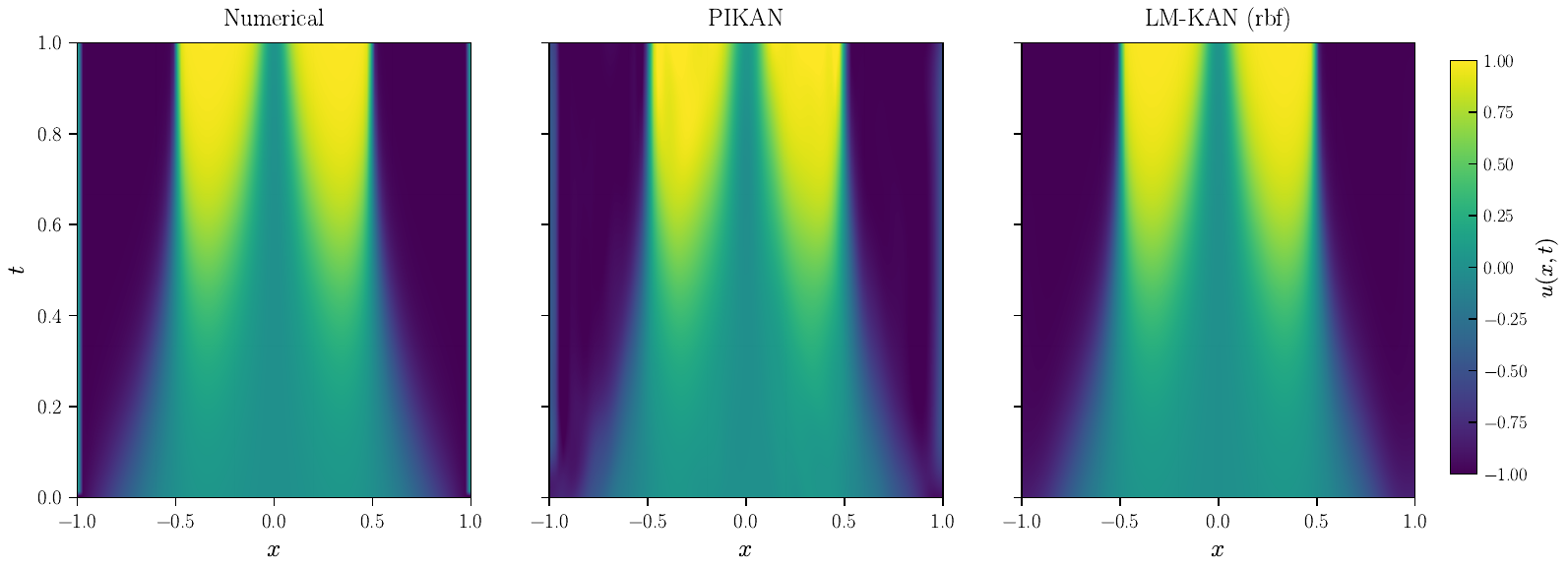}
        \caption{Spatiotemporal solution heatmaps for the numerical reference, PIKAN prediction, and LM-KAN prediction. Both models capture the main evolution pattern, with LM-KAN showing closer agreement with the reference field.}
        \label{fig:allen_case2_heatmaps}
    \end{subfigure}

    \vspace{0.3cm}

    \begin{subfigure}[t]{0.95\textwidth}
        \centering
        \includegraphics[width=\textwidth]{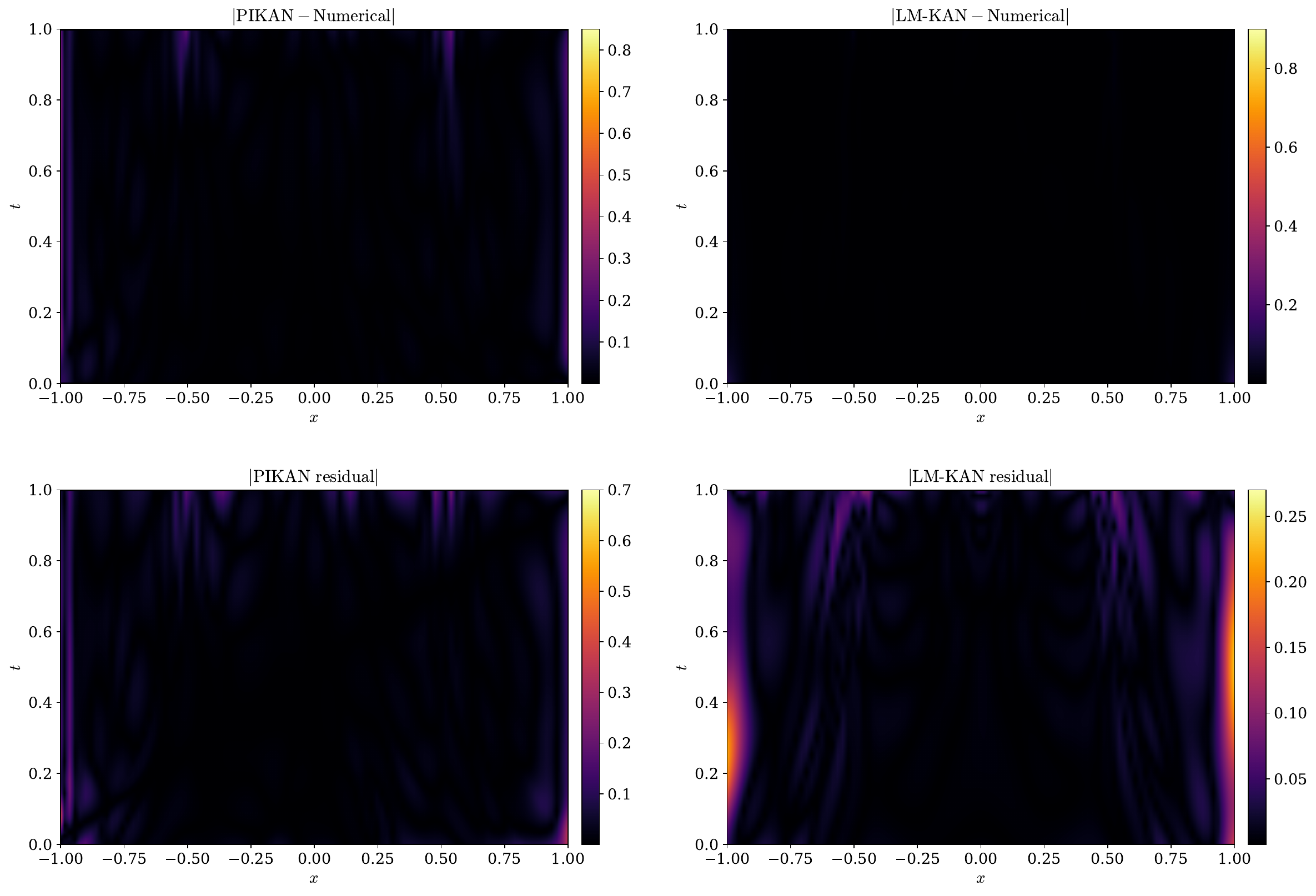}
        \caption{Absolute error and PDE residual heatmaps for PIKAN and LM-KAN. LM-KAN exhibits both smaller prediction error and lower residual magnitude over most of the spatiotemporal domain.}
        \label{fig:allen_case2_error_residual}
    \end{subfigure}
    \caption{Spatiotemporal comparison for Allen--Cahn Case 2. The predicted solution fields, error maps, and PDE residual distributions show that both models capture the overall dynamics, while LM-KAN provides closer agreement with the numerical reference and stronger physics consistency across the domain.}
    \label{fig:allen_case2_heatmap_residual_overview}
\end{figure}

A broader view of the predicted solution fields is given in Fig.~\ref{fig:allen_case2_heatmap_residual_overview}(a), which compares the spatiotemporal heatmaps of the numerical reference, PIKAN prediction, and LM-KAN prediction. Both models successfully reproduce the global spatiotemporal evolution of the Allen--Cahn solution, including the formation and persistence of the central transition structure. The qualitative agreement is good in both cases, confirming that the main dynamics are captured over the full domain. Nevertheless, LM-KAN appears visually closer to the reference field, especially in regions where the solution varies more sharply.

This trend is further confirmed by the absolute error and PDE residual maps shown in Fig.~\ref{fig:allen_case2_heatmap_residual_overview}(b). The absolute error \(|\mathrm{prediction}-\mathrm{numerical}|\) for LM-KAN is uniformly smaller across most of the spatiotemporal domain than that of PIKAN. Likewise, the PDE residual magnitude is substantially lower for LM-KAN, indicating that it satisfies the governing equation more accurately in this data-assisted setting.

\begin{figure}
    \centering
    \begin{subfigure}[t]{0.7\textwidth}
        \centering
        \includegraphics[width=\textwidth]{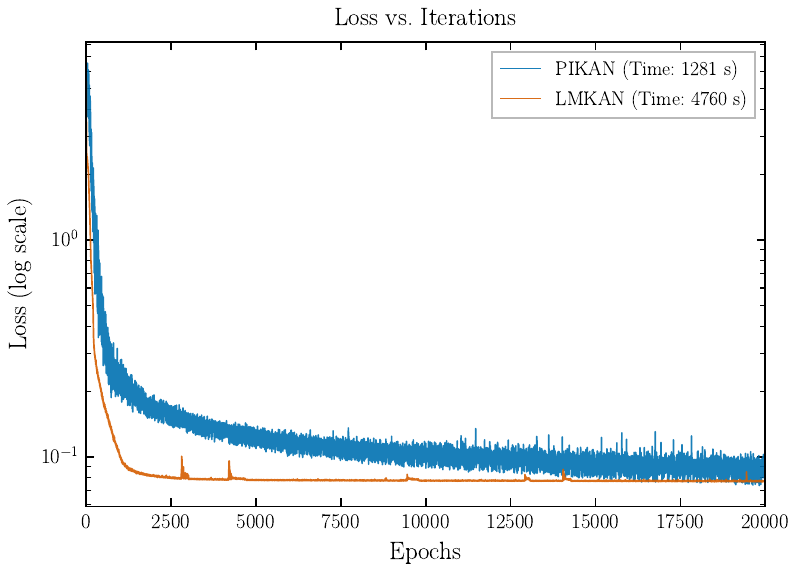}
        
    \end{subfigure}
    \caption{Training loss comparison for Allen--Cahn Case 2.}
    \label{fig:allen_case2_loss_overview}
\end{figure}

The training behavior is summarized in Fig.~\ref{fig:allen_case2_loss_overview}. As shown in Fig.~\ref{fig:allen_case2_loss_overview}, both PIKAN and LM-KAN converge stably during training. Unlike Case~1, however, PIKAN is faster in terms of wall-clock time. In the RBF-based setting, for \(20{,}000\) epochs, PIKAN trains in \(1281.289\)~s, while LM-KAN takes \(4759.98\)~s.  Both models have the same number of trainable parameters, namely \(700\). The final loss values are \(8.606\times10^{-2}\) for PIKAN and \(7.6563\times10^{-2}\) for LM-KAN. A quantitative comparison between the two models is reported in Table~\ref{tab:allen_case2_comparison}. Overall, Case~2 shows that LM-KAN can yield slightly better final accuracy, although this improvement comes with a significantly larger training cost.

\begin{table}[t]
\centering
\caption{Comparison between PIKAN and LM-KAN for Allen--Cahn Case 2 after \(20{,}000\) training epochs. LM-KAN achieves a slightly smaller final loss, while PIKAN is substantially faster to train; both models use the same number of trainable parameters.}
\label{tab:allen_case2_comparison}
\renewcommand{\arraystretch}{1.2}
\begin{tabular}{c|c|c|c}
\hline
Model & Parameters & Training time (s) & Final loss \\
\hline
\textbf{PIKAN}  & 700 & 1281.289 & \(8.606\times10^{-2}\) \\
\textbf{LM-KAN} & 700 & 4759.98 & \(7.6563\times10^{-2}\) \\
\hline
\end{tabular}

\end{table}

Overall, these results indicate that the more challenging Allen--Cahn configuration benefits from the geometry-adaptive representation of LM-KAN. The profile comparisons in Fig.~\ref{fig:allen_case2_profiles}, the spatiotemporal fields in Fig.~\ref{fig:allen_case2_heatmaps}, and the error/residual maps in Fig.~\ref{fig:allen_case2_error_residual} consistently show that LM-KAN provides more accurate solution reconstruction and stronger physics consistency across the domain. The training curves in Fig.~\ref{fig:allen_case2_loss_overview} further show that this improved accuracy is obtained at the expense of substantially higher computational cost, while PIKAN remains the faster alternative.

\subsection{Burgers' Equation}
\label{burger_eqn}

The one-dimensional viscous Burgers' equation is a classical nonlinear advection--diffusion model, originally introduced by Bateman~\cite{Burger0} and later studied by Burgers~\cite{Burger1}. It arises in several areas of applied mathematics and physics, including fluid dynamics, nonlinear wave propagation, and plasma physics~\cite{Burger2,Burger3,Burger4}. From the perspective of scientific machine learning, Burgers' equation is an important benchmark because it combines nonlinear convection and diffusion in a single PDE. The nonlinear term \(u\,u_x\) tends to steepen the solution profile, while the viscous term \(\nu u_{xx}\) smooths it. As a result, the problem tests whether the model can accurately capture transport, dissipation, and evolving localized gradients at the same time.

For LM-KAN, this example is especially useful because it examines whether the learned geometric adaptation can improve representation efficiency in regions where the solution changes more rapidly. In this sense, Burgers' equation tests not only accuracy, but also how well the model allocates resolution across the space--time domain under nonlinear dynamics.

We consider Burgers' equation with the following initial and boundary conditions~\cite{Burgercase12}:
\begin{equation}
\begin{aligned}
u_t(x,t)+u(x,t)u_x(x,t)&=\nu u_{xx}(x,t), \\
u(x,0)&=\sin(\pi x), \\
u(0,t)&=0, \\
u(1,t)&=0,
\end{aligned}
\label{burger}
\end{equation}
where \(\nu=0.1\), \(x\in[0,1]\), and \(t\in[0,1]\).

Let \(\hat{u}(x,t)\) denote the network approximation. The PDE residual is defined by
\begin{equation}
\mathcal{R}_\theta(x,t)=\hat{u}_t+\hat{u}\hat{u}_x-\nu \hat{u}_{xx}.
\end{equation}
Accordingly, the residual, initial-condition, boundary-condition, and data-driven losses are given by
\begin{align}
\mathcal{L}_{\mathrm{r}} &=
\frac{1}{N_r}\sum_{i=1}^{N_r}
\left|\mathcal{R}_\theta(x_r^i,t_r^i)\right|^2, \nonumber\\
\mathcal{L}_{\mathrm{ic}} &=
\frac{1}{N_{\mathrm{ic}}}\sum_{j=1}^{N_{\mathrm{ic}}}
\left(\hat{u}(x_{\mathrm{ic}}^j,0)-\sin(\pi x_{\mathrm{ic}}^j)\right)^2, \nonumber\\
\mathcal{L}_{\mathrm{bc}} &=
\frac{1}{N_{\mathrm{bc}}}\sum_{k=1}^{N_{\mathrm{bc}}}
\left|\hat{u}(0,t_{\mathrm{bc}}^k)\right|^2
+
\frac{1}{N_{\mathrm{bc}}}\sum_{k=1}^{N_{\mathrm{bc}}}
\left|\hat{u}(1,t_{\mathrm{bc}}^k)\right|^2, \nonumber\\
\mathcal{L}_{\mathrm{data}} &=
\frac{1}{N_{\mathrm{data}}}\sum_{p=1}^{N_{\mathrm{data}}}
\left|\hat{u}(x_{\mathrm{data}}^p,t_{\mathrm{data}}^p)-u(x_{\mathrm{data}}^p,t_{\mathrm{data}}^p)\right|^2, \nonumber\\
\mathcal{L}_{\mathrm{final}}^{\mathrm{DD}} &=
\mathcal{L}_{\mathrm{r}}+\mathcal{L}_{\mathrm{ic}}+\mathcal{L}_{\mathrm{bc}}+\mathcal{L}_{\mathrm{data}}.
\label{burger_loss}
\end{align}
Here \(N_r\), \(N_{\mathrm{ic}}\), \(N_{\mathrm{bc}}\), and \(N_{\mathrm{data}}\) denote the numbers of residual, initial-condition, boundary-condition, and supervised data points, respectively.

In this experiment we adopt a data-driven physics-informed strategy. A numerical reference solution is first computed, and a small subset of solution values is incorporated into training through the additional term \(\mathcal{L}_{\mathrm{data}}\). In particular, sharp-feature points corresponding to 10\% of the time grid are sampled and matched to the numerical RK45 solution, thereby providing localized supervision in regions where steep gradients emerge. This hybrid construction is especially useful for Burgers' equation, since the nonlinear advective term can produce rapidly varying transient structures that are often more difficult to capture accurately using a purely residual-based objective. The supervised data help anchor the solution near these sharp regions, while the physics-informed residual enforces consistency with the governing PDE.

For Burgers' equation, the PIKAN baseline uses the architecture $[2,8,4,1]$ with cubic B-spline parameterization on a grid of size 5. In contrast, the LM-KAN model employs two learned-metric basis layers of width 8 with \(K=8\) RBF basis functions per input dimension, \(\gamma=2.5\), and metric hidden width 9 (Table~\ref{tab:burgers_arch}). This larger LM-KAN configuration was chosen deliberately, since Burgers' equation develops sharp spatial gradients and requires higher local representational resolution near the evolving front. Unlike the Allen--Cahn experiments, the Burgers setup does not enforce a tightly matched parameter budget; instead, the LM-KAN architecture is enlarged to test whether geometry-adaptive basis placement is beneficial in the presence of shock-forming dynamics.

\begin{table}
\centering
\caption{Architecture comparison for Burgers' equation. The LM-KAN model is intentionally larger than the PIKAN baseline, using both larger hidden width and more basis functions per input dimension to better resolve steep gradients and shock-like structures.}
\label{tab:burgers_arch}
\renewcommand{\arraystretch}{1.2}
\small
\begin{tabularx}{\textwidth}{|c|X|c|c|X|c|c|}
\hline
Model & Architecture & Depth & Resolution & Basis & Metric hidden & Output \\
\hline
PIKAN & $[2,8,4,1]$ & 2 & Grid $=5$ & Cubic B-spline + $\sin$ & -- & 1 \\
\hline
LM-KAN & $(2 \rightarrow 8 \rightarrow 8 \rightarrow 1)$ & 2 & $K=8$ & RBF, $\gamma=2.5$ & 9 & 1 \\
\hline
\end{tabularx}

\end{table}

\begin{figure}
    \centering
    \begin{subfigure}[t]{0.48\textwidth}
        \centering
        \includegraphics[width=\textwidth]{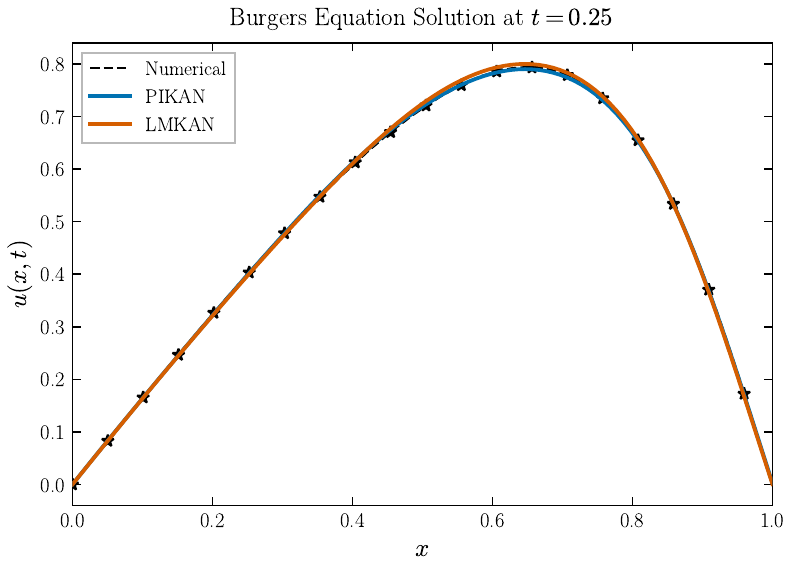}
        \caption{\(t=0.25\)}
    \end{subfigure}
    \hfill
    \begin{subfigure}[t]{0.48\textwidth}
        \centering
        \includegraphics[width=\textwidth]{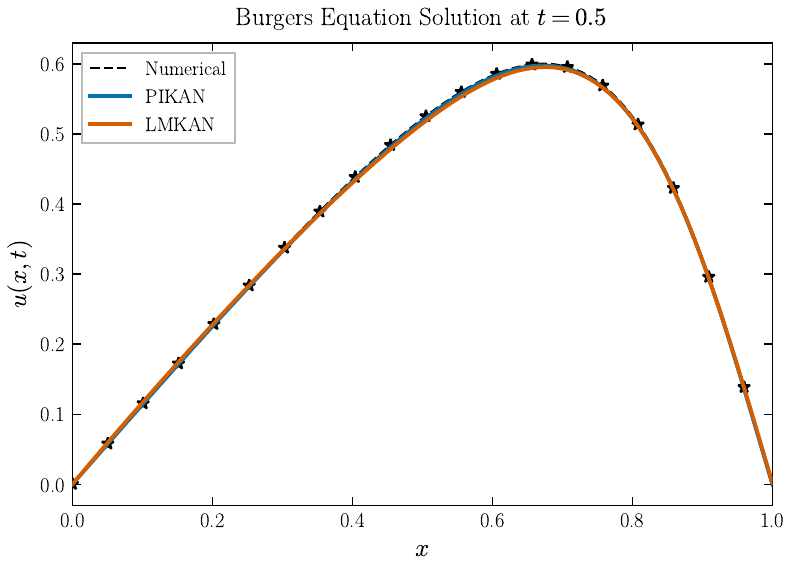}
        \caption{\(t=0.5\)}
    \end{subfigure}

    \vspace{0.3cm}

    \begin{subfigure}[t]{0.48\textwidth}
        \centering
        \includegraphics[width=\textwidth]{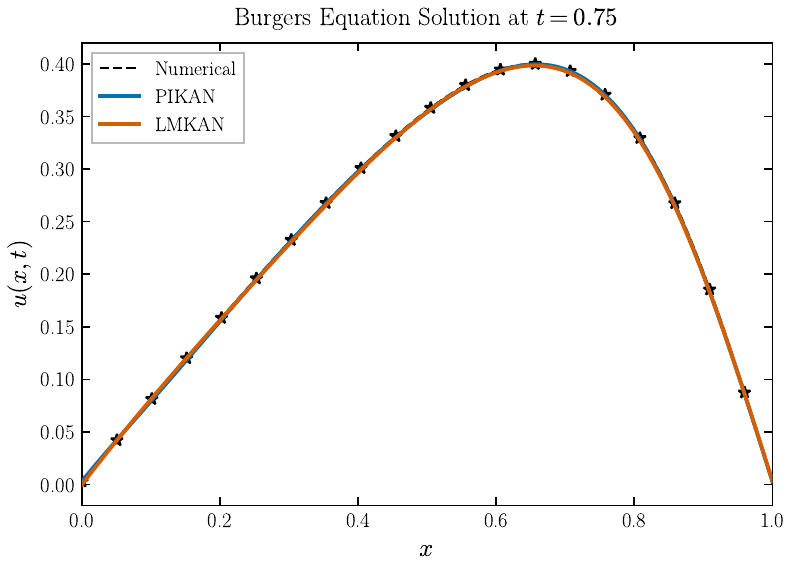}
        \caption{\(t=0.75\)}
    \end{subfigure}
    \hfill
    \begin{subfigure}[t]{0.48\textwidth}
        \centering
        \includegraphics[width=\textwidth]{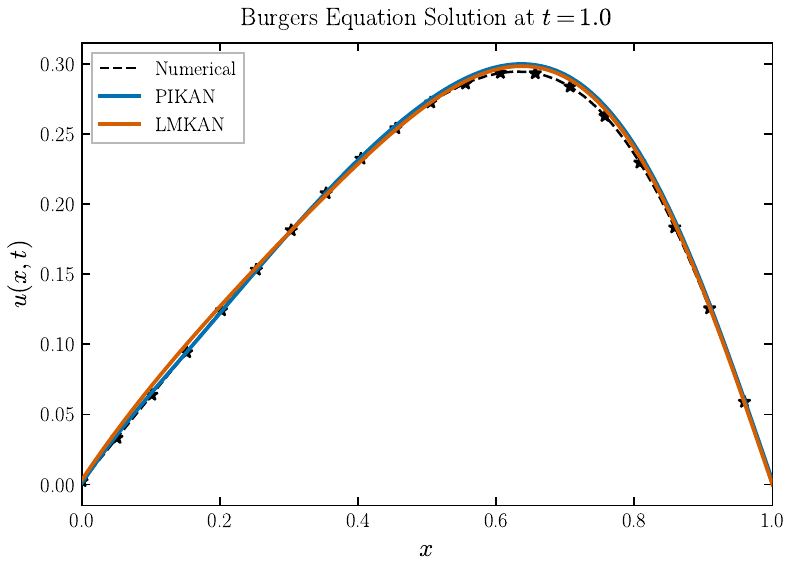}
        \caption{\(t=1.0\)}
    \end{subfigure}
    \caption{Comparison of the numerical solution with the PIKAN and LM-KAN predictions for Burgers' equation at four representative time instances. Both models remain in close agreement with the numerical reference over the full time interval, showing that they accurately capture the nonlinear advection--diffusion evolution.}
    \label{fig:burger_profiles}
\end{figure}

\begin{figure}
    \centering
    \begin{subfigure}[t]{0.7\textwidth}
        \centering
        \includegraphics[width=\textwidth]{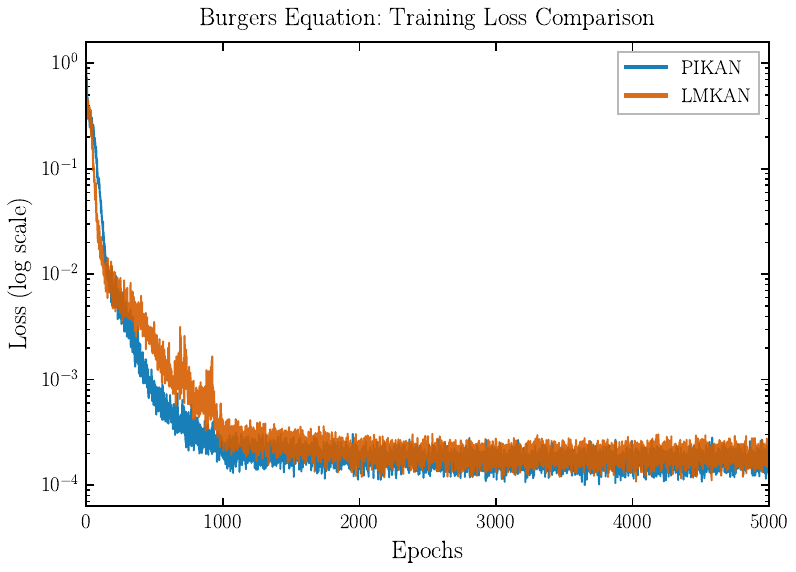}
        \caption{Training loss comparison between PIKAN and LM-KAN.}
        \label{fig:burger_loss_compare}
    \end{subfigure}
    \caption{Training behavior for Burgers' equation. Both models exhibit rapid early-stage loss decay and then stabilize in the \(10^{-4}\) regime. PIKAN reaches a slightly smaller final loss, while LM-KAN attains comparable accuracy with lower wall-clock training time.}
    \label{fig:burger_training}
\end{figure}

\begin{table}
\centering
\caption{Comparison between PIKAN and LM-KAN for Burgers' equation after \(5000\) training epochs. PIKAN attains a slightly smaller final loss, while LM-KAN completes training in less wall-clock time despite using more trainable parameters.}
\label{tab:burger_pikan_lmkan}
\renewcommand{\arraystretch}{1.2}
\begin{tabular}{c|c|c|c}
\hline
Model & Parameters & Training time (s) & Final loss \\
\hline
\textbf{PIKAN}  & 520  & 295.39 & 0.000099 \\
\textbf{LM-KAN} & 1229 & 206.48 & 0.000103 \\
\hline
\end{tabular}

\end{table}

To first assess pointwise predictive accuracy, Fig.~\ref{fig:burger_profiles} compares the numerical and predicted solution profiles at \(t=0.25\), \(0.5\), \(0.75\), and \(1.0\). At all four time instants, both PIKAN and LM-KAN closely follow the numerical solution, indicating that each model successfully captures the nonlinear advection--diffusion dynamics. The agreement remains strong not only at earlier times, when the solution amplitude is larger and gradients are more pronounced, but also at later times, when viscosity progressively smooths the solution profile. These profile comparisons confirm that both models correctly learn the temporal evolution of Burgers' dynamics.

At the end of training, the two models attain nearly the same final loss: \(9.9\times10^{-5}\) for PIKAN and \(1.03\times10^{-4}\) for LM-KAN. However, LM-KAN achieves this with a noticeably smaller wall-clock training time, requiring \(206.48\) seconds compared with \(295.39\) seconds for PIKAN, despite having more trainable parameters (1229 for LM-KAN versus 520 for PIKAN). A quantitative comparison between the two models is reported in Table~\ref{tab:burger_pikan_lmkan}. This suggests that the learned-metric representation improves computational efficiency while maintaining essentially the same predictive quality.

Fig. ~\ref{fig:burger_training} shows the evolution of the training loss for the two models. Both curves decrease sharply during the early epochs and then flatten into a similar convergence regime. Although PIKAN ends with a marginally smaller loss value, the difference is very small, and both models exhibit stable optimization behavior throughout training. When viewed together with the training times, this result indicates that LM-KAN reaches essentially the same accuracy level more efficiently.

\begin{figure}
    \centering
    \begin{subfigure}[t]{1.0\textwidth}
        \centering
        \includegraphics[width=\textwidth]{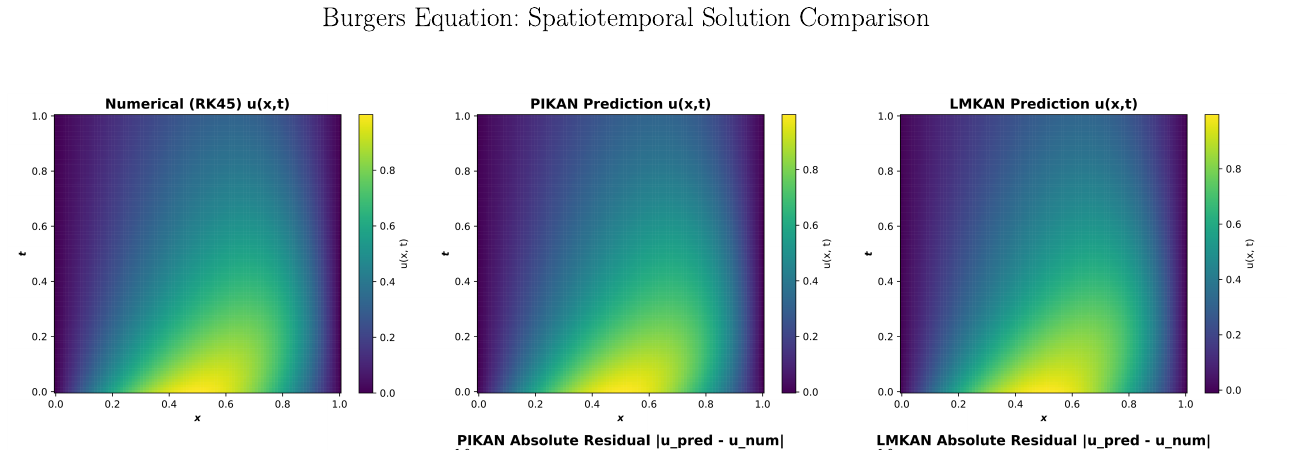}
        \caption{Spatiotemporal solution comparison among the numerical reference, PIKAN, and LM-KAN.}
    \end{subfigure}

    \vspace{0.3cm}

    \begin{subfigure}[t]{0.8\textwidth}
        \centering
        \includegraphics[width=\textwidth]{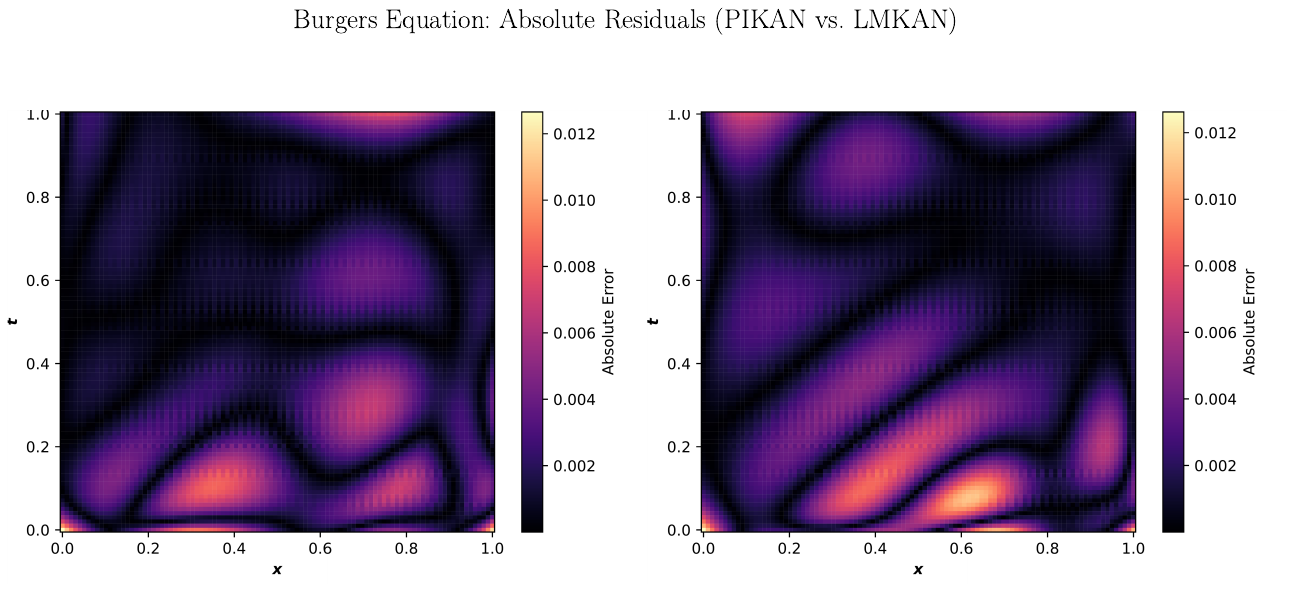}
        \caption{Absolute error maps \(|u_{\mathrm{pred}}-u_{\mathrm{num}}|\) for PIKAN and LM-KAN.}
    \end{subfigure}
    \caption{Global space--time comparison for Burgers' equation. The top panel shows that both learned models reproduce the overall transport--diffusion pattern of the numerical solution very accurately. The bottom panel shows that the largest errors are localized mainly near early times and in regions with stronger spatial variation, while the error remains small over most of the domain for both methods.}
    \label{fig:burger_heatmap_residual}
\end{figure}

A more global picture is provided in Fig.~\ref{fig:burger_heatmap_residual}. The spatiotemporal heatmaps in Fig.~\ref{fig:burger_heatmap_residual}(a) show that both PIKAN and LM-KAN reproduce the full solution field very closely, including the gradual decay and redistribution induced by the balance between nonlinear transport and viscosity. The absolute error maps in Fig.~\ref{fig:burger_heatmap_residual}(b) further show that the largest discrepancies are confined to localized regions, mainly near early times and near sharper transient structures. Outside these regions, both models maintain very small errors across most of the domain.

Overall, Burgers' equation provides a meaningful benchmark for GeoKAN-type models because it tests nonlinear transport, diffusion balance, and localized gradient resolution simultaneously. The profile comparisons in Fig.~\ref{fig:burger_profiles}, the training behavior in Fig.~\ref{fig:burger_training}, and the global space--time results in Fig.~\ref{fig:burger_heatmap_residual} all show that LM-KAN preserves the predictive quality of PIKAN while reducing training time. This supports the usefulness of the learned geometric adaptation for nonlinear PDE learning.

\subsection{The Lorenz System}
\label{Lorentz_System}

The Lorenz system is a classical nonlinear system of ordinary differential equations originally introduced as a simplified model for atmospheric convection \cite{Lorenz2}. It is one of the most widely studied benchmark problems in nonlinear dynamics, owing to its rich transient behavior and strong coupling among state variables. In the present work, we consider a non-chaotic regime of the Lorenz system and use it to compare the original physics-informed KAN (PIKAN) with the geometry-adaptive LM-KAN model.

The governing equations are given by \cite{Lorenz1}
\begin{equation}
\begin{aligned}
x'(t) &= \sigma\bigl(y(t)-x(t)\bigr), \\
y'(t) &= x(t)\bigl(\rho-z(t)\bigr)-y(t), \\
z'(t) &= x(t)y(t)-\beta z(t),
\end{aligned}
\qquad
\begin{cases}
x(0)=1,\\
y(0)=1,\\
z(0)=1,
\end{cases}
\label{lorentz_equation}
\end{equation}
where \(t\in[0,20]\).

Let \(\hat{x}(t)\), \(\hat{y}(t)\), and \(\hat{z}(t)\) denote the learned solution components. The corresponding residuals are defined as
\begin{equation}
\begin{aligned}
\mathcal{R}_{\theta}^{1}(t) &= \hat{x}'(t)-\sigma\bigl(\hat{y}(t)-\hat{x}(t)\bigr), \\
\mathcal{R}_{\theta}^{2}(t) &= \hat{y}'(t)-\bigl(\hat{x}(t)(\rho-\hat{z}(t))-\hat{y}(t)\bigr), \\
\mathcal{R}_{\theta}^{3}(t) &= \hat{z}'(t)-\bigl(\hat{x}(t)\hat{y}(t)-\beta\hat{z}(t)\bigr).
\end{aligned}
\end{equation}
The residual and initial-condition losses are then given by
\begin{equation}
\begin{aligned}
\mathcal{L}_{\mathrm{r}}
&= \frac{1}{N_r}\sum_{i=1}^{N_r}
\left(
\left|\mathcal{R}_{\theta}^{1}(t_r^i)\right|^2
+
\left|\mathcal{R}_{\theta}^{2}(t_r^i)\right|^2
+
\left|\mathcal{R}_{\theta}^{3}(t_r^i)\right|^2
\right), \\
\mathcal{L}_{\mathrm{ic}}
&= (\hat{x}(0)-1)^2+(\hat{y}(0)-1)^2+(\hat{z}(0)-1)^2,
\end{aligned}
\end{equation}
and the final loss function is
\begin{equation}
\mathcal{L}_{\mathrm{final}}^{\mathrm{DF}}=\mathcal{L}_{\mathrm{r}}+\mathcal{L}_{\mathrm{ic}}.
\label{Lorentz_loss}
\end{equation}

For the Lorenz system, in the architecture the input is the scalar time variable \(t\), while the output is the three-dimensional state vector \((x(t),y(t),z(t))\). The PIKAN baseline is implemented as a wavelet-based KAN with architecture $[1,8,16,3]$. The LM-KAN architecture uses three learned-metric basis layers of width 4 with \(K=5\) RBF basis functions per layer and metric hidden width 9, followed by a linear head producing the three state components simultaneously (Table~\ref{tab:lorenz_arch}). Here, the RBF parameter is reduced to \(\gamma=0.5\), yielding smoother and wider kernels that are better suited to long-horizon trajectory learning.

\begin{table}
\centering
\caption{Architecture comparison for the Lorenz system. Since the task maps a scalar time input to the three coupled state variables \((x,y,z)\), the output dimension is 3 for all models.}
\label{tab:lorenz_arch}
\renewcommand{\arraystretch}{1.2}
\small
\begin{tabularx}{\textwidth}{|c|X|c|c|X|c|c|}
\hline
Model & Architecture & Depth & Resolution & Basis & Metric hidden & Output \\
\hline
PIKAN (Wav-KAN) & $[1,8,16,3]$ & 2 & -- & Wavelet basis & -- & 3 \\
\hline
LM-KAN & $(1 \rightarrow 4 \rightarrow 4 \rightarrow 4 \rightarrow 3)$ & 3 & $K=5$ & RBF, $\gamma=0.5$ & 9 & 3 \\
\hline
\end{tabularx}

\end{table}

\begin{figure}
    \centering
    \begin{subfigure}[t]{0.92\textwidth}
        \centering
        \includegraphics[width=\textwidth]{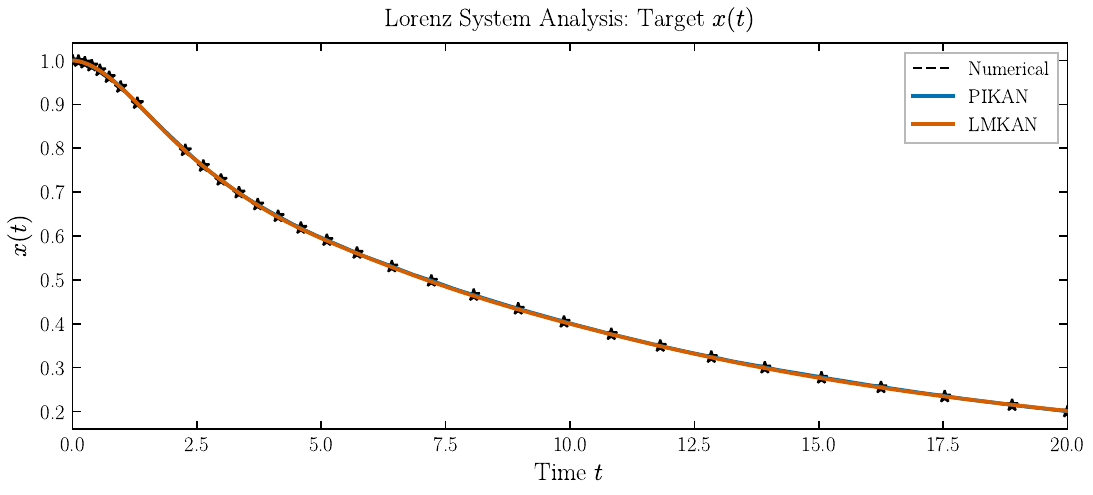}
        \caption{Comparison for the \(x(t)\) trajectory.}
        \label{fig:lorenz_xt}
    \end{subfigure}
    \hfill
    \begin{subfigure}[t]{0.92\textwidth}
        \centering
        \includegraphics[width=\textwidth]{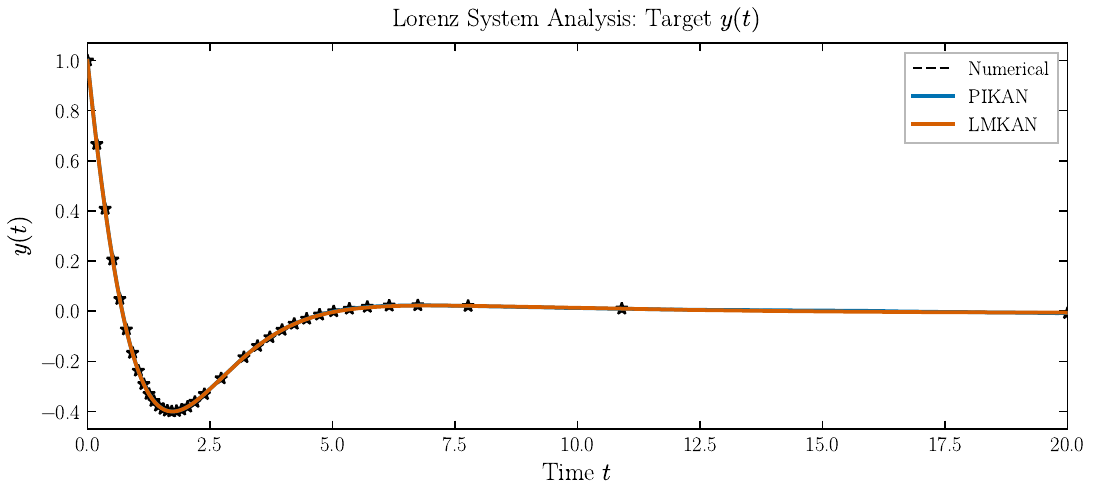}
        \caption{Comparison for the \(y(t)\) trajectory.}
        \label{fig:lorenz_yt}
    \end{subfigure}

    \vspace{0.3cm}

    \begin{subfigure}[t]{0.92\textwidth}
        \centering
        \includegraphics[width=\textwidth]{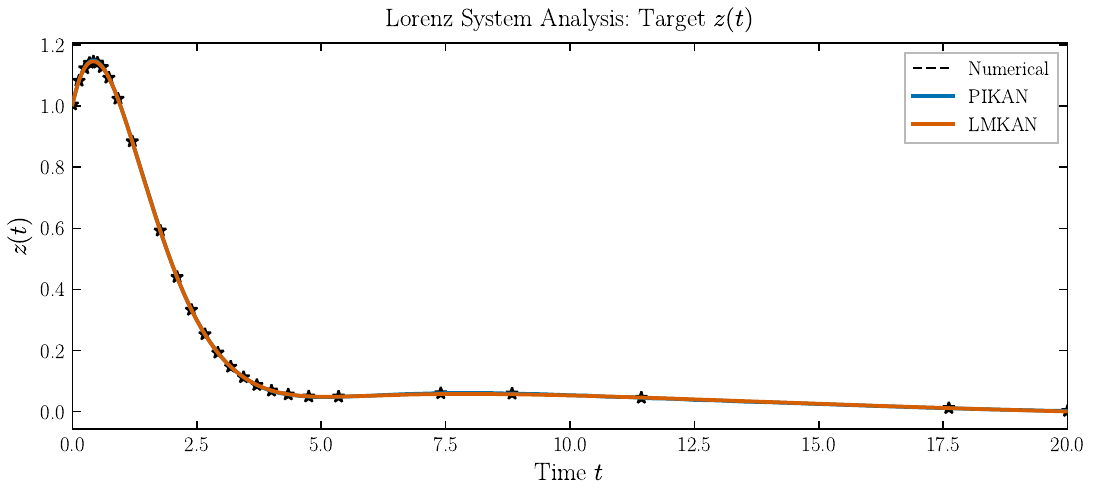}
        \caption{Comparison for the \(z(t)\) trajectory.}
        \label{fig:lorenz_zt}
    \end{subfigure}
    \caption{Solution comparison for the Lorenz system. Both PIKAN and LM-KAN closely follow the numerical solution for all three state variables over \(0\le t\le 20\).}
    \label{fig:lorenz_trajectories}
\end{figure}

 We use \(N_r=100\) collocation points uniformly distributed over \(0\le t\le 20\). Both models are trained for \(10^4\) epochs using the AdamW optimizer with learning rate \(\eta=10^{-3}\). The corresponding solution trajectories are shown first in Fig.~\ref{fig:lorenz_trajectories}. For all three state variables, both learned models closely match the numerical reference solution over the full time interval. The \(x(t)\) trajectory in Fig.~\ref{fig:lorenz_xt} is captured accurately by both approaches, with the predicted curves nearly overlapping the numerical solution. A similarly close agreement is observed for \(y(t)\) in Fig.~\ref{fig:lorenz_yt}, including its initial transient and subsequent smooth evolution. The \(z(t)\) trajectory in Fig.~\ref{fig:lorenz_zt} is also well resolved, showing that both models successfully learn the coupled system dynamics. Overall, the trajectory plots indicate that both PIKAN and LM-KAN provide accurate solution approximations throughout the interval \(0\le t\le 20\).

\begin{figure}
    \centering
    \begin{subfigure}[t]{0.72\textwidth}
        \centering
        \includegraphics[width=\textwidth]{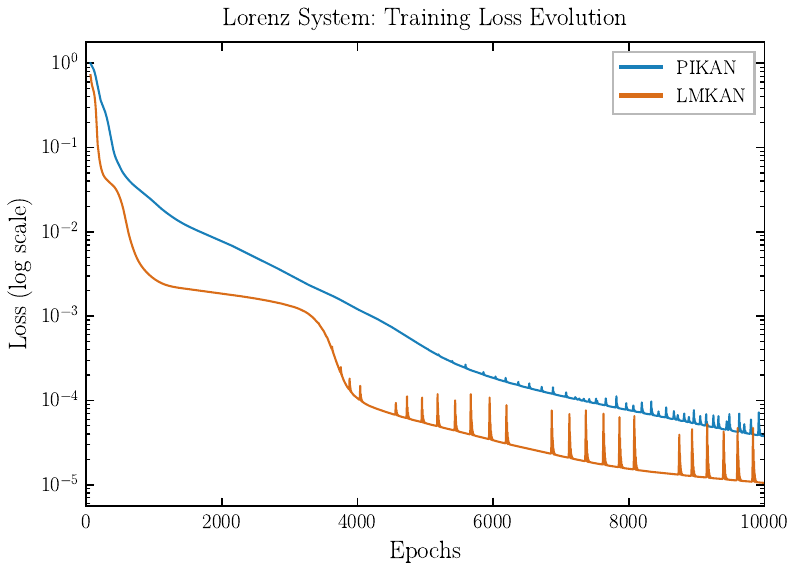}
        \label{fig:lorenz_loss_compare_sub}
    \end{subfigure}
    \caption{Training loss comparison between PIKAN and LM-KAN for the Lorenz system. Both models exhibit steady loss decay over training, while LM-KAN reaches a lower final loss than PIKAN after \(10^4\) epochs.}
    \label{fig:lorenz_loss_compare}
\end{figure}

The training-loss evolution is shown next in Fig.~\ref{fig:lorenz_loss_compare}. Both models achieve stable optimization and reduce the loss by several orders of magnitude. However, LM-KAN converges to a slightly smaller final loss than PIKAN. At the end of training, the final loss is \(2.7\times 10^{-5}\) for PIKAN and \(5\times 10^{-6}\) for LM-KAN, confirming the stronger final fit achieved by the geometry-adaptive model.

A quantitative comparison between the two models is reported in Table~\ref{tab:lorenz_comparison}. Although LM-KAN requires a longer training time, it achieves the smallest final loss while keeping the parameter count close to that of PIKAN. Specifically, PIKAN uses 736 trainable parameters and completes training in 232.46 seconds, whereas LM-KAN uses 777 trainable parameters and requires 527.37 seconds. Thus, for this Lorenz-system experiment, LM-KAN provides improved final accuracy at the expense of higher computational cost.

\begin{table}
\centering
\caption{Comparison between PIKAN and LM-KAN for the Lorenz system after \(10^4\) training epochs. LM-KAN attains a lower final loss, while PIKAN remains more computationally efficient.}
\renewcommand{\arraystretch}{1.2}
\begin{tabular}{c|c|c|c}
\hline
Model & Parameters & Training time (s) & Final loss \\
\hline
\textbf{PIKAN}  & 736 & 232.46 & 0.000027 \\
\textbf{LM-KAN} & 777 & 527.37 & 0.000005 \\
\hline
\end{tabular}
\label{tab:lorenz_comparison}
\end{table}

\subsection{Helmholtz Equation}
\label{helmholtz_eqn}

The one-dimensional Helmholtz equation is a canonical frequency-domain model for wave propagation in inhomogeneous media. In the present work, it serves as a demanding benchmark because the solution is complex-valued and combines rapid phase oscillations with spatially varying amplitude. As the wavelength decreases, the effective wavenumber increases, and the field becomes progressively more oscillatory. This makes the Helmholtz problem a useful test of whether the model can preserve phase accuracy, amplitude transitions, and global wave geometry under increasing representational difficulty. 

Following the structure-preserving Helmholtz formulation in~\cite{Khairullin2026HelmholtzPINN}, we consider a one-dimensional monochromatic scattering problem for a complex-valued field \(u(z)\) on the interval \(z\in[0,1]\):
\begin{equation}
\begin{aligned}
u_{zz}(z)+k^2(z)u(z)&=0, \\
\left(u_z+ik_- \, u\right)\big|_{z=0}&=2ik_- \, A_{\mathrm{inc}}, \\
\left(u_z-ik_+ \, u\right)\big|_{z=1}&=0,
\end{aligned}
\label{helmholtz}
\end{equation}
where \(A_{\mathrm{inc}}\) denotes the incident-wave amplitude, and \(k_-\) and \(k_+\) are the exterior wavenumbers on the left and right boundaries, respectively. The spatially varying wavenumber is related to the complex relative permittivity \(\epsilon(z)\) through
\begin{equation}
k^2(z)=\epsilon(z)\left(\frac{2\pi}{\lambda}\right)^2,
\end{equation}
where \(\lambda\) is the driving wavelength. Since \(\epsilon(z)\) may be complex-valued, the solution exhibits both oscillatory and attenuative behavior, which makes the learning problem substantially more challenging than standard real-valued smooth benchmarks. 

Let \(\hat{u}(z)\) denote the network approximation. The Helmholtz residual is defined by
\begin{equation}
\mathcal{R}_\theta(z)=\hat{u}_{zz}(z)+k^2(z)\hat{u}(z).
\end{equation}
Accordingly, the residual and boundary-condition losses are given by
\begin{align}
\mathcal{L}_{\mathrm{r}} &=
\frac{1}{N_r}\sum_{i=1}^{N_r}
\left|\mathcal{R}_\theta(z_r^i)\right|^2, \nonumber\\
\mathcal{L}_{\mathrm{bc}} &=
\frac{1}{N_{\mathrm{bc}}}\sum_{j=1}^{N_{\mathrm{bc}}}
\left|
\hat{u}_z(0)+ik_-\hat{u}(0)-2ik_-A_{\mathrm{inc}}
\right|^2 \nonumber\\
&\quad+
\frac{1}{N_{\mathrm{bc}}}\sum_{j=1}^{N_{\mathrm{bc}}}
\left|
\hat{u}_z(1)-ik_+\hat{u}(1)
\right|^2.
\end{align}
If reference solution samples are additionally used, the data-driven loss is written as
\begin{align}
\mathcal{L}_{\mathrm{data}} &=
\frac{1}{N_{\mathrm{data}}}\sum_{p=1}^{N_{\mathrm{data}}}
\left|
\hat{u}(z_{\mathrm{data}}^p)-u(z_{\mathrm{data}}^p)
\right|^2,
\end{align}
and the final training objective becomes
\begin{equation}
\mathcal{L}_{\mathrm{final}}^{\mathrm{DD}}
=
\mathcal{L}_{\mathrm{r}}+\mathcal{L}_{\mathrm{bc}}+\mathcal{L}_{\mathrm{data}},
\label{helmholtz_loss}
\end{equation}
where \(N_r\), \(N_{\mathrm{bc}}\), and \(N_{\mathrm{data}}\) denote the numbers of residual, boundary, and supervised data points, respectively. This is the direct Helmholtz-PINN formulation; the reference paper notes that such second-order residuals become increasingly stiff in high-frequency regimes, whereas reduced first-order formulations can substantially improve optimization behavior. 

For the Helmholtz equation, we compare the PIKAN baseline implemented with EfficientKAN against the LM-KAN model under an approximately matched parameter budget. Both models are trained over the same wavelength interval,
\[
\lambda \in \left[\frac{1}{30}, \frac{1}{10}\right],
\]
using the same residual-based training objective. This ensures that the comparison reflects differences in representation and inductive bias rather than differences in overall model size.

The PIKAN baseline uses an EfficientKAN architecture of the form $[2,12,12,2]$ with cubic B-spline parameterization on a grid of size 5. In contrast, the LM-KAN model employs a learned-metric architecture with input dimension 2, output dimension 2, hidden width 4, and depth 2. For this oscillatory problem, we use the Fourier-based LM-KAN variant (Eq. \eqref{eq:lm_forward_expanded_fourier}), since its post-warp sinusoidal dictionary is naturally aligned with the wave-like structure of the Helmholtz solution. It uses \(K=16\) Fourier basis functions per input dimension, \(\gamma=0.5\), and a metric hidden width of 18. As shown in Table~\ref{tab:helmholtz_arch}, the two models have nearly identical parameter counts, with 1920 parameters for PIKAN and 1928 for LM-KAN. Hence, any performance difference in this benchmark cannot be attributed simply to model scale. This matched-capacity setup is particularly important for the Helmholtz problem, since the benchmark is designed to test how well each model handles oscillatory complex-valued wavefields over multiple wavelengths. In this setting, LM-KAN is not given a larger budget; instead, it is evaluated at essentially the same parameter scale as the EfficientKAN-based PIKAN baseline, so that the effect of learned geometric adaptation can be assessed directly.

\begin{table}
\centering
\caption{Architecture comparison for the Helmholtz equation. The PIKAN and LM-KAN models are matched to nearly the same parameter count so that differences in performance reflect architectural bias rather than model size.}
\label{tab:helmholtz_arch}
\renewcommand{\arraystretch}{1.2}
\small
\begin{tabularx}{\textwidth}{|c|X|c|c|X|c|c|}
\hline
Model & Architecture & Depth & Resolution & Basis & Metric hidden & Output \\
\hline
PIKAN & $[2,12,12,2]$ & 2 & Grid $=5$ & Cubic B-spline & -- & 2 \\
\hline
LM-KAN & $(2 \rightarrow 4 \rightarrow 4 \rightarrow 2)$ & 2 & $K=16$ & Fourier, $\gamma=0.5$ & 18 & 2 \\
\hline
\end{tabularx}
\end{table}

\begin{figure}
    \centering
    \includegraphics[width=0.98\textwidth]{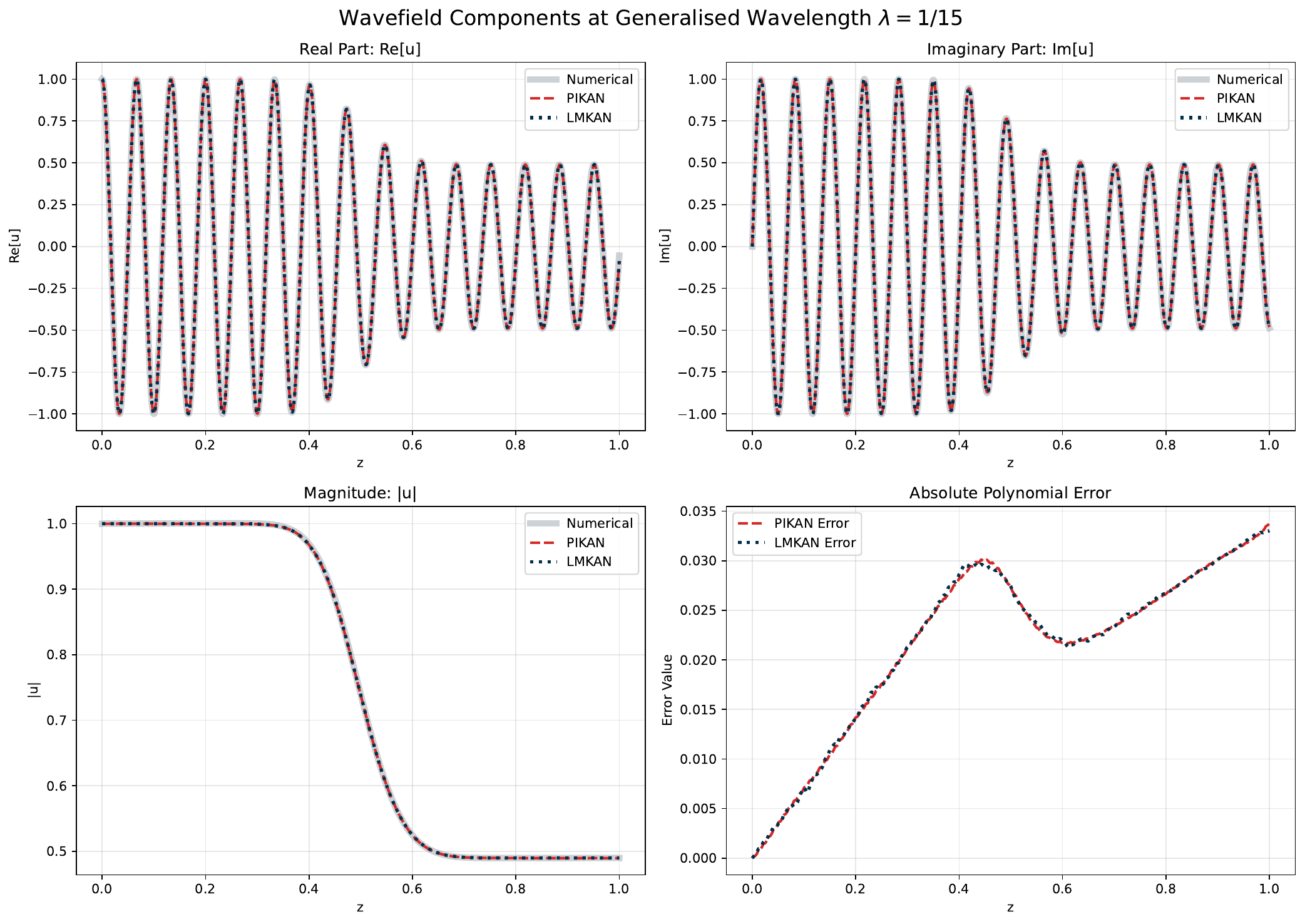}
    \caption{Wavefield components for the Helmholtz benchmark at \(\lambda=1/15\). The figure shows \(\mathrm{Re}[u]\), \(\mathrm{Im}[u]\), \(|u|\), and the absolute error for the reference solution, PIKAN, and LM-KAN. At this wavelength, both models closely reproduce the reference field, while LM-KAN remains slightly more tightly aligned across the oscillatory components.}
    \label{fig:helmholtz_wavefield_115}
\end{figure}

For the present comparison, we evaluate PIKAN and LM-KAN at three wavelengths, \(\lambda=1/15\), \(1/20\), and \(1/25\). These cases provide a controlled progression from moderately oscillatory to strongly oscillatory regimes. In all cases, the reference solution is compared against the learned real part, imaginary part, magnitude, spatiotemporal field structure, and phase-space orbit geometry. This allows us to assess not only pointwise agreement, but also whether the learned representation preserves the qualitative structure of the complex wavefield under wavelength tension. The original Helmholtz study similarly emphasizes that performance differences become more pronounced as the oscillatory burden increases. 
The one-dimensional wavefield comparisons provide the clearest view of model accuracy across increasing oscillatory difficulty. Fig. ~\ref{fig:helmholtz_wavefield_115} corresponds to the case \(\lambda=1/15\), where both models reproduce the reference solution closely in the real and imaginary parts, and both capture the smooth magnitude transition accurately. In this relatively easier regime, the difference between the methods is modest, although LM-KAN remains slightly more tightly aligned with the reference in the oscillatory components.

\begin{figure}
    \centering
    \includegraphics[width=0.98\textwidth]{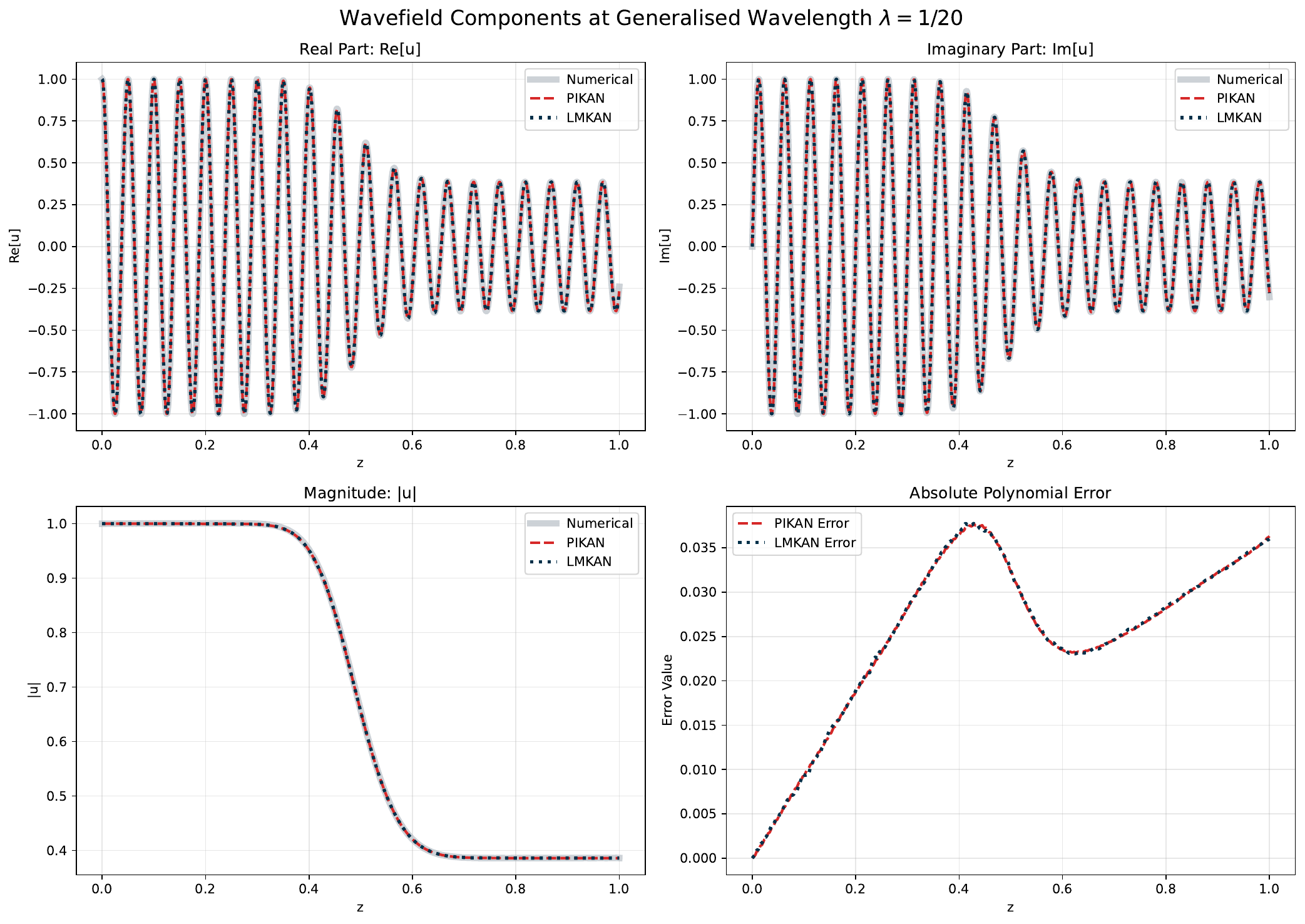}
    \caption{Wavefield components for the Helmholtz benchmark at \(\lambda=1/20\). As the wavelength decreases, the oscillations become denser and the approximation task becomes more sensitive to phase mismatch. Both models continue to track the reference well, but LM-KAN preserves the wavefield structure slightly more accurately through the transition region and in the error profile.}
    \label{fig:helmholtz_wavefield_120}
\end{figure}

As the wavelength decreases to \(\lambda=1/20\), shown in Fig. ~\ref{fig:helmholtz_wavefield_120}, the oscillations become denser and the approximation becomes more sensitive to phase mismatch. Both models still track the reference well, but the error level increases relative to the \(\lambda=1/15\) case. In particular, the transition region near the middle of the domain becomes a stricter test because the model must preserve both oscillatory phase and amplitude-envelope variation. LM-KAN continues to maintain slightly better alignment with the reference across \(\mathrm{Re}[u]\), \(\mathrm{Im}[u]\), and \(|u|\).

\begin{figure}
    \centering
    \includegraphics[width=0.98\textwidth]{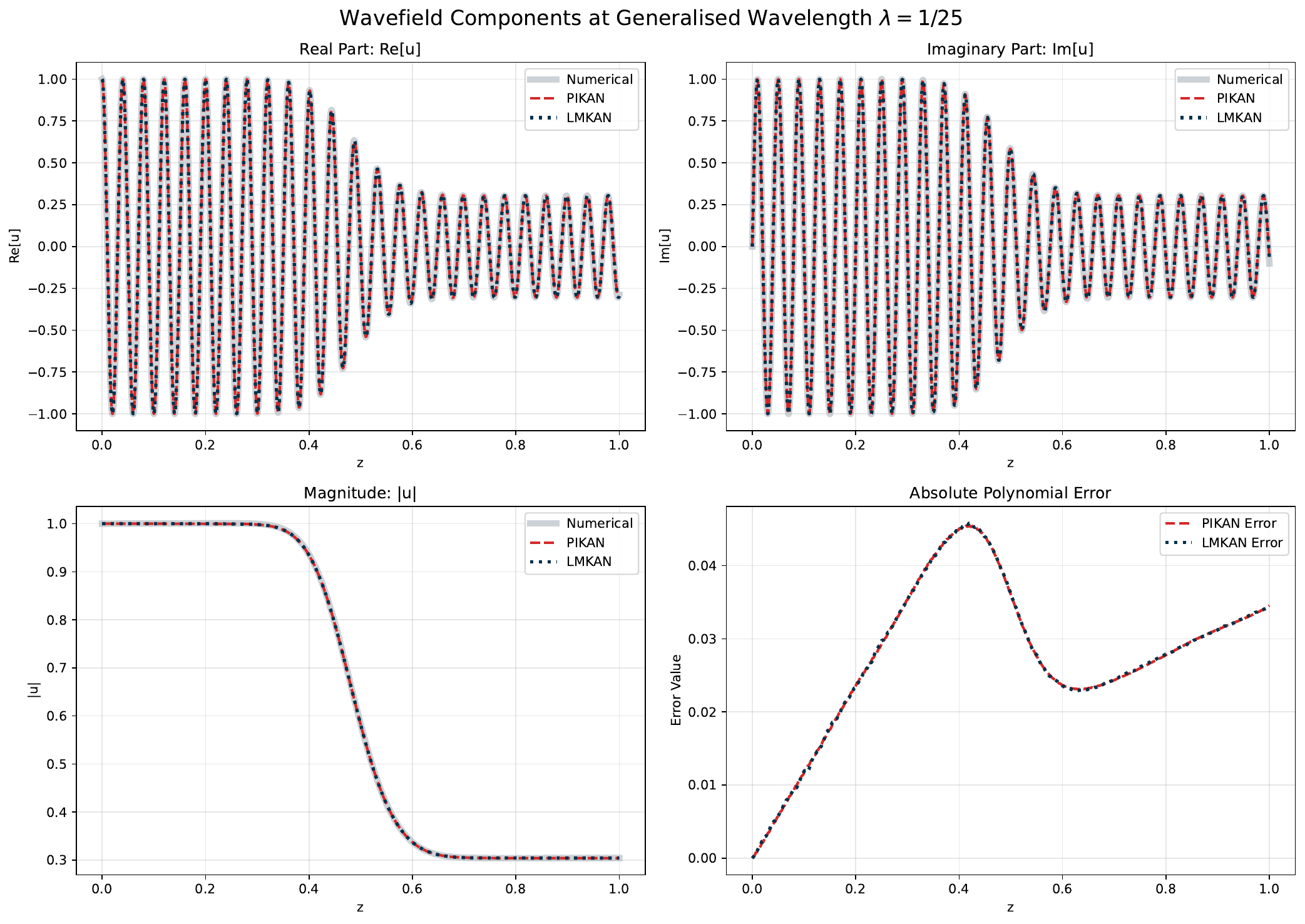}
    \caption{Wavefield components for the Helmholtz benchmark at \(\lambda=1/25\). This is the most oscillatory case considered. Although both models retain strong qualitative agreement with the reference solution, LM-KAN better preserves the oscillatory structure and remains marginally closer in the absolute-error distribution, especially near the main transition region.}
    \label{fig:helmholtz_wavefield_125}
\end{figure}

The hardest case, \(\lambda=1/25\), is shown in Fig. ~\ref{fig:helmholtz_wavefield_125}. Here the field contains the highest oscillation density among the tested cases, and the absolute error becomes largest overall. Even in this regime, both models preserve the qualitative structure of the reference solution, but LM-KAN remains marginally closer in the oscillatory components and through the downstream tail. This trend supports the view that the learned geometric adaptation becomes more useful as the solution develops finer-scale oscillatory structure.

\begin{figure}
    \centering
    \begin{subfigure}[t]{0.98\textwidth}
        \centering
        \includegraphics[width=\textwidth]{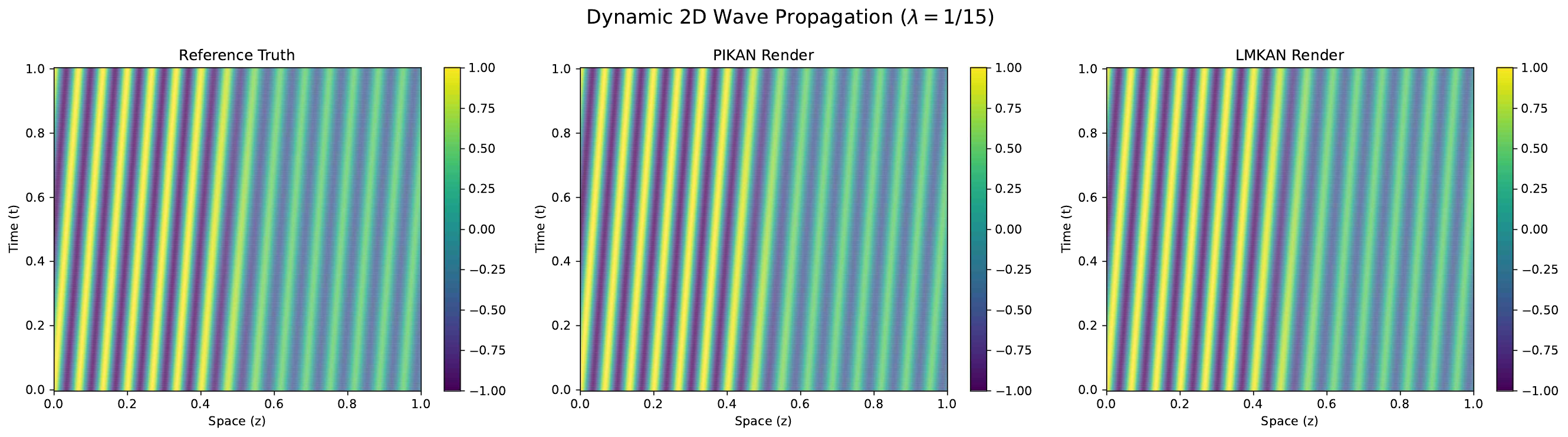}
        \caption{\(\lambda=1/15\)}
        \label{fig:helmholtz_spacetime_115}
    \end{subfigure}
    \hfill
    \begin{subfigure}[t]{0.98\textwidth}
        \centering
        \includegraphics[width=\textwidth]{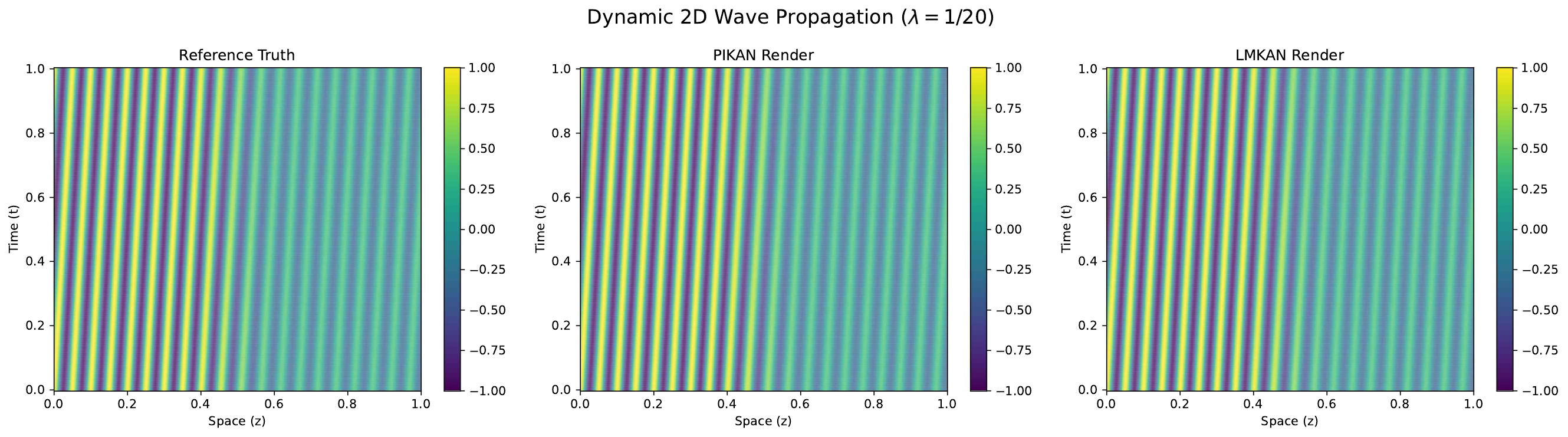}
        \caption{\(\lambda=1/20\)}
        \label{fig:helmholtz_spacetime_120}
    \end{subfigure}
    \hfill
    \begin{subfigure}[t]{0.98\textwidth}
        \centering
        \includegraphics[width=\textwidth]{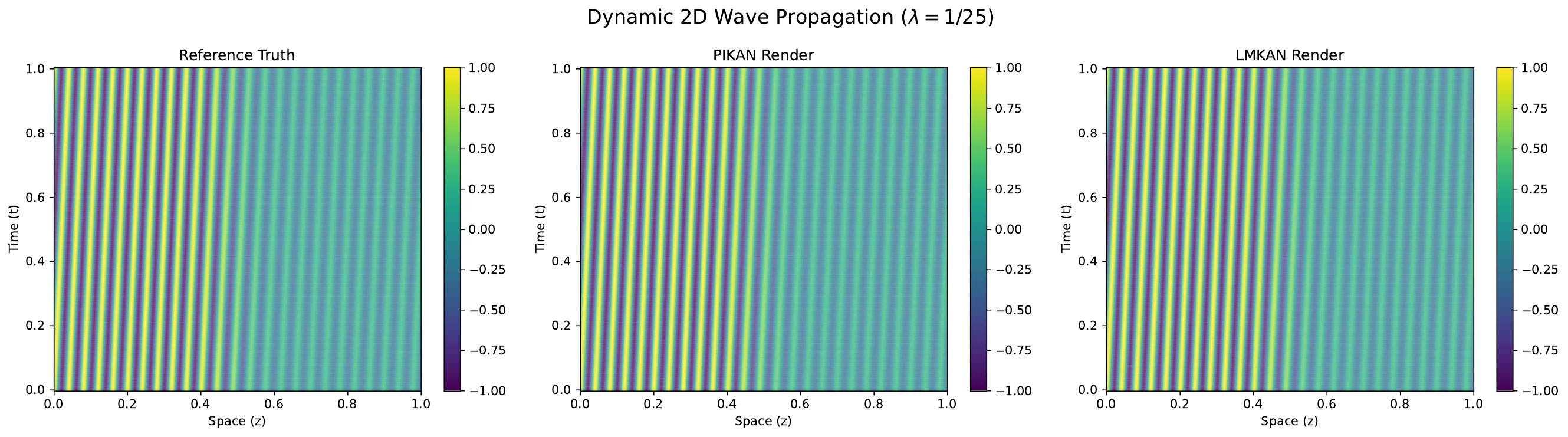}
        \caption{\(\lambda=1/25\)}
        \label{fig:helmholtz_spacetime_125}
    \end{subfigure}
    \caption{Spatiotemporal rendering of the Helmholtz wavefield for three wavelengths. Each panel compares the reference field with the EfficientKAN and LM-KAN reconstructions. The stripe pattern becomes progressively denser as \(\lambda\) decreases, revealing the increasing oscillatory complexity of the benchmark.}
    \label{fig:helmholtz_spacetime_all}
\end{figure}

The same trend appears in the spatiotemporal visualizations of Fig. ~\ref{fig:helmholtz_spacetime_all}. For \(\lambda=1/15\), the stripe pattern is relatively well resolved by both methods, and the rendered fields appear visually close to the reference. At \(\lambda=1/20\), the frequency of the stripe pattern increases, making any local phase drift more noticeable. At \(\lambda=1/25\), the problem becomes clearly high-frequency, and the benefit of LM-KAN is more evident: its rendering remains closer to the reference stripe geometry and contrast pattern, while EfficientKAN exhibits slightly more visible deviation in the fine oscillatory structure. These spatiotemporal plots therefore support the one-dimensional error analysis by showing that LM-KAN preserves the global propagation texture more faithfully as the wavelength decreases.

\begin{figure}
    \centering
    \begin{subfigure}[t]{0.98\textwidth}
        \centering
        \includegraphics[width=\textwidth]{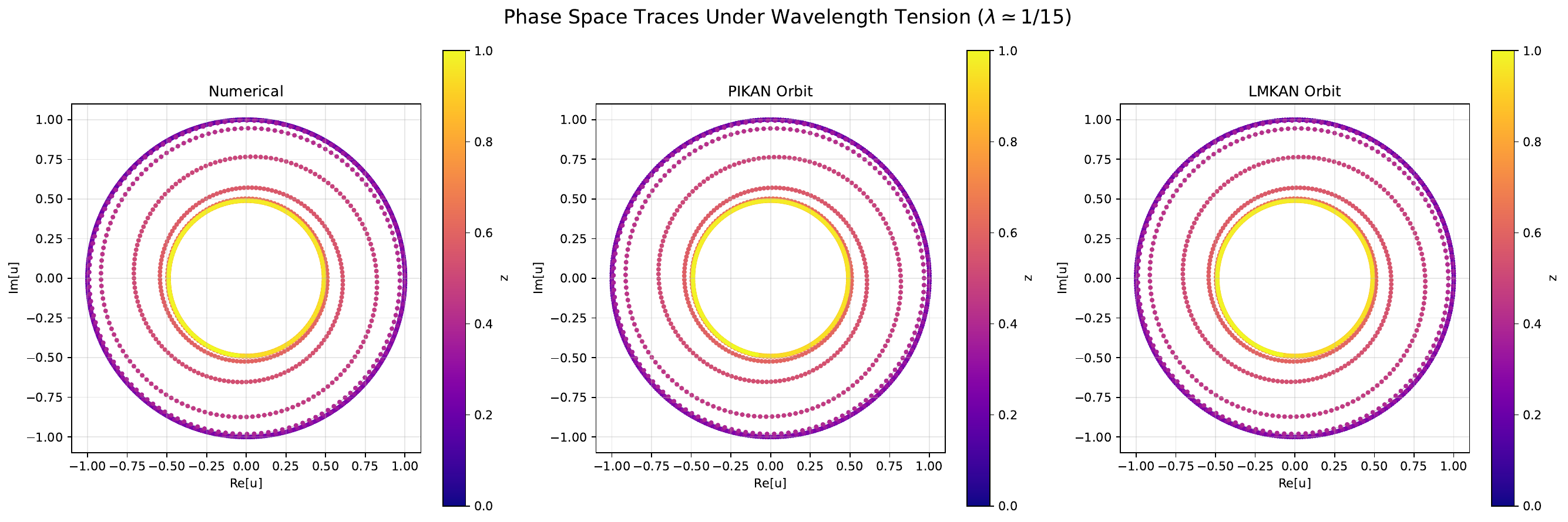}
        \caption{\(\lambda=1/15\)}
        \label{fig:helmholtz_phase_115}
    \end{subfigure}
    \hfill
    \begin{subfigure}[t]{0.98\textwidth}
        \centering
        \includegraphics[width=\textwidth]{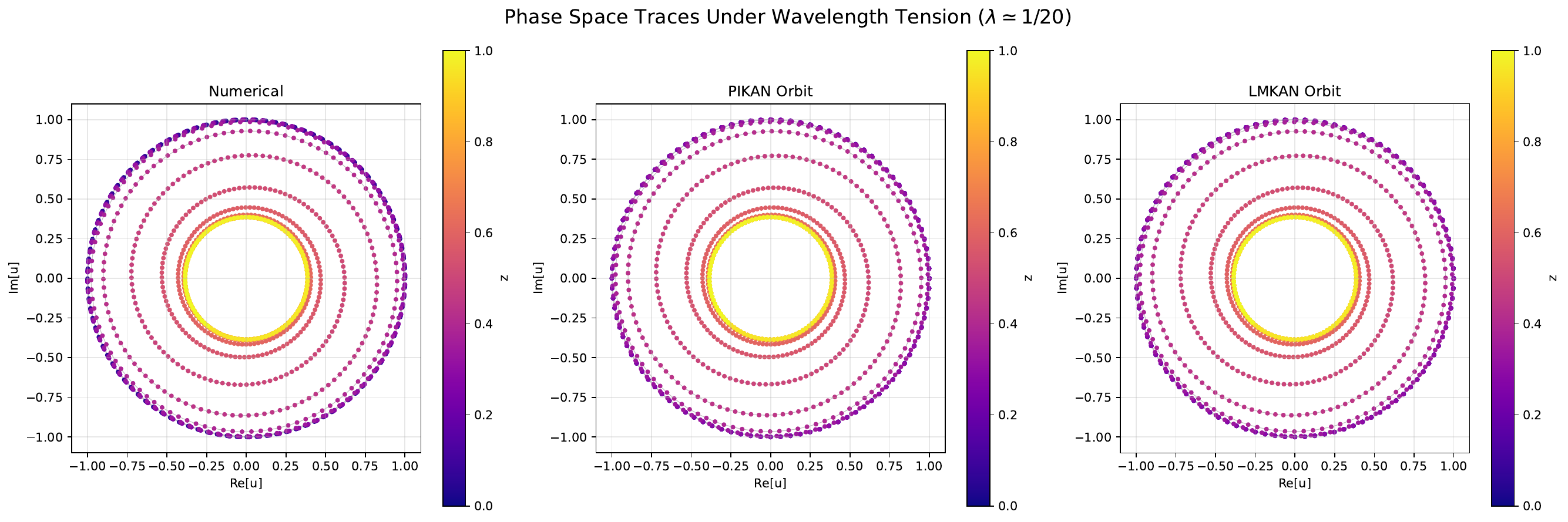}
        \caption{\(\lambda=1/20\)}
        \label{fig:helmholtz_phase_120}
    \end{subfigure}
    \hfill
    \begin{subfigure}[t]{0.98\textwidth}
        \centering
        \includegraphics[width=\textwidth]{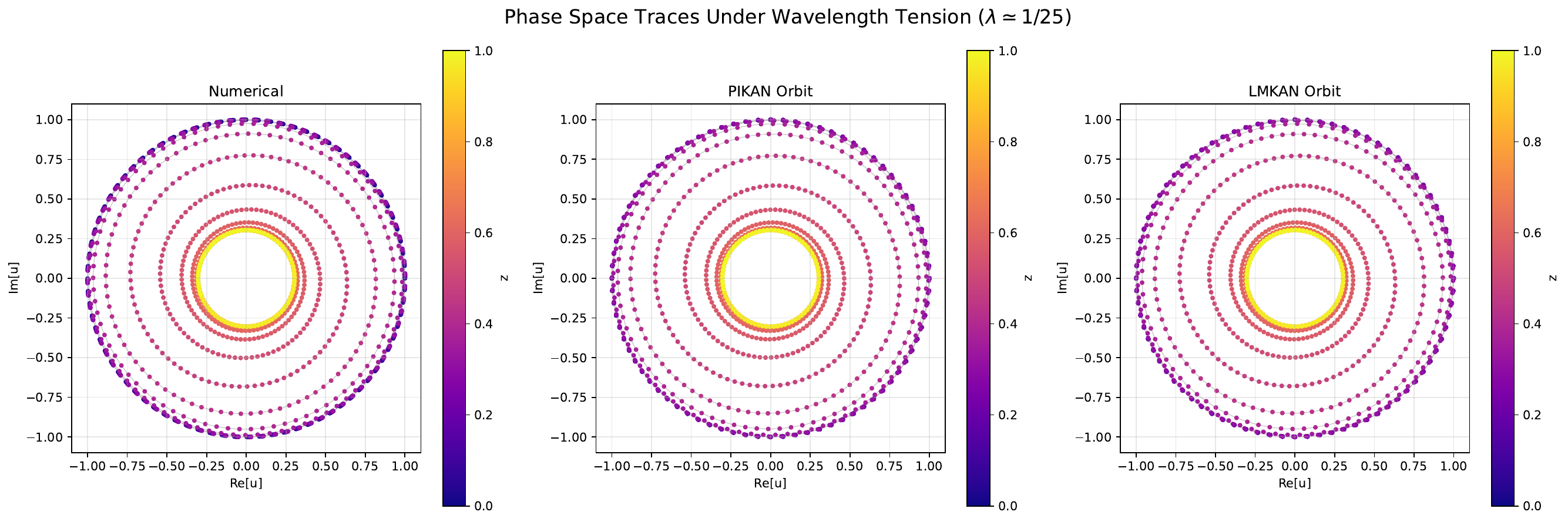}
        \caption{\(\lambda=1/25\)}
        \label{fig:helmholtz_phase_125}
    \end{subfigure}
    \caption{Phase-space traces \((\Re[u],\Im[u])\) for the Helmholtz benchmark. The reference, EfficientKAN, and LM-KAN trajectories are shown for each wavelength. Preservation of these nested orbits provides a geometric diagnostic of phase consistency in the learned complex-valued field.}
    \label{fig:helmholtz_phase_all}
\end{figure}

A further geometric view is provided by the phase-space traces in Fig. ~\ref{fig:helmholtz_phase_all}, where the complex field is visualized through its trajectory in the \((\Re[u],\Im[u])\) plane. At \(\lambda=1/15\), both methods reproduce the nested elliptical orbit structure very well. At \(\lambda=1/20\), the orbit geometry remains well preserved, but small deviations become more visible as the loops tighten and the field becomes more oscillatory. At \(\lambda=1/25\), LM-KAN again appears slightly closer to the reference nested structure, especially in maintaining the shape and spacing of the inner and outer loops. Since these phase portraits encode the joint evolution of real and imaginary components, they provide strong evidence that LM-KAN is not merely matching amplitude pointwise, but is also preserving the global geometry of the complex wave more accurately. 

\begin{figure}
    \centering
    \includegraphics[width=0.72\textwidth]{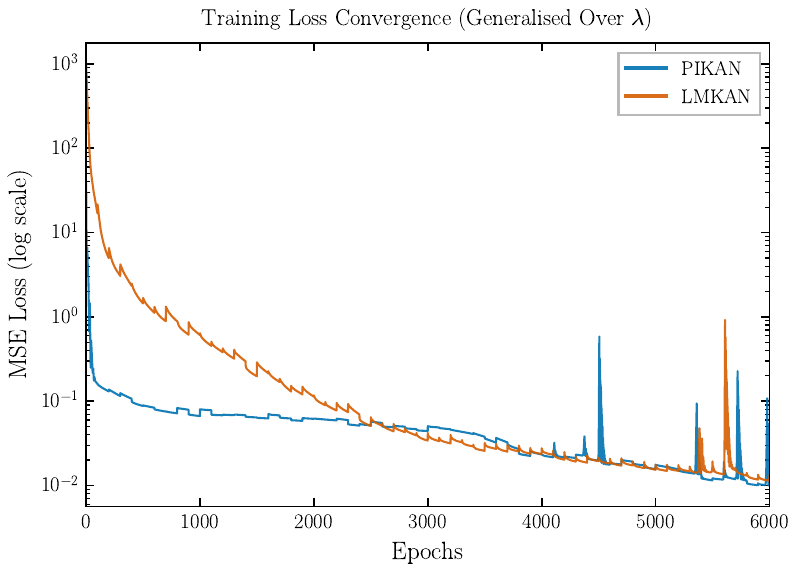}
    \caption{Training loss convergence for the Helmholtz benchmark. LM-KAN begins from a larger initial loss, but its optimization progresses more efficiently and reaches a lower final loss than PIKAN, indicating improved convergence for this oscillatory complex-valued problem.}
    \label{fig:helmholtz_training_loss}
\end{figure}

Fig.~\ref{fig:helmholtz_training_loss} compares the training behavior of the two models, and Table~\ref{tab:train_compare} summarizes the corresponding loss values and runtimes. Both networks reduce the physics loss substantially during training, but they do so in different ways. The EfficientKAN-based PIKAN model starts from a much smaller initial loss, whereas LM-KAN begins from a considerably larger value. Despite this unfavorable starting point, LM-KAN converges more effectively as training proceeds and reaches the lower final logged loss, \(4.83\times10^{-3}\), compared with \(8.79\times10^{-3}\) for EfficientKAN. This corresponds to an approximate \(45\%\) reduction in final loss. EfficientKAN does attain a slightly smaller temporary minimum during training, \(4.49\times10^{-3}\) versus \(4.83\times10^{-3}\), but it does not preserve that advantage through the end of optimization. In addition, LM-KAN completes training in \(76.77\) seconds, whereas EfficientKAN requires \(166.42\) seconds, so the LM-KAN model is more than twice as fast and reduces training time by about \(53.9\%\). Since the two architectures are essentially parameter matched, with 1920 parameters for PIKAN and 1928 for LM-KAN, this improvement cannot be attributed to model size alone. Rather, it suggests that the learned metric together with the Fourier basis provides a more effective inductive bias for the multi-wavelength Helmholtz problem, leading to faster and better final convergence.

\begin{table}
\centering
\caption{Comparison between PIKAN and LM-KAN for the Helmholtz benchmark. LM-KAN attains a lower final loss and trains substantially faster, while the two models remain nearly matched in parameter count. EfficientKAN reaches a slightly smaller temporary minimum during training, but does not retain this advantage at the end of optimization.}
\renewcommand{\arraystretch}{1.2}
\begin{tabular}{c|c|c|c|c|c}
\hline
Model & Parameters & Initial loss & Best logged loss & Training time (s) & Final loss \\
\hline
\textbf{PIKAN}  & 1920 & $5.95\times10^{1}$ & $4.49\times10^{-3}$ & 166.42 & $8.79\times10^{-3}$ \\
\textbf{LM-KAN} & 1928 & $2.73\times10^{3}$ & $4.83\times10^{-3}$ & 76.77 & $4.83\times10^{-3}$ \\
\hline
\end{tabular}
\label{tab:train_compare}
\end{table}

Overall, the Helmholtz results show a clear and consistent trend. At the largest tested wavelength, both models already provide accurate approximations of the reference solution. As the wavelength decreases and the solution becomes more oscillatory, the advantage of LM-KAN becomes more apparent. In the one-dimensional wavefield comparisons, LM-KAN remains slightly closer to the reference in the real and imaginary components and generally exhibits a smaller final error profile. In the spatiotemporal visualizations, it preserves the stripe structure of the propagating field more faithfully as oscillatory complexity increases. In the phase-space plots, it also maintains better agreement with the reference orbit geometry, which indicates stronger preservation of the joint real-imaginary wave dynamics.

The training results reinforce the same conclusion. Although EfficientKAN briefly reaches a slightly lower intermediate minimum, LM-KAN attains the better final solution and does so in substantially less time. This is especially significant because the two models are compared under an almost identical parameter budget. Taken together, these observations show that LM-KAN provides the stronger overall performance for the Helmholtz benchmark, offering a better balance of final accuracy and computational efficiency.

From the perspective of the present work, this result is important because the Helmholtz problem directly probes whether learned geometric adaptation is beneficial in highly oscillatory regimes. The answer suggested by these experiments is affirmative. LM-KAN appears better able to allocate representational capacity to regions where the complex wavefield changes rapidly, and this leads to improved optimization, better structural fidelity, and more robust behavior as the wavelength decreases. Thus, the Helmholtz benchmark supports the broader claim that geometry-adaptive KAN models are particularly effective for wave problems with multi-scale oscillatory structure.

\section{Discussion and Conclusion}
\label{diss}

In this work, we introduced Geometric KAN (GeoKAN), a family of geometry-aware Kolmogorov--Arnold models for supervised function approximation and physics-informed differential equation solving. The main novelty of GeoKAN is that it does not construct the representation only in fixed Euclidean coordinates. Instead, it learns a task-dependent metric and uses that learned geometry to define the coordinates in which the features are built. This is the key difference from earlier models. MLPs learn nonlinear maps on a fixed domain, while standard KANs improve the representation by replacing scalar weights with learnable univariate functions that are still evaluated directly on the original inputs. GeoKAN extends this idea by learning the geometry itself. In the layerwise formulation (Eqns. \eqref{eq:generic_geokan_layer}--\eqref{eq:generic_geokan_readout}), the model first learns a positive metric, then forms warped coordinates, and only afterwards constructs the features. In the physics-informed setting, the loss remains the standard residual-based one in Eq.~\eqref{eq:pikan_general_loss}. The novelty is therefore architectural rather than variational. Within this framework, LM-KAN, defined in Eq.\eqref{eq:lm_forward_expanded_rewrite} and Eq. \eqref{eq:lm_forward_expanded_fourier}, is the main geometry-aware surrogate considered in this work.

The supervised data-fitting experiments show that this geometric adaptation is useful across a broad range of targets. In the matched-capacity benchmark, GeoKAN models outperform the MLP baseline and often also Efficient-KAN on functions with oscillatory, localized, discontinuous, and multiscale structure. As shown in Table~\ref{tab:datafit_main_results}, GeoKAN-NNMetric and LM-KAN-Wav achieve the best or tied-best results on several of the most difficult targets, including the high-frequency sinusoid, the step discontinuity, the narrow needle, the multiscale envelope-modulated sinusoid, and the sawtooth. The same overall trend is visible in Fig.~\ref{fig:datafit_all_targets} and Fig.~\ref{fig:datafit_selected_targets}, where the geometry-aware models track the difficult targets more closely than the fixed-geometry baselines. Since the parameter counts in Table~\ref{tab:datafit_param_counts} are closely matched, these gains cannot be attributed simply to larger models. Rather, they indicate that the learned geometric warp changes how approximation power is distributed across the domain. Regions with rapid variation, sharp transitions, or fine-scale structure can be represented more effectively, while smoother regions need less representational effort. This is precisely the type of setting in which a geometry-adaptive construction should be expected to help, and the data-fitting results provide clear evidence that it does.

The physics-informed experiments lead to the same general conclusion. We studied Allen--Cahn equations in two regimes, Burgers' equation, the Lorenz system, and the Helmholtz equation, given by Eqns. \eqref{allen_1}, \eqref{allen_2}, \eqref{burger}, \eqref{lorentz_equation}, and \eqref{helmholtz}. These problems cover different types of difficulty, including moving interfaces, steep gradients, nonlinear coupling, and oscillatory wave behavior. For Allen--Cahn Case~1, both models recover the solution accurately. The main difference is that PIKAN reaches the smaller final loss, while LM-KAN trains much faster. For Allen--Cahn Case~2, which is the harder regime, LM-KAN gives the better overall approximation. This suggests that the learned metric helps the model resolve sharper interface structure, although it comes with a higher training cost. For Burgers' equation, both models again produce very accurate solutions, but LM-KAN reaches nearly the same final loss in less wall-clock time. This indicates that the learned warp is useful when the solution develops localized steepening. For the Lorenz system, both methods reproduce the trajectories well, while LM-KAN attains a lower final loss and PIKAN remains faster. This shows that the benefit of geometry adaptation is not limited to PDEs with spatial structure, but can also appear in coupled nonlinear ODE systems. The Helmholtz equation gives the strongest evidence in favor of the proposed approach. In this case, the two models are nearly parameter matched, so the comparison mainly reflects the architecture itself. LM-KAN preserves the oscillatory structure more robustly, especially as the wavelength decreases, and it reaches a lower final loss in substantially less time. Since this experiment uses the Fourier version of LM-KAN, it also shows that performance improves when the basis is well matched to the structure of the target problem.

Overall, the experiments show that GeoKAN is most effective when the target has non-uniformity across the domain. Fixed-geometry models tend to distribute representational capacity more uniformly, whereas GeoKAN can adapt that allocation through the learned metric. This interpretation is consistent with Fig.~\ref{fig:kan_geokan} and Table~\ref{tab:geokan_taxonomy_short}, and it helps explain why the gains are strongest for localized, multiscale, and oscillatory problems. At the same time, GeoKAN is not uniformly superior in every respect. In some cases, such as Allen--Cahn Case~1, PIKAN still achieves the lowest final loss. In other cases, such as Allen--Cahn Case~2 and the Lorenz system, LM-KAN improves the final approximation but requires more training time. The most balanced conclusion is therefore not that GeoKAN dominates all earlier models in every metric, but that it provides a more flexible and often more effective representational mechanism, especially for problems with sharp transitions, localized structure, or oscillatory complexity.

Several directions remain open for future work. One natural extension is to move beyond diagonal metrics and consider anisotropic or fully coupled metric tensors, especially for higher-dimensional PDEs where directional effects may matter. Another direction is to make the basis more adaptive, since the present results already show that wavelet, RBF, and Fourier variants have different strengths. It would also be valuable to test GeoKAN on more demanding scientific machine-learning tasks, including higher-dimensional PDEs, inverse problems, operator learning, and wave propagation in heterogeneous media.

\section*{Data and code availability}
All codes used in this study can be found in \href{https://github.com/AI-and-Quantum-Computing/GeoKAN}{https://github.com/AI-and-Quantum-Computing/GeoKAN}.

The proposed model was implemented and evaluated on a Linux-based system (6.8.0-57-generic) powered by a 56-core x86\_64 CPU (112 logical processors) and 754.53 GB of system memory.

\section*{Acknowledgments}

This work was supported by Army Research Office (ARO) (grant W911NF-23-1-0288; program manager Dr.~James Joseph). The views and conclusions contained in this document are those of the authors and should not be interpreted as representing the official policies, either expressed or implied, of ARO, or the U.S. Government. The U.S. Government is authorized to reproduce and distribute reprints for Government purposes notwithstanding any copyright notation herein.

\FloatBarrier
\bibliographystyle{unsrt}
\bibliography{references}

\end{document}